\documentclass[preprint,12pt]{elsarticle}

\usepackage{amssymb}
\usepackage[acronym]{glossaries}
\usepackage{csquotes}
\usepackage{multirow}
\usepackage{multicol}
\usepackage{hyperref}
\usepackage{booktabs}
\usepackage{threeparttable}
\usepackage[top=2.0cm,left=2.3cm,right=2.3cm,bottom=2.0cm]{geometry}
\usepackage{verbatim}
\usepackage{setspace}
\usepackage{array}
\usepackage{enumitem}

\usepackage{longtable}
\usepackage{pdflscape}

\usepackage{graphicx}

\usepackage[font=normalsize]{caption}

\newacronym{bpm}{BPM}{business process management}
\newacronym{ai}{AI}{artificial intelligence}
\newacronym{ml}{ML}{machine learning}
\newacronym{kpi}{KPI}{key performance indicator}
\newacronym{nlp}{NLP}{natural language processing}
\newacronym{xai}{XAI}{explainable artificial intelligence}
\newacronym{rpa}{RPA}{robotic process automation}
\newacronym{gan}{GAN}{generative adversarial network}
\newacronym{gnn}{GNN}{graph neural network}
\newacronym{cnn}{CNN}{convolutional neural network}
\newacronym{is}{IS}{information systems}

\newacronym{svm}{SVM}{support vector machine}
\newacronym{nn}{NN}{neural network}
\newacronym{dnn}{DNN}{deep neural network}
\newacronym{lstm}{LSTM}{long short-term memory}
\newacronym{mlp}{MLP}{multi-layer perceptron}
\newacronym{gru}{GRU}{gated recurrent unit}
\newacronym{erp}{ERP}{enterprise resource planning}
\newacronym{crm}{CRM}{customer relationship management}

\journal{Expert Systems with Applications}
\biboptions{authoryear}
\begin{document}

\begin{frontmatter}



\title{Machine Learning in Business Process Management: A Systematic Literature Review}

\author[add1]{Sven Weinzierl}\corref{cor1}
\ead{sven.weinzierl@fau.de}
\author[add2]{Sandra Zilker}
\ead{sandra.zilker@th-nuernberg.de}
\author[add1]{Sebastian Dunzer}
\ead{sebastian.dunzer@fau.de}
\author[add1]{Martin Matzner}
\ead{martin.matzner@fau.de}

\cortext[cor1]{Corresponding author}

\address[add1]{Institute of Information Systems, Friedrich-Alexander-Universit\"at N\"urnberg-Erlangen, Germany}
\address[add2]{Professorship for Business Analytics, Technische Hochschule Nürnberg Georg Simon Ohm, Germany}

\begin{abstract}
Machine learning (ML) provides algorithms to create computer programs based on data without explicitly programming them.
In business process management (BPM), ML applications are used to analyse and improve processes efficiently.
Three frequent examples of using ML are providing decision support through predictions, discovering accurate process models, and improving resource allocation.
This paper organises the body of knowledge on ML in BPM. We extract BPM tasks from different literature streams, summarise them under the phases of a process's lifecycle, explain how ML helps perform these tasks and identify technical commonalities in ML implementations across tasks.
This study is the first exhaustive review of how ML has been used in BPM. We hope that it can open the door for a new era of cumulative research by helping researchers to identify relevant preliminary work and then combine and further develop existing approaches in a focused fashion. Our paper helps managers and consultants to find ML applications that are relevant in the current project phase of a BPM initiative, like redesigning a business process.
We also offer -- as a synthesis of our review -- a research agenda that spreads ten avenues for future research, including applying novel ML concepts like federated learning, addressing less regarded BPM lifecycle phases like process identification, and delivering ML applications with a focus on end-users.
\end{abstract}


\begin{highlights}

\item We describe BPM lifecycle phases and define BPM tasks in view of ML applications.
\item We present an overview of ML applications in the field of BPM.
\item We provide a technical summary of ML applications of each BPM lifecycle phase.
\item We derived ten findings from our literature review.
\item We set up an agenda with ten future research directions for advancing research on ML applications in BPM.

\end{highlights}


\end{frontmatter}


\section{Introduction}
\label{sec:introduction}
Organisations have made great strides in digitising their business processes~\citep{beverungen2021seven}, so information systems now produce large amounts of process data~\citep{vanderAalst2016action}.
\Gls{bpm} research offers approaches to create value from such data~\citep{vanderAalst2016action}, including process mining~\citep[e.g.][]{van.2011}, business activity monitoring \citep[e.g.][]{mccoy2002business}, predictive business process monitoring~\citep[e.g.][]{grigori.2004}, and anomaly detection \citep[e.g.][]{bezerra2009anomaly}. 
\Gls{ml} is popular as an analytical capability at the core of \gls{bpm} approaches.
Because of the increased availability of event data, the emergence of off-the-shelf \gls{ml} libraries, and advances in hardware, \gls{ml} approaches are increasingly used to solve \gls{bpm} tasks. 
According to \citet[p.~xv]{mitchell.1997}, \textquote{[t]he field of machine learning is concerned with the question of how to construct computer programs that automatically improve with experience}. 
\Gls{ml} provides algorithms that learn patterns and structures from data (e.g. examples or observations) and capture those in mathematical models (e.g. functions) through automated improvement procedures~\citep{bishop2006pattern}.

Before \gls{ml} can be used in applications tailored to specific \gls{bpm} tasks (e.g. anomaly detection), underlying \gls{ml} models are developed by completing the following four steps: Data input, feature engineering, model building, and model assessment~\citep{janiesch2021machine,goodfellow2016deep,liu1998feature}. 
Once these models are developed, \gls{ml} applications can create value from process data, such as cost reduction or risk mitigation~\citep{marquez2017predictive}. In short, \gls{ml} supports organisations in improving their business processes~\citep{mendling2018machine}.

While \gls{ml} in \gls{bpm} bears considerable potential for improving organisational operations, we believe that an overview of \gls{ml} applications in \gls{bpm} can help scholars propose contributions in this research area and practitioners find existing \gls{ml} applications that address a problem they are facing.
This systematic literature review contributes to research and practice in the following five ways:

\begin{enumerate}
    \item We describe \gls{bpm} lifecycle phases and define \gls{bpm} tasks in view of \gls{ml} applications. 
    \item We present an overview of \gls{ml} applications in the field of \gls{bpm}.
    \item We provide a technical summary of \gls{ml} applications of each \gls{bpm} lifecycle phase. 
    \item We derived findings from our literature review. 
    \item We set up a future research agenda for advancing research on \gls{ml} applications in \gls{bpm}.
\end{enumerate}

The remainder of this paper is structured as follows. After we present the core concepts of \gls{bpm} and \gls{ml} in Section~\ref{sec:background}, we provide an overview of related literature reviews in Section~\ref{sec:related}. 
Then, in Section~\ref{sec:research}, we describe our applied research method, after which we provide an overview of \gls{ml} applications in \gls{bpm}, and describe \gls{bpm} lifecycle phases and define \gls{bpm} tasks in the view of \gls{ml} applications (Section~\ref{sec:results}). Subsequently, in Section~\ref{sec:synthesis_results_findings}, we synthesise the results from the previous section and present findings, which we derived from the synthesised results. In Section~\ref{sec:discussion}, we present directions for future research, implications for research and practice, and limitations of our literature review. 

\section{Background} 
\label{sec:background}
 
This paper focuses on applications based on \gls{ml} models in the \gls{bpm} domain. 
Therefore, in this section, we provide overviews of \gls{bpm} and its lifecycle, as well as the process of \gls{ml} model development in general. 

\subsection{BPM and its Lifecycle}
\label{subsec:-bpm}
\begin{figure}[ht]
    \centering
    \includegraphics[width=0.7\textwidth]{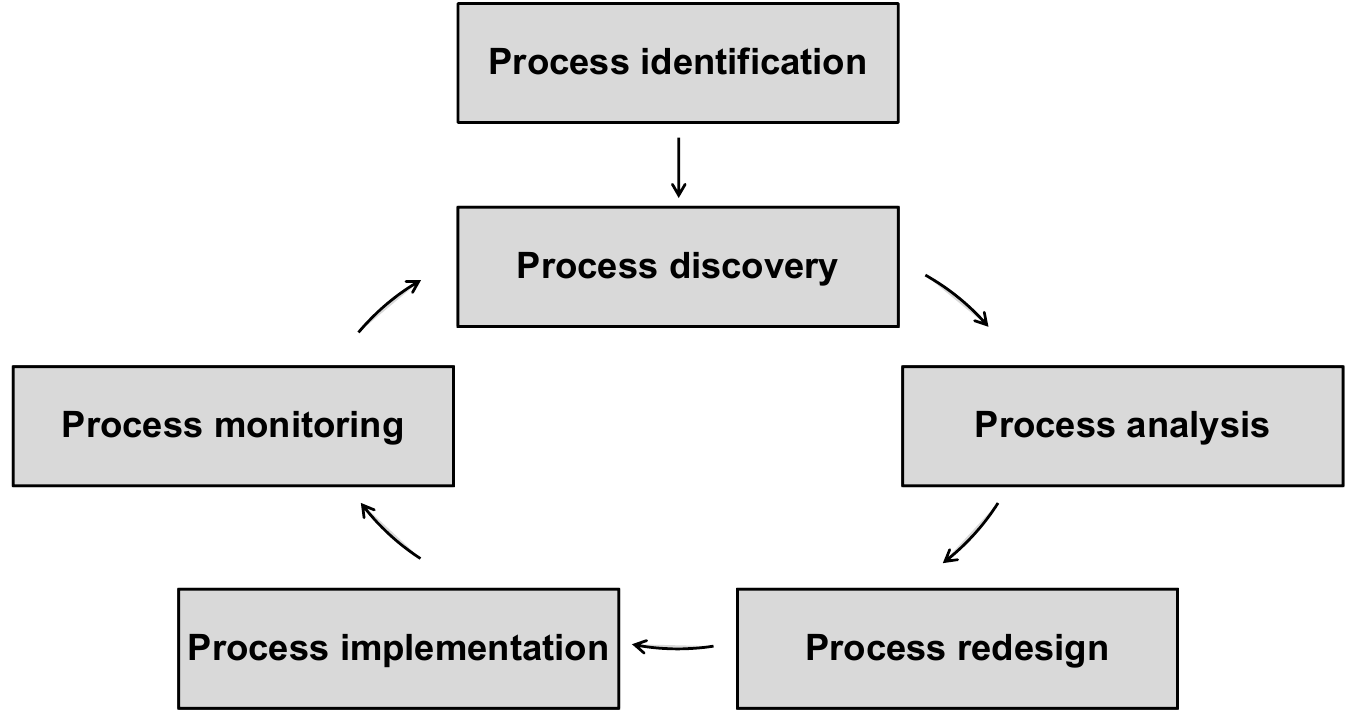}
    \caption{\gls{bpm} lifecycle based on \protect\citet{dumas2018fundamentals}}
    \label{fig:bpmlifecycle}
\end{figure}

\gls{bpm} is a management discipline aiming to increase a business's competitive advantage by facilitating continuous improvement in organisational operations~\citep{Trkman2010}.
To achieve this, \gls{bpm} proposes management practices that cultivate an end-to-end view of customer-oriented processes across functional boundaries~\citep{Trkman2010}. 
%
%
\Gls{bpm} tasks are classified along lifecycle models to analyse, redesign, implement, and monitor business processes continuously~\citep{DeMorais2014}.
For this paper, we align with \citet{Recker2016} in using the \gls{bpm} lifecycle of \citet{dumas2018fundamentals} to classify existing \gls{ml} applications in \gls{bpm} tasks.
We also consider definitions from other \gls{bpm} lifecycle models in describing each of the phases \citep[e.g.][]{DeMorais2014,Houy2010,Muehlen2006}.
\gls{bpm} lifecycle model of \citet{dumas2018fundamentals} has six phases through which (except \textit{process identification}, the entry point to \gls{bpm}) business processes pass multiple times to facilitate constant improvement~(see Figure~\ref{fig:bpmlifecycle}).  

\textbf{\textit{Process identification.}} The process identification phase identifies, delimits, and interrelates processes and their stakeholders. Taking a cross-functional perspective allows processes' architectures to be set up~\citep{dumas2018fundamentals} that aid in identifying the processes that should be improved along the lifecycle~\citep{Houy2010, Trkman2010}. An activity at this stage includes analyses of process environment and organisation~\citep{Muehlen2006}.

\textbf{\textit{Process discovery.}} The process discovery phase documents the current state 
of processes in the form of as-is process models.\footnote{Other lifecycle models refer to this phase as process design~\citep{DeMorais2014}.} To enhance documentation, organisations may, for instance, specify a process in detail, model it in formal process-modelling languages, and conduct process walkthroughs~\citep{Houy2010,Muehlen2006,Zairi1997}.

\textbf{\textit{Process analysis.}} In the process analysis phase, the resulting process documentations and as-is models are assessed to identify the issues related to a process~\citep{dumas2018fundamentals}. Examples of activities in this phase are simulating a process, calculating cost and cycle-time, applying process mining, and defining target metrics~\citep{DeMorais2014,Muehlen2006,Zairi1997}.
 
\textbf{\textit{Process redesign.}} Based on the results of the process analysis phase, the process redesign phase elaborates changes to processes that will improve them~\citep{dumas2018fundamentals}, resulting in redesigned to-be process models that include information for process operation~\citep{Houy2010,DeMorais2014}. Process analysis tools can help to determine process changes that should be implemented~\citep{dumas2018fundamentals}.  

\textbf{\textit{Process implementation.}} Redesigned processes must be embedded in organisations' information systems~\citep{Muehlen2006} by using organisational change mechanisms to facilitate work according to redesigned processes and adjusting the IT systems required for the redesigned processes to be executed. In short, the process implementation phase yields executable process models~\citep{dumas2018fundamentals}.

\textbf{\textit{Process monitoring.}} During the execution of a redesigned process, conformance and performance are controlled using data about process executions \citep{Houy2010}. Corrective actions can be taken if errors occur or a process faces performance issues. However, if new issues arise, the lifecycle must be repeated~\citep{dumas2018fundamentals}.

\subsection{ML Model Development} 
\label{subsec:-ML}

\begin{figure}[ht]
    \centering
    \includegraphics[width=\textwidth]{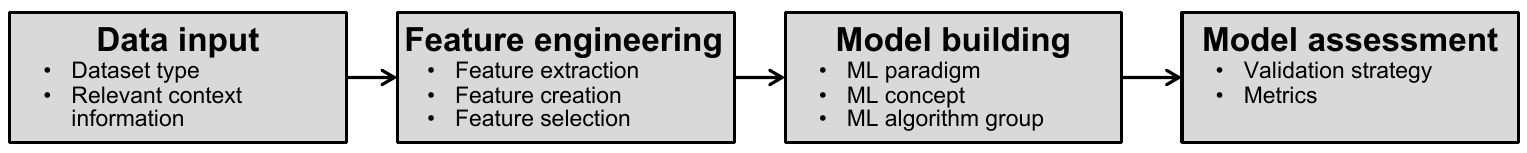}
    \caption{\gls{ml} model development phases based on \protect\citet{janiesch2021machine}, \citet{goodfellow2016deep}, and \citet{liu1998feature}}
    \label{fig:modeldevelopment}
\end{figure}

The process of developing an \gls{ml} model consists of four phases, as depicted in Figure~\ref{fig:modeldevelopment}, enriched with relevant concepts~\citep{janiesch2021machine,goodfellow2016deep,liu1998feature}. In what follows, these four phases are described.

\textbf{\textit{Data input.}} Data of various types can be inputs to \gls{ml} models. For example, a dataset can consist of tabular, event log, or text data. The perspectives to which the event characteristics refer, such as time, resources, or data flow, can also be included for event logs~\citep{leoni2016general}.

\textbf{\textit{Feature engineering.}} 
Feature engineering, which consists of feature extraction, creation, and selection, is performed to transform and simplify data, thus facilitating model building~\citep{liu1998feature}. 
%
%
Generally, a feature is an independent variable in a dataset \citep{hastie2009elements}. 
Feature extraction retrieves features from the data input and encodes them (e.g. categorical features as one-hot encoded features), (manual) feature creation adds additional features (e.g. temporal features created based on the control-flow and timestamp information of an event log), and feature selection removes features that, for example, have little effect on the learning target.

\textbf{\textit{Model building.}} \gls{ml} algorithms are applied to build mathematical models based on data \citep{bishop2006pattern}. As the research field \gls{ml} provides many \gls{ml} algorithms, we differentiate them based on three dimensions \citep{janiesch2021machine}: 
The \emph{\gls{ml} paradigm} they follow, a dimension that distinguishes among the approaches supervised, unsupervised, and reinforcement learning and those in-between (i.e. semi- and self-supervised learning); the \emph{\gls{ml} concept}, which refers to the approach (e.g. deep learning) an \gls{ml} algorithm uses to address a learning problem in terms of such aspects, as the expected input data, the number of \gls{ml} models, or the learning target; and the \emph{\gls{ml} algorithm group}, referring to how the algorithm builds models from data. Algorithms of the same group follow similar principles for model building, including model structure and parameter fitting. Table~\ref{tab:ml-approaches} shows an overview of the ML dimensions.

\textbf{\textit{Model assessment.}} Several validation strategies can be used to test model generalisability, and metrics can be used to measure model properties~\citep[e.g.][]{hastie2009elements}. 
While validation strategies include split validation and cross-validation, metrics can refer to quality (e.g. accuracy for classification models \citep{ferri2009experimental}) and runtime (e.g. training time for model training~\citep{wang2012efficient}).

\begin{table}[ht!]
\centering
\caption{Overview of \gls{ml} dimensions}
\label{tab:ml-approaches}
\scriptsize
\resizebox{\textwidth}{!}{
\begin{threeparttable}
\begin{tabular}{p{0.1cm}p{3cm}p{14cm}}
\toprule
&\textbf{Dimension} & \textbf{Description} \\ \midrule
& Supervised learning & The algorithm learns from examples \citep{Alpaydin2014}; that is, it learns from data instances that are labelled with their correct output \citep{Kotsiantis2006}.\\
&Semi-supervised learning& The algorithm uses labelled input data and unlabelled input data for model training \citep{Zhu2005SemiSupervisedLL}, which is more effective than unsupervised learning and needs less human effort than supervised learning.\\
\multirow{5}*{\rotatebox[origin=c]{90}{\textbf{\gls{ml} paradigm}}} & Reinforcement learning &
The input data for reinforcement learning algorithms is generated through the algorithm's interactions with its environment and changes over time~\citep{Alpaydin2014}. Hence, the algorithm relies on feedback that is received during training and is rewarded for steps towards or achieving a desired objective and punished otherwise \citep{Brynjolfsson2017,Russel2016,Kaelbling1996}. Therefore, the algorithm is like an agent that interacts with a dynamic environment via perception and action \citep{Kaelbling1996}.\\
&Self-supervised learning&Using unlabelled input data \citep{jaiswal2021survey, Longlong2021} and training an \gls{nn} model are based on pseudo-labels created from input data to solve a \textquote{pretext task} (e.g. predicting a word based on its surrounding words in a sentence). Then the model is used to solve a \textquote{downstream task} using supervised or unsupervised learning. \\
&Unsupervised learning & Identifying and learning inherent patterns in the input data is based on structural properties \citep{jordan2015machine}. 
In contrast to supervised learning, there are no labelled examples for model training~\citep{Russel2016}.\\

\midrule
&Multi-task learning & The algorithm learns a model that addresses multiple related problems \citep[e.g.][]{zhang2021survey}. \\
&Active learning & An accurate model can be learned with a low volume of labelled training instances if an \gls{ml} algorithm is allowed to choose the training data from which it learns~\citep[e.g.][]{settles2009active}.\\
\multirow{9}*{\rotatebox[origin=c]{90}{\textbf{\gls{ml} concept}}}&Online/ incremental learning & The algorithm uses the available data and updates the model before a prediction is required or after the last observation is made \citep[e.g.][]{blum1998line}.\\
& Transfer learning & The algorithm first learns a model on a task. Then part or all of the model is used as the starting point for a related task \citep[e.g.][]{pan2009survey}. \\
& Ensemble learning & The algorithm learns two or more models on the same training set and the outputs from all models are combined \citep[e.g.][]{dietterich2002ensemble}. \\
& Deep learning & The algorithm learns multiple representations instead of a single representation to identify complex structures in data \citep[e.g.][]{lecun2015deep}. \\
& Meta learning & 
The algorithm learns from the output of other \gls{ml} algorithms that learn from a training set~\citep[e.g.][]{nichol2018first}.\\ 
&Federated learning & Multiple data owners collectively learn and use a shared model while keeping all of the local training data private \citep[e.g.][]{yang2019federated}.
\\
&Multi-view learning& The algorithm learns one or more models from multiple views of training data \citep[e.g.][]{xu2013survey}.  \\ \midrule

&Instance-based & \gls{ml} algorithms build models directly from the training instances (e.g. support vector machine \citep{vapnik1995support}).
\\
&Decision tree-based & \gls{ml} algorithms build tree-structured models, in which leaves represent class labels and branches represent conjunctions of the features that lead to these labels (e.g. classification and regression trees \citep{breiman1984classification}).
\\
\multirow{9}*{\rotatebox[origin=c]{90}{\textbf{\gls{ml} algorithm group}}}&Clustering-based & \gls{ml} algorithms build models to discover natural groups (i.e. clusters) in the data's feature space (e.g. k-means \citep{forgy1965cluster}).
\\
&Artificial-neural-network-based & \gls{ml} algorithms build models consisting of one or more neurons that are connected via edges and structured into one or more layers from input to output (e.g. multi-layer perceptron \citep{rumelhart1985learning}).
\\
&Regression-based & \gls{ml} algorithms build models by separating data points through a fitted regression line (e.g. logistic regression \citep[e.g.][]{hastie2009elements}).
\\
&Bayesian-based & \gls{ml} algorithms build models using Bayesian statistics (e.g. na\"ive Bayes \citep[e.g.][]{lewis1998naive}).
\\
&Rule-based & \gls{ml} algorithms build models as a set of relational rules that represent knowledge (e.g. RIPPER \citep{COHEN1995115}).\tnote{1}
\\
&Reinforcement-learning-based & \gls{ml} algorithms build models on how an agent should behave in a particular environment by performing actions and observing the results (e.g. Q-learning \citep{watkins1989learning}).
\\
&Genetic-based & \gls{ml} algorithms build models by iteratively updating a population (pool of hypotheses), evaluating each population member based on a fitness measure, and selecting the best fitting members to produce the next generation (e.g. genetic algorithm \citep[e.g.][]{de1993using}).
\\
\bottomrule
\end{tabular}
\begin{tablenotes}
    \item[1] In this study, we associate the term \textquote{rule-based} with rule-based \gls{ml} \citep{furnkranz2012foundations}, where some form of a learning algorithm is applied to automatically identify a set of relational rules for pattern detection or prediction, and not traditional rule-based \gls{ai} systems  \citep{hayes1985rule}, where a human expert with domain knowledge manually constructs a set of rules for knowledge representation or reasoning.
\end{tablenotes}
\end{threeparttable}
}
\end{table}

\section{Related Literature Reviews}
\label{sec:related}
Previous literature reviews in \gls{bpm} consider \gls{ml} an approach to gain insights from process data. These reviews investigate the application of certain \gls{ml} types for certain \gls{bpm} lifecycle phases (e.g. process monitoring) and approaches (e.g. process mining).
One group of literature reviews focuses on predictive business process monitoring, where \gls{ml} is an approach for building predictive models.
\citet{marquez2017predictive} provide a global overview of the domain, while \citet{di2018predictive} investigate algorithms' tasks, input data, families, and tools. 
\citet{verenich2019survey}, \citet{teinemaa2019outcome}, and \citet{stierlexpbpm} focus on techniques for predicting remaining time, techniques for classification-based process outcome prediction, and techniques that use \gls{xai} approaches, while \citet{Neu2021} and \citet{rama2020deep} conduct a systematic literature review on deep-learning approaches for predictive business process monitoring.

Previous literature reviews also address process mining. 
\citet{tiwari2008review} perform a literature review on process mining, specifically process discovery.
\citet{maita2015process} conduct a systematic literature review on applying \glspl{nn} and \glspl{svm} for data-mining tasks in process mining.
\citet{Taymouri2021} conduct a systematic literature review on analysis methods for process variants, and consider \gls{ml} a family of algorithms used in these methods. 
\citet{folino2021ai} perform two systematic literature reviews, one for \gls{ai}~-based process-mining approaches exploiting domain knowledge; second, for process-mining approaches that address auxiliary \gls{ai} tasks jointly with target process-mining tasks. Both consider \gls{ml} a subset of \gls{ai}.  
\citet{herm2021symbolic} conduct a literature review on intelligent \gls{rpa}, in which the authors consider \gls{ml} an approach for transforming symbolic \gls{rpa} into intelligent \gls{rpa}. 
\citet{wanner2020machine} focus on the combination of \gls{ml} and complex event processing in their literature review.
Finally, \citet{koj2023anomaly} perform a systematic literature review on anomaly detection for business process event logs.
Table~\ref{tab:related} provides an overview of related (systematic) literature reviews and positions our work in this context.

\begin{table}[htb!]
\centering
\caption{Overview of related literature reviews}
\label{tab:related}
\scriptsize
\resizebox{\textwidth}{!}{
\begin{tabular}{@{}p{4.5cm}p{6cm}p{6cm}@{}}
\toprule
\textbf{Reference} & \textbf{Scope of review} & \textbf{Focus of review} \\ \midrule

\citet{tiwari2008review} 
& \textbf{General:} Papers published (1998) -- 2005
& \textbf{Broad:} Process mining, specifically process discovery
\\

\citet{maita2015process}          & \textbf{General:} Journal and conference papers published 2004 -- 2014 
& \textbf{Specific:} \Glspl{nn} and \glspl{svm} in process mining \\

\citet{marquez2017predictive}          & \textbf{Specific:} Journal and conference papers, and book chapters published 2010 -- 2017 
& \textbf{Specific:} Predictive business process monitoring                 \\
\citet{di2018predictive}                    &    \textbf{General:} Journal and conference papers published (2005) -- 2018
&  \textbf{Specific:} Predictive business process monitoring               \\
 \citet{verenich2019survey}                             &           
 \textbf{General:} Papers published 2005 -- 2017
 & 
 \textbf{Very specific:} Remaining-time-prediction task in predictive business process monitoring         \\
 
%
\citet{teinemaa2019outcome} & 
\textbf{General:} Papers published (2005) -- 2017 
 &
 \textbf{Very specific:} Outcome prediction task in predictive business process monitoring               \\
 \citet{stierlexpbpm}                                                 &        
 \textbf{Specific:} Journal and conference papers published (2014) -- 2020
 &  \textbf{Very specific:} \Gls{xai} approaches used in predictive business process monitoring               \\
        \citet{Neu2021}  & 
          \textbf{Very specific:} Papers published (2017) -- 2020
        &   \textbf{Very specific:} Deep-learning approaches used in predictive business process monitoring \\
\citet{rama2020deep}                   &
\textbf{Very specific:} Papers published (2017) -- 2020
&    
\textbf{Very specific:} Deep-learning approaches used in predictive business process monitoring 
\\
\citet{Taymouri2021}          
&
\textbf{General:} Papers published 2003 -- 2019 
&  
\textbf{Specific:} Business process variant analysis
\\                           

\citet{folino2021ai}                           &   
\textbf{General:} (i) Journal and conference papers, and book chapters published (2009) -- 2020; (ii) journal and conference papers, and book chapters published (2008) -- 2020
%
&   \textbf{Specific:} (i) \Gls{ai}-based process-mining approaches exploiting domain knowledge; (ii) process-mining approaches addressing auxiliary \gls{ai} tasks jointly with target process-mining tasks              \\               
\citet{herm2021symbolic}
& \textbf{Specific:} Papers published (2015) -- 2020
& \textbf{Specific:} \Gls{ai}-based \gls{rpa}
\\
\citet{wanner2020machine} & \textbf{General:} Papers published (2007) -- 2018& \textbf{Specific}: Combining complex event processing and \gls{ml}\\

\citet{koj2023anomaly} & \textbf{General:} Journal and conference papers published (2000) -- 2021& \textbf{Very specific}: Anomaly-detection task in pattern detection\\

\midrule

\textbf{Our work}          &   \textbf{General}: Journal and conference papers published (1998) -- 2022  & \textbf{Broad:} \gls{ml} applications for tasks in all \gls{bpm} lifecycle phases\\ \bottomrule
\end{tabular}
}
\begin{tablenotes}
\small
\item The year in round brackets indicates the year of the oldest paper in the final coding table.
\end{tablenotes}
\end{table}

Our review has a more general scope and a broader focus than the other reviews we mention. 
The scope of our review is general, as we do not limit the time horizon because we want to consider all \gls{bpm} papers that propose \gls{ml} applications, regardless of when they were published. 
The focus of our review is broad, as we consider \gls{ml} applications for \gls{bpm} that rely on any type of \gls{ml} and can tackle any \gls{bpm} task, regardless of the phase of the \gls{bpm} lifecycle to which they belong. 
%
Conducting a literature review with this scope and focus allows us to give an overview of how \gls{ml} has been applied in \gls{bpm}, describe \gls{bpm} lifecycle phases and define \gls{bpm} tasks in the view of \gls{ml} applications, provide a technical summary of \gls{ml} applications of each \gls{bpm} lifecycle phase, derive findings, and set up an agenda that can advance research on \gls{ml} applications in~\gls{bpm}.

\section{Research Method}
\label{sec:research}
We conducted a systematic literature review using a descriptive perspective and provide guidance based on the review's results~\citep{pare.2015}.
We follow \citet{vom2015standing} and the concept-centric notion of~\citet{webster2002analyzing}.
Further, in line with~\citet{beese2019simulation}, we carried out the review in two steps, first applying a structured \emph{search process} to gather relevant papers before developing a coding scheme and using a \emph{coding process} following \citet{Recker2016}. 

\subsection{Search Process}
\begin{figure}[htb!]
 \centering
 \includegraphics[width=\textwidth]{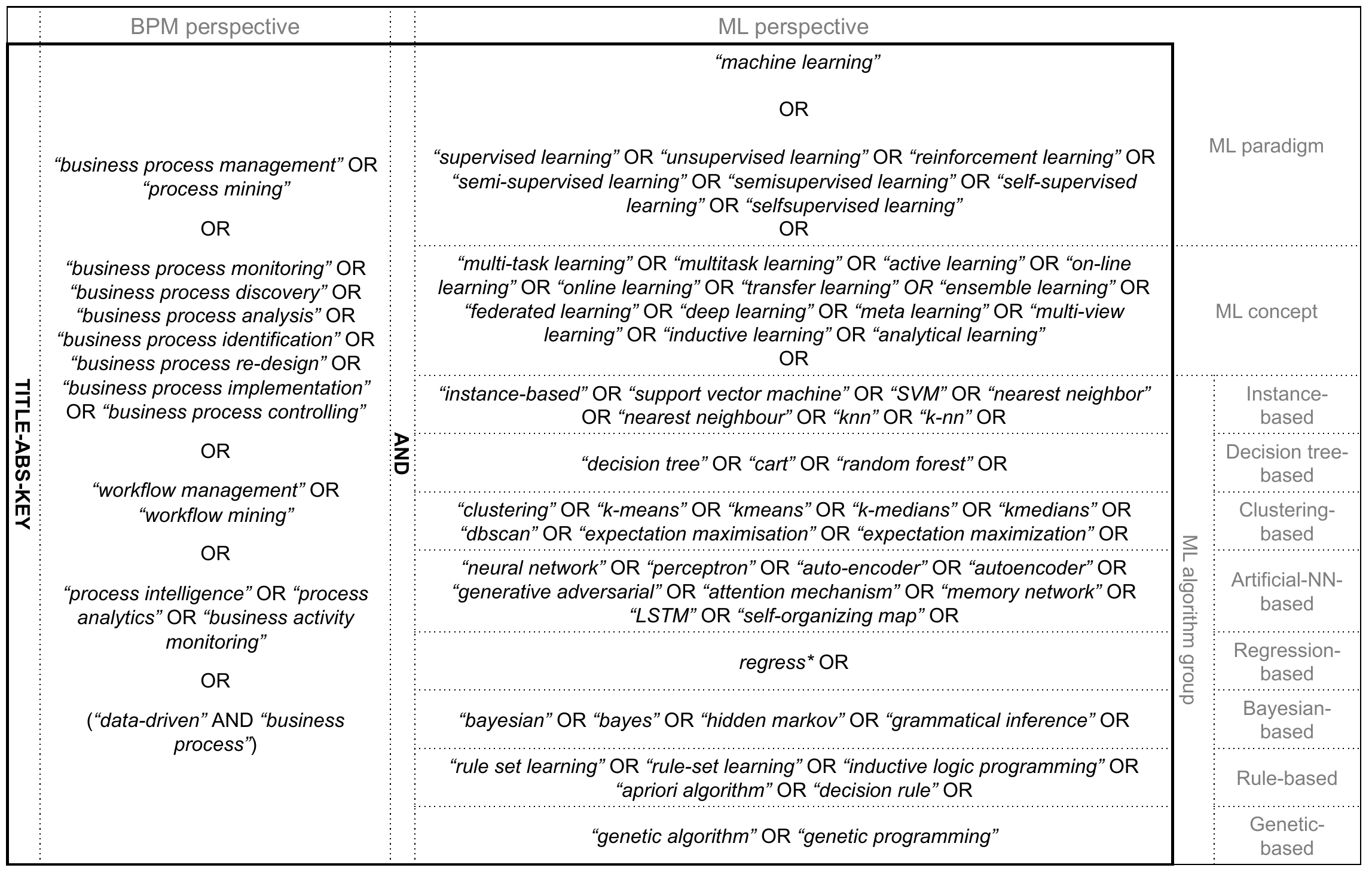}
 \caption{Search string (in Scopus syntax)}
 \label{fig:searchstring}
\end{figure}

We chose a systematic, sequential procedure to find a representative set of papers for our review~\citep{vom2015standing, cooper1988organizing}. We used the databases \emph{Scopus}, \emph{IEEE Xplore}, and \emph{Web of Science} to conduct the search process, as the combination of these databases covers a wide range of academic papers in the \gls{bpm} domain. 
We defined the keywords for our search string based on our addressed research gap and the dimensions described in Section~\ref{subsec:-bpm} and Section~\ref{subsec:-ML} combining \gls{bpm}-related keywords with \gls{ml}-related keywords \citep{vom2015standing}. The search string is depicted in Figure~\ref{fig:searchstring} (see~\ref{sec:searchprocessspecifications} for a detailed description of the creation of the search string).

\begin{figure}[ht]
 \centering
 \includegraphics[width=0.9\textwidth]{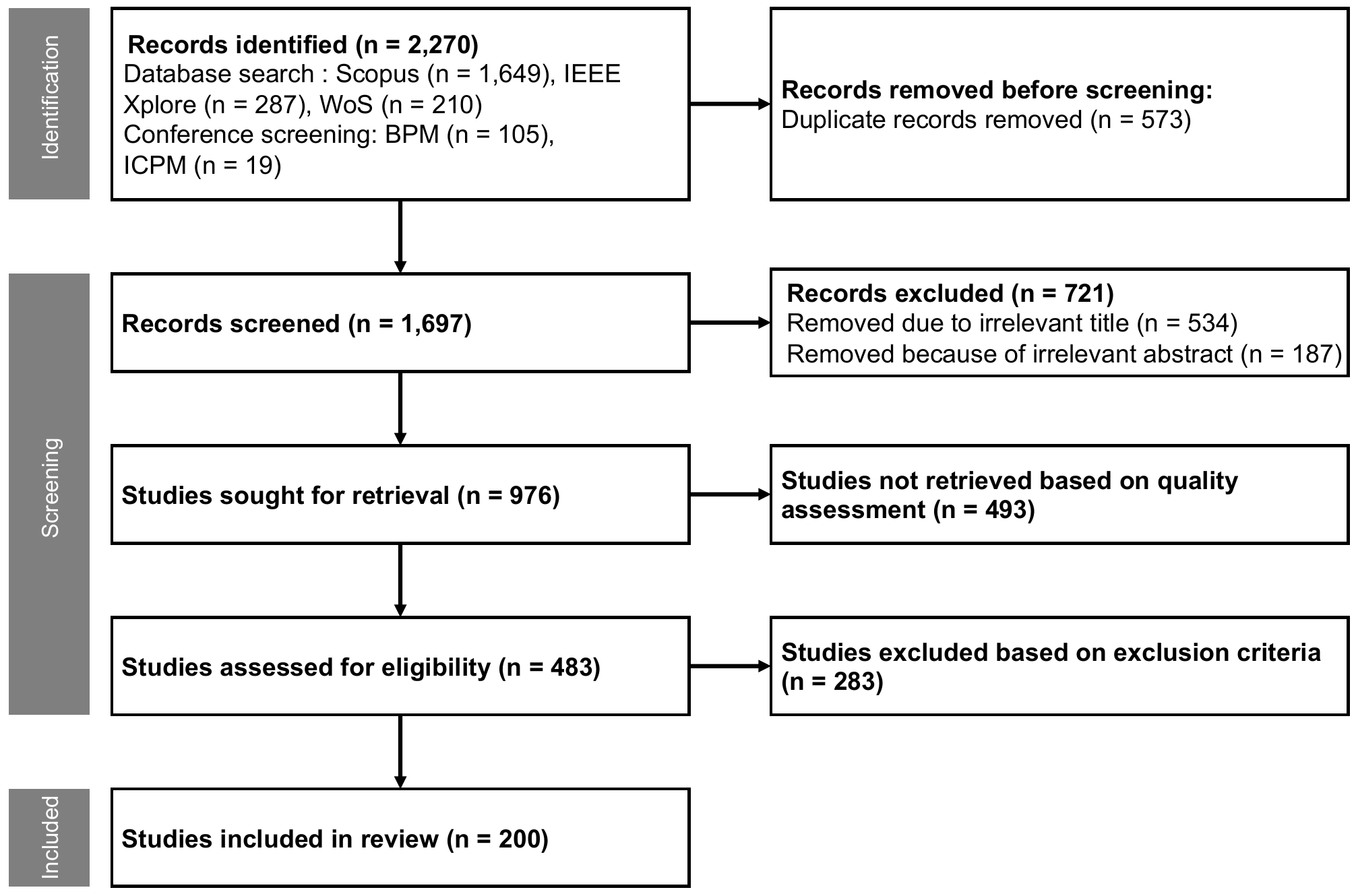}
 \caption{Phases of the search process following the PRISMA framework}
 \label{fig:searchprocess}
\end{figure}

We also screened the main proceedings of topical conferences (i.e. the \emph{International Conference on Business Process Management (BPM)} and the \emph{International Conference on Process Mining (ICPM)}). Our search and screening, executed in July 2022, retrieved 2,270 papers. 
Next, we proceeded in a structured manner to determine the final set of relevant papers. The search process is depicted in Figure \ref{fig:searchprocess}.

After removing duplicates, we scanned the title and abstract of each paper for relevance to our topic and checked the papers that remained for whether they met our quality criteria~\citep{okoli2015guide}, for example, whether they meet a length requirement of at least four pages.
Lastly, to assess eligibility, we screened the full text of each remaining paper in terms of our defined exclusion criteria~\citep{vom2015standing}. The quality and exclusion criteria can be found in~\ref{sec:searchprocessspecifications}.
At the end of this process, we had 200 papers. 

\subsection{Coding Process} 
We followed \citet{hruschka2004reliability} adjusted to three coders by first developing an initial coding scheme,  coding the entire sample second, and, third, reconciling the coders' results. 
In the first step, we adapted \citeauthor{Recker2016}'s (\citeyear{Recker2016}) coding scheme and \citeauthor{janiesch2021machine}'s (\citeyear{janiesch2021machine}) \gls{ml}-model development process to set up the \gls{bpm}-related and \gls{ml}-related dimensions of our coding matrix~\citep{webster2002analyzing}. For the \gls{ml}-related dimensions, we used the categories as defined in the search string (see Appendix~\ref{sec:searchprocessspecifications}).
Two authors conducted two coding iterations of ten randomly sampled papers per iteration per person. After each iteration, the authors reflected on the coding scheme's sufficiency and completeness and added or adapted dimensions and concepts to better fit the review's objectives~\citep{webster2002analyzing}. The coders also defined each concept textually to ensure a common understanding, which finalised the coding scheme. 

In the second step, three authors conducted the final coding, analysing and coding each of the 200 papers in the final set. The papers were distributed equally among the coders to deal with the large number of papers.
To clarify each paper's positioning along the \gls{bpm} lifecycle and its goal, the paper's \gls{bpm} aim, such as predicting the next activity or predicting remaining time, was extracted as free text. After all papers were coded, one researcher clustered the tasks (e.g. predictive business process monitoring) per lifecycle phase~\citep{webster2002analyzing}.

In the third step, the different codes were reconciled in a single table. Since there were multiple coders involved, we conducted an inter-coder reliability analysis using 25 randomly selected papers to ensure consistency among the coders. 
We calculated the percentage agreement, Krippendorff's $\alpha$ \citep{krippendorff2018content}, and Fleiss' $\kappa$ \citep{fleiss1973equivalence}, which are established metrics for three or more coders~\citep{lombard2002}, and found high inter-coder reliability. Further information on the analysis and the results can be found in Section~\ref{sec:intercoderReliability}. 
The coders pointed out any ambiguities in the coding table during the coding process, which could be resolved later in open discussions to ensure uniformity.

For greater transparency, our coding table can be found in Section \ref{sec:codingtable} and in an interactive concept matrix.

\subsection{Interactive Concept Matrix}
\begin{figure}[ht]
    \centering
    \includegraphics[width=0.6\textwidth]{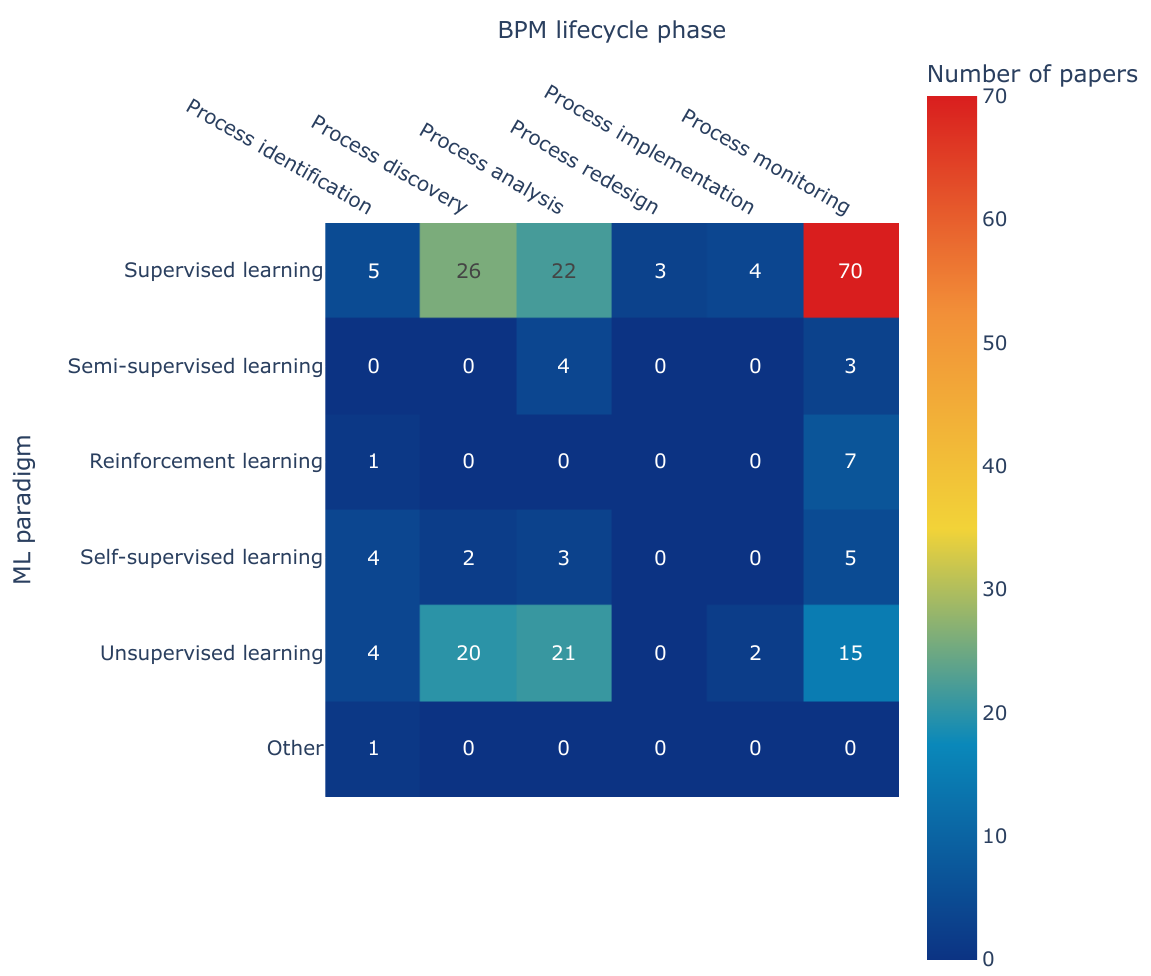}
    \caption{Exemplary figure of the interactive concept matrix for the intersection of the \gls{bpm} lifecycle phases and the \gls{ml} paradigms}
    \label{fig:intersection}
\end{figure}

One aim of our systematic literature review is to make the data coded from the collected papers transparent and accessible in a useful manner. While the coding table in Section \ref{sec:codingtable} fulfils the first aspect, its usability is limited because of its size (number of papers coded and the number of coding categories).
Therefore, we created an interactive concept matrix (including a filterable table and two exemplary visualisations of the results)\footnote{The interactive concept matrix can be found here: \href{https://literaturedashboardbpm.herokuapp.com/}{https://literaturedashboardbpm.herokuapp.com/} After publication, we aim to provide the content openly available via the university's website.} that synthesises the concepts the literature considers from a descriptive perspective. The interactive concept matrix is structured in three sheets, each allowing the user to filter certain concepts to gain insights. On the first sheet \textquote{Concept development over time}, the user sees how the concepts the literature considers evolve, for example, per \gls{bpm}~lifecycle~phase. Using the second sheet, \textquote{Intersection of concepts}, the user can gain insights into how two concepts intersect. One example can be the intersection of the \gls{bpm} lifecycle phases and the \gls{ml} paradigms, where the heatmap colour indicates the number of papers per intersection category, as shown in Figure~\ref{fig:intersection}.

From the third sheet, \textquote{Complete coding table}, the user can glean a holistic perspective of all concepts the literature considers and filter the complete concept matrix for certain values, such as specific authors, publication years, or lifecycle phases.

\section{Results}
\label{sec:results}

\begin{figure}[ht]
    \centering
    \includegraphics[width=0.9\textwidth]{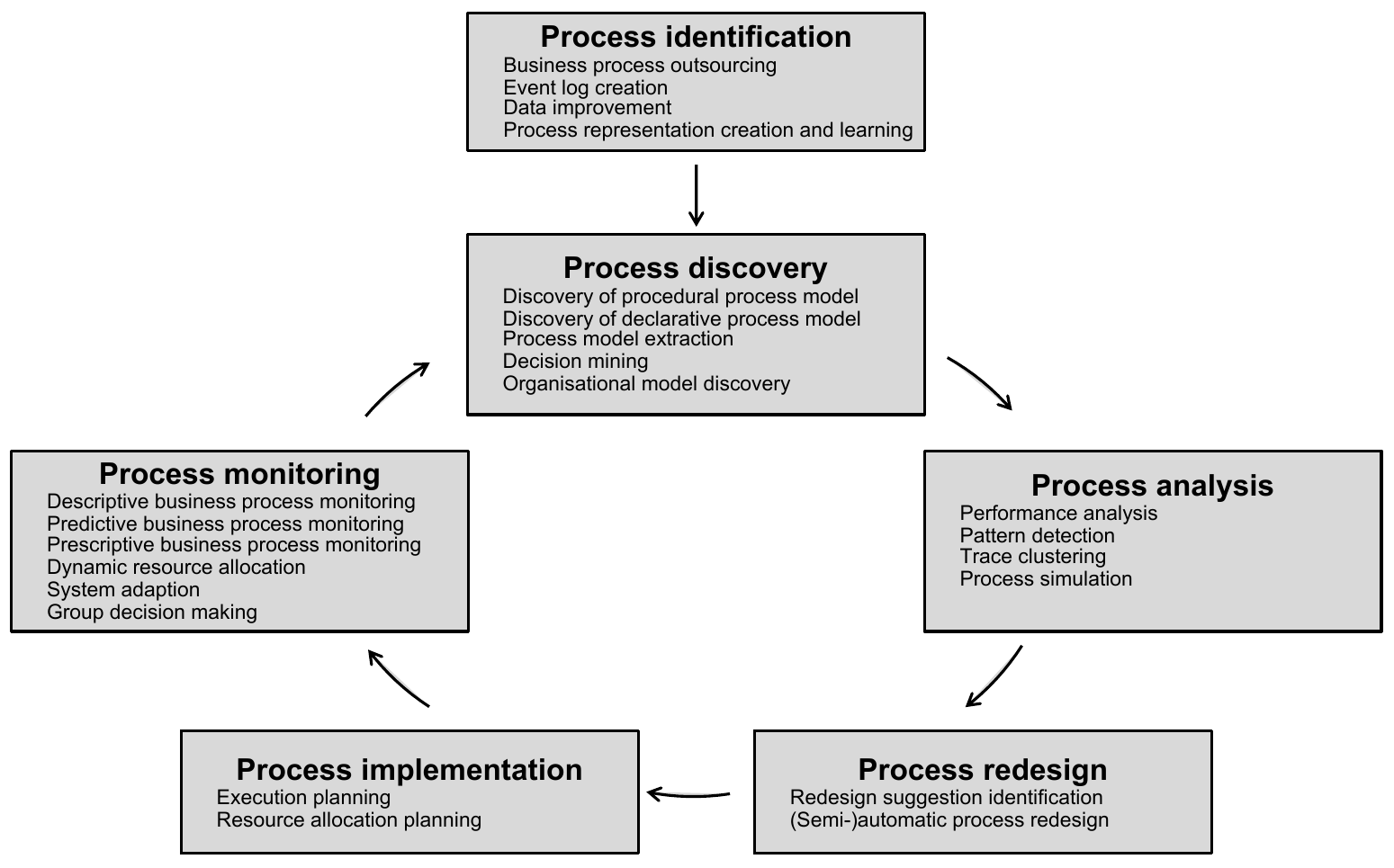}
    \caption{\gls{bpm} lifecycle of \protect\citet{dumas2018fundamentals} enriched with the identified tasks}
    \label{fig:bpmlifecycle_inclprobs}
\end{figure}

This section presents an overview of \gls{ml} applications in \gls{bpm} along the six phases of the \gls{bpm} lifecycle and the \gls{bpm} tasks derived from our literature review. Figure~\ref{fig:bpmlifecycle_inclprobs} shows the \gls{bpm} lifecycle by \citet{dumas2018fundamentals}, enriched with tasks we identified. 
In addition, descriptions of \gls{bpm} lifecycle phases and definitions of \gls{bpm} tasks in the view \gls{ml} applications are provided.


\subsection{Process Identification}

\begin{figure}[ht]
    \centering
    \includegraphics[scale = 0.5]{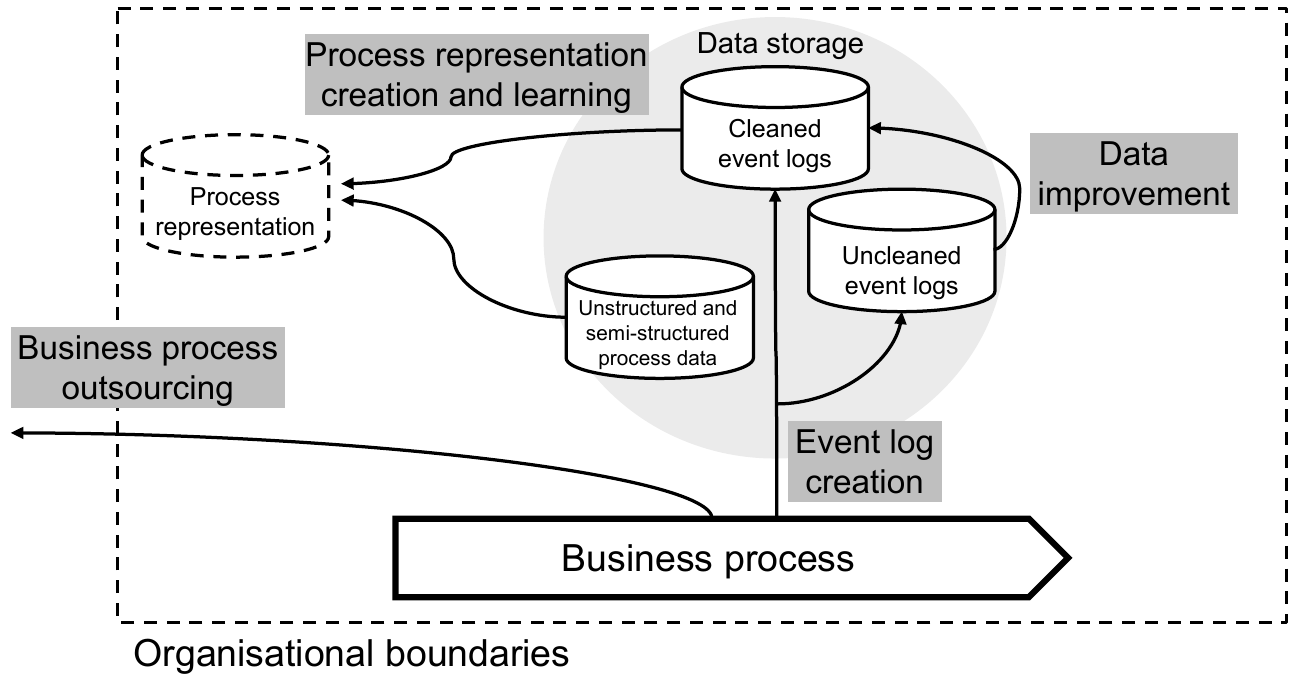}
    \caption{Schematic representation of the tasks in the process identification phase}
    \label{fig:ProcessIdentification_tasks}
\end{figure}

In the process identification phase, organisations collect information about their business processes. Tasks in this phase are identifying processes in the organisational landscape, determining which business processes can be optimised and whether an optimisation should happen within or outside the organisation. \Gls{ml} applications can support different tasks in this phase, especially by unlocking unstructured data, repairing missing data, and supporting outsourcing decision-making. Moreover, creating data representations facilitates thorough analyses in later phases of the \gls{bpm} lifecycle. Figure~\ref{fig:ProcessIdentification_tasks} summarises the tasks in the process identification phase.

\subsubsection{Business Process Outsourcing}
A strategic task of process identification is to decide whether a business process or a part of a business process can be optimised regarding time or cost when outsourced to another organisation. \Gls{ml} applications can help identify process indicators that propose a high probability of successful outsourcing. Therefore, \citet{ciasullo2018business} propose a reinforcement-learning-based framework that supports outsourcing decisions of business processes.

\subsubsection{Event Log Creation}
After adding process mining to many organisations' business intelligence portfolios, event logs gained importance quickly. Event logs are the standard data form for process-mining applications. While structured data in information systems can directly be compiled into an event log, \gls{ml} applications can also turn raw data (e.g. video, image, or plain text) into event logs.
Therefore, \citet{sim2022automatic} propose a method to create an event log from raw event data using \gls{cnn} models.
\citet{tello2019machine} provide a framework that uses supervised-learning models for mapping low-level event data onto high-level process activities.  
An event log can also be created for further analysis by detecting process instances (i.e. sequences of events) in e-mails with a clustering model~\citep{jlailaty2017business} or processing video data with a computer-vision-based approach \citep{kratsch2022shedding}.

\subsubsection{Data Improvement}
As part of process identification, process analysts identify process data in sources and sinks. Process data is often represented by event logs that hold a potential value but can lack quality when they include missing or incorrect data \citep{bose2013wanna}. 
\gls{ml} applications can reconstruct missing values in event data or identify and replace incorrect event data. 
As \gls{ml} applications for data improvement increase event-log quality, they leverage accurate analyses.  
Missing activities in event log data are repaired using a self-organising map model for trace clustering \citep{xu2019profile}. 
\citet{nolle2020deepalign} correct anomalies using \gls{gru} models with bi-directional beam search.
\citet{nguyen2019autoencoders} clean event log data using an autoencoder model.

\subsubsection{Process Representation Creation and Learning}
Process identification also deals with finding compact, abstract, and computational process representations on which \gls{ml} can build applications for addressing \gls{bpm} tasks in subsequent phases of the \gls{bpm} lifecycle. Therefore, \gls{bpm} research has developed \gls{ml} applications to create and learn process representations from process data.

Supervised-learning approaches can produce dense and accurate process representations~\citep{seeliger2021learning} and event abstractions~\citep{tax2016event}.
\citet{koninck2018act2vec} propose self-supervised learning techniques to create representations of activities, traces, event logs, and process models. \citet{guzzo2021multi} also use self-supervised learning to create process representations considering multi-dimensional aspects of event log traces.

\subsection{Process Discovery}
\begin{figure}[ht]
    \centering
    \includegraphics[scale = 0.5]{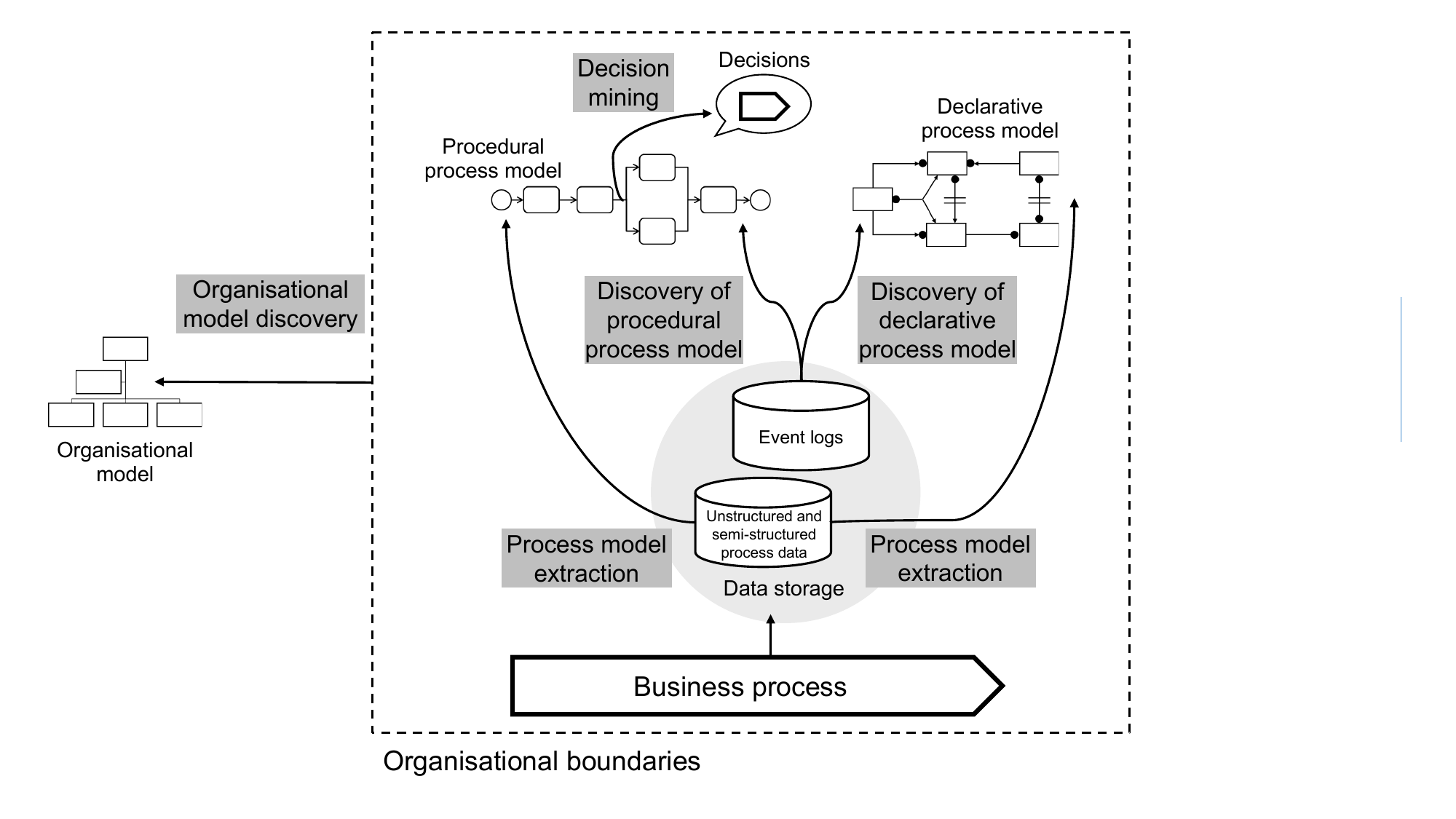}
    \caption{Schematic representation of the tasks in the process discovery phase}
    \label{fig:ProcessDiscovery_tasks}
\end{figure}

Process discovery focuses on generating as-is models that describe a process as it is executed according to event data. Process models can become complex depending on the number of variants and the length of traces in a process. Unsupervised learning applications can focus on process variants to improve process model comprehension. While \gls{ml} applications can detect general process model structures in data, they can also mine specific types of process models, such as procedural and declarative models. Moreover, \gls{ml} applications can optimise discovered process models regarding specific process model quality criteria. Consequently, \gls{ml} applications support many tasks in the process discovery phase and automate it to a certain extent. Figure~\ref{fig:ProcessDiscovery_tasks} overviews the process discovery tasks.

\subsubsection{Discovery of Procedural Process Model}
Discovering procedural process models is a task to gain transparency over a process as it is executed. 
Discovered procedural process models can be complex, so they are often simplified using heuristics or filtering options. 
\gls{ml} applications can identify process structures or optimise parameters and event-log sampling for discovery algorithms to create accurate and comprehensible procedural process models. Additionally, \gls{ml} applications can choose the best-suited discovery algorithm based on the input data and produce specific procedural models (e.g. reference models). 
%
Early works in \gls{ml}-based \gls{bpm}, such as~\citet{herbst2000machine} and \citet{herbst2004workflow, herbst2000integrating}, focus on the discovery of process model structures using Bayesian algorithms. 

Other studies address trace-clustering-based process model discovery.  
\citet{greco2004mining} propose an algorithm that incorporates a k-means model to cluster workflow executions, where a cluster is a global constraint of a discovered process. In a later work \citep{greco2006discovering}, the same authors present an extension of their algorithm for discovering conforming process models. \citet{garcia2014controlled} propose a trace-clustering-based process discovery technique that allows users to control process model complexity. Closely connected to these studies, \citet{qiao2011towards} present a clustering-based approach for retrieving business process models.

Process discovery applications that rely on genetic algorithms use target metrics like fitness \citep{van2005genetic}, either fitness, replay, precision, generalisation, or simplicity \citep{buijs2012genetic}, or completeness, precision, and simplicity together~\citep{vazquez2014genetic, vazquez2015prodigen}, to create optimised process models. Genetic algorithms can also be combined with graph-based representations to analyse complex processes~\citep{turner2008genetic}. 

Other studies that consider discovery of procedural process model address various discovery tasks, such as the creation of process models for unseen event log data using \gls{gnn} models~\citep{sommers2021process}, the creation of process models using hidden Markov models containing equations and rules~\citep{sarno2016hidden}, and, building on the latter, the inclusion of invisible prime tasks and parallel control-flow patterns~\citep{sarno2016coupled}. 

Moreover, probabilistic process models using a Bayesian algorithm~\citep{silva2005probabilistic} are discovered.
\citet{lu2016synchronization} discover synchronisation-core-based process models, where a decision-tree model is used to group traces into clusters, from which synchronisation cores of activity dependencies are derived. 
\citet{ferreira2009discovering} discover process models from event stream data by estimating model parameters and case assignments using an expectation-maximisation procedure.  
\citet{ferreira2013mining} discover hierarchical process models that capture the relationships between macro-level activities and micro-level events using a hierarchical Markov model with expectation-maximisation-algorithm-fitted parameters.

Another group of studies discovers certain forms of process models, such as process trees using a k-means model to handle context data in addition to control-flow data~\citep{shraga2020process} or connections between events using a logistic regression model~\citep{maruster2002process}. Building on the latter, \citet{muarucster2006rule} propose a method that uses the RIPPER algorithm to create rule sets representing process-activity relations. 

Finally, specific procedural process discovery applications are meta-process discovery for process discovery algorithm selection using an \gls{svm} model for reference model selection~\citep{wang2012efficient}, reference model discovery using a genetic algorithm~\citep{martens2014genetic} or a clustering algorithm \citep{li2010minadept}, behavioural pattern model discovery using a clustering algorithm~\citep{diamantini2016behavioral}, and discovery of configurable process model using a genetic algorithm~\citep{buijs2013mining}.

\subsubsection{Discovery of Declarative Process Model}
In contrast to procedural process models, declarative models express specific rules, which should not be infringed during execution. 
The discovery of declarative process models finds constraints in an event log. \gls{ml} applications support both discovering declarative models and consistency checking of such constraints.
%
\citet{leno2018correlating} facilitate the discovery of multi-perspective declarative process models using a clustering model for selecting traces and a classification model to identify declarative constraints. \citet{leno2020automated} additionally present a redescription-mining-based approach for declarative constraint identification that uses two decision-tree models. %
Inductive logic programming allows for the discovery of declarative models~\citep{lamma2007inducing,chesani2009exploiting} and learning integrity constraints for declarative modelling~\citep{lamma2007applying}.
\citet{maggi2012efficient} use the Apriori algorithm to find declarative constraints, and \citet{maggi2018parallel} combine the Apriori algorithm with a sequence analysis algorithm to discover declarative constraints.

\subsubsection{Process Model Extraction}
Process models can be modelled in diagram software or paper drawings, or described in plain text. Interpreting these process models and translating them into computationally interpretable process models can be done with \gls{ml} applications.
Plain-text process descriptions can serve as foundations for extracting procedural process models using a hierarchical multi-grained \gls{dnn}~\citep{qian2020approach} and declarative process models using a discovery algorithm that incorporates a BERT model~\citep{lopez2021declarative}.
\citet{polanvcivc2020empirical} extract digital process models from hand-drawn process diagrams using a \gls{cnn} model. \citet{kim2002document} propose a document-based discovery approach that applies case-based reasoning to effectively reuse the design outputs.

\subsubsection{Decision Mining}
When discovering processes, it is essential to understand decision points where process flows split. Decision mining comprises identifying decision points and conditions, which lead to a course of action. \gls{ml} applications can automate identifying decision points and decision dependencies. 
Process models can be enhanced by using decision-tree models to detect dependencies in the data at the decision points in the process model, as \citet{rozinat2006decision} do in one of the first approaches. \citet{bazhenova2016discovering} use decision trees to semi-automatically derive decision models from event logs. 
Both \citet{mannhardt2016decision} and \citet{effendi2017discovering} present a technique for discovering overlapping rules in event logs using decision-tree models. 
\Citet{leoni2013discovering} propose a technique for discovering branching conditions by combining decision-tree models with invariant discovery techniques. 
Lastly, \citet{sarno2013decision} propose a decision-mining approach that uses a decision-tree model and considers multi-choice workflow patterns.

\subsubsection{Organisational Model Discovery}
Processes are not executed in isolation but in the context of an organisation. Consequently, event data contains information about organisational functions. \gls{ml} applications can create models of the executing organisation based on event logs using clustering models~\citep{yang2018finding}.

\subsection{Process Analysis}
\begin{figure}[ht]
    \centering
    \includegraphics[scale = 0.5]{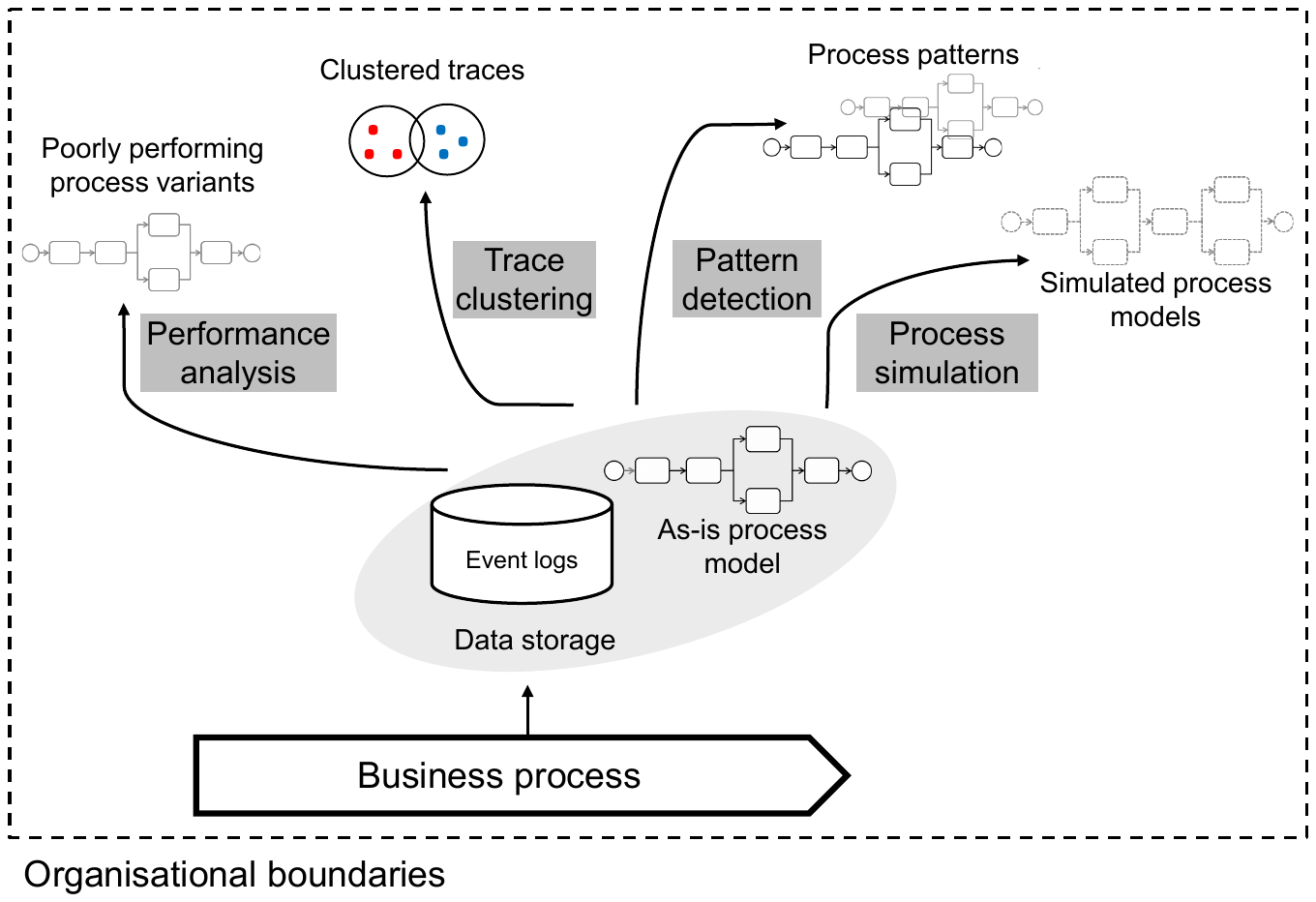}
    \caption{Schematic representation of the tasks in the process analysis phase}
    \label{fig:ProcessAnalysis_tasks}
\end{figure}

Process analysis assesses as-is processes using process models and process data. Process analysis takes a retrospective perspective in analysing a process. Process analysts determine \glspl{kpi}, assess process performance, look for problematic instances, and simulate processes to evaluate process redesigns.
All these tasks can be supported with \gls{ml} applications' pattern recognition capabilities, unsupervised learning capabilities for identifying similar process instances regarding control flow or performance, and predictive capabilities to determine realistic simulation parameters. Figure~\ref{fig:ProcessAnalysis_tasks} shows a schematic representation of the tasks in the process analysis phase.

\subsubsection{Performance Analysis}
When assessing the performance of processes and process instances, organisations determine \glspl{kpi} to look for poorly performing processes and the cause of their performance failure. In doing this, analysts find starting points for process redesign. \gls{ml} applications support performance analysis, for example, by repairing scarce performance data, aggregating \glspl{kpi}, and identifying causal relations between data and process outcome. 
\citet{Berkenstadt2020} propose an approach using techniques from queuing theory and supervised learning to predict process performance indicators. 
%
\citet{wetzstein2009monitoring} use rule-based learning to recompose low-level process indicators into high-level process metrics. 
To enable fairness-aware process mining and performance analysis, \citet{qafari2019fairness} add situation-specific discrimination to event logs and train decision-tree models. \citet{Es-Soufi2016} propose an approach that detects process patterns using process mining and creates performance-related predictions per activity of a process pattern using a supervised-learning model.

Training logical decision-tree models based on logical event log encoding~\citep{ferreira2015using} and uplift tree models~\citep{bozorgi2020process} can reveal causal relationships in process performance and specific control-flow actions. 
\Citet{savickas2018belief} use belief network models to analyse process performance with domain-specific data.
In contrast to the previously mentioned approaches, \citet{theis2020adversarial} compute the generalisability of process models using \gls{gan} models.
To support performance analyses in general, \citet{leoni2014general} and \citet{leoni2016general} present a process analysis framework to correlate different process characteristics using regression and decision-tree models.

\subsubsection{Pattern Detection}
Process analysis further strives to identify favourable and problematic process patterns. Before pattern recognition capabilities of \gls{ml} applications could be leveraged, process analysts manually compared process variants to find workarounds, anomalous behaviour or time series patterns, which could be improved or transferred to other instances. 
Process analysts can thereby provide input for redesigning process variants. 
Analysts can use \gls{ml} applications with autoencoder models to identify anomalies even when they lack domain knowledge, as~\citet{nolle.2018} do without, and \citet{krajsic2020lambda} do within, a lambda architecture designed for event steams.  
Another approach uses a word2vec model to encode activities in event log data and a (one-class) \gls{svm} model to detect anomalies~\citep{junior2020anomaly}.
\citet{rogge2014temporal} detect temporal anomalies in the activity execution duration using Bayesian models. 

\citet{cuzzocrea2015multi,cuzzocrea2016multi} detect deviant process instances using multi-view, multi-dimensional ensemble learning in event logs.
Building on their techniques, \citet{cuzzocrea2016robust} present a probabilistic-based framework for robust detection of deviance. 
\citet{stierle2021technique} derive activity relevance-scores from an attention-layer of a \gls{gnn} model that detects deviant process instances.    

\citet{weinzierl2022detecting} present a method for detecting workarounds using an autoencoder model that removes noisy process instances and a \gls{cnn} model that maps process instances to workaround classes.
\citet{yeshchenko2019comprehensive} detect concept drifts in event log data using visual analytics and time-series-based clustering.
Finally, supervised learning and conformance checking \citep{valdes2022conformance} or supervised learning and graph kernels~\citep{valdes2022graphkernel} can prove useful in identifying time-series patterns in event log data.

\subsubsection{Trace Clustering}
Clustering approaches can automatically group process instances or instance variants to analyse processes quickly. Improvements, problems, and advantages identified in one variant may apply to other instances or instance variants in the same cluster. Therefore, clustering traces can support process analysts.  

The most popular trace-clustering approach in \gls{bpm} is distance-based trace clustering. Distance-graph models~\citep{ha2016trace}, graph similarity metrics~\citep{weerdt2012leveraging}, and similarity in heterogeneous information networks~\citep{nguyen2016process} allow for distance-based trace clustering. 
\citet{delias2019non} take the outranking relations theory into account for trace clustering. 
Other studies address trace clustering using a co-training-based strategy with multiple trace profiles~\citep{appice2015co}, frequent-item-set mining~\citep{seeliger2018finding}, and the Levenshtein distance~\citep{bose2009context} to consider process-context data.
High-level process analysis is achieved by employing hierarchical clustering \citep{jung2008hierarchical,jung2009hierarchical} or clustering using graph partitioning \citep{sarno2013clustering} based on graph similarity of different business process models. 

Previous research also uses model-based trace clustering and approaches that include expert and domain knowledge via process model metrics~\citep{koninck2017approach}, the extension with must-link and cannot-link relationships between process instances \citep{koninck2021expert}, and active learning~\citep{weerdt2013active}. 
Super-instance-based trace clustering enables combining distance- and model-based clustering~\citep{koninck2019scalable}. \citet{boltenhagen2019generalized} use distance-based quality criteria in model-based trace clustering.

Prior research also describes certain applications of trace clustering. 
\citet{folino2015mining} use a logical trace-clustering model to create comprehensible process models. 
\citet{varela2019process} apply trace clustering to mine configuration flows optimised for a particular type of user. 
\citet{yang2017data} use distance-based clustering to set up a recommender system. 
\citet{ferreira2007approaching} apply a mixture of Markov chains learned with the expectation-maximisation algorithm to create traces from identified tasks.
\citet{koninck2017explaining} present a trace clustering approach with instance-level explanations. 

\subsubsection{Process Simulation}
Process analysis assesses potential process changes by simulating redesigned process models. \gls{ml} applications can aid in creating realistic simulation conditions, as the conditions are learned from a mass of data instead of hard-coded rules. 
While \citet{camargo2022learning} facilitate data-driven process simulation using two \gls{lstm} models, one for processing time prediction and one for waiting time prediction, \citet{khodyrev2014discrete} use decision and regression tree models to predict \gls{kpi} values for short-term process simulation.

\subsection{Process Redesign}
\begin{figure}[ht]
    \centering
    \includegraphics[scale = 0.5]{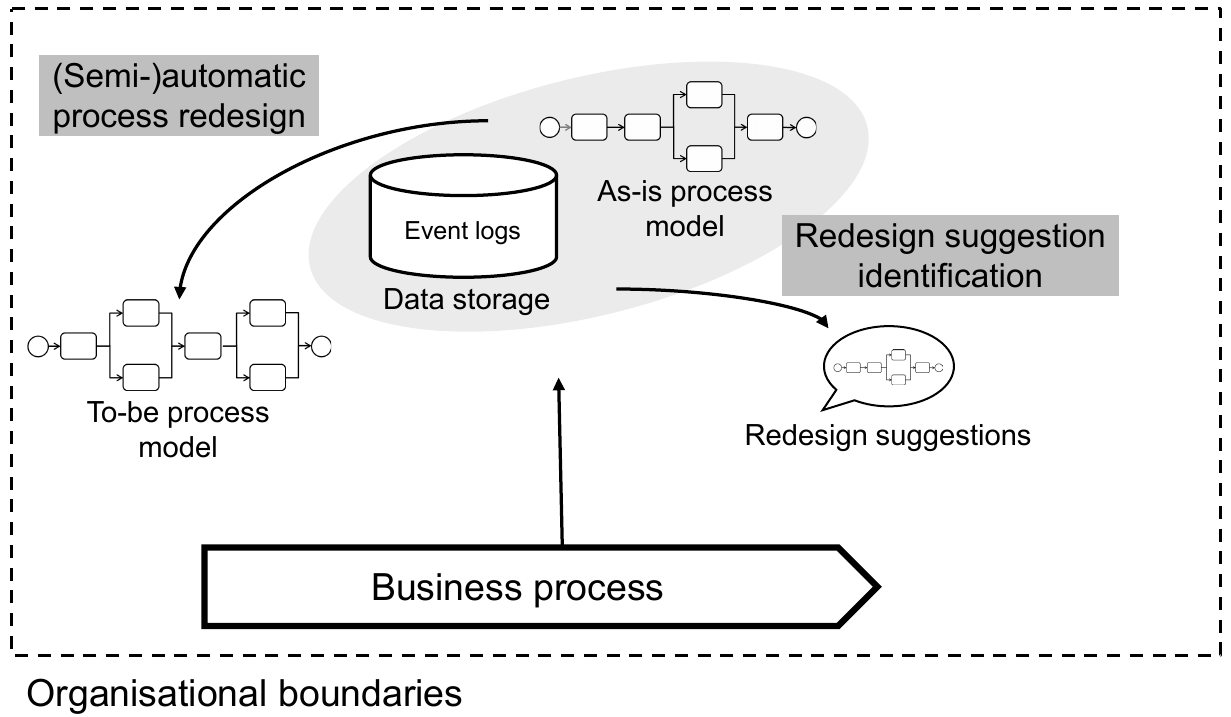}
    \caption{Schematic representation of the tasks in the process redesign phase}
    \label{fig:ProcessRedesign_tasks}
\end{figure}

Process redesign's goal is to improve an as-is process by applying process changes to the model and creating a to-be process model. Process designers use, for example, experience-based redesign heuristics, describing concrete measures to make a process efficient or less prone to error.   
\Gls{ml} applications can support process redesigners by automatically suggesting process redesign measures based on data or by performing (semi-)automatic process redesign, where more appropriate or optimised process designs are identified. Figure~\ref{fig:ProcessRedesign_tasks} provides an overview of these tasks.

\subsubsection{Redesign Suggestion Identification}
A task in process redesign is to elaborate redesign suggestions based on ideas, heuristics, or new technology. \Gls{ml} applications can identify unfavourable patterns in process instances and how to redesign them, for example, geared towards a certain target (e.g. customer satisfaction or throughput time).  
Therefore, research in this area addresses redesign suggestion identification. 
\citet{Mustansir2022} use language models and \gls{dnn} models to detect suggestions for process redesign in textual customer feedback data. 

\subsubsection{(Semi-)automatic Process Redesign}
A subset of \gls{ml} applications in process redesign focuses on semi-automatic process redesign. 
More concretely, \gls{ml} applications addressing this task use genetic algorithms to perform multi-objective optimisation of business processes \citep{vergidis2007optimisation} or to determine an objective prioritisation of process redesigns~\citep{afflerbach2017design}.

\subsection{Process Implementation}
\begin{figure}[ht]
    \centering
    \includegraphics[scale = 0.5]{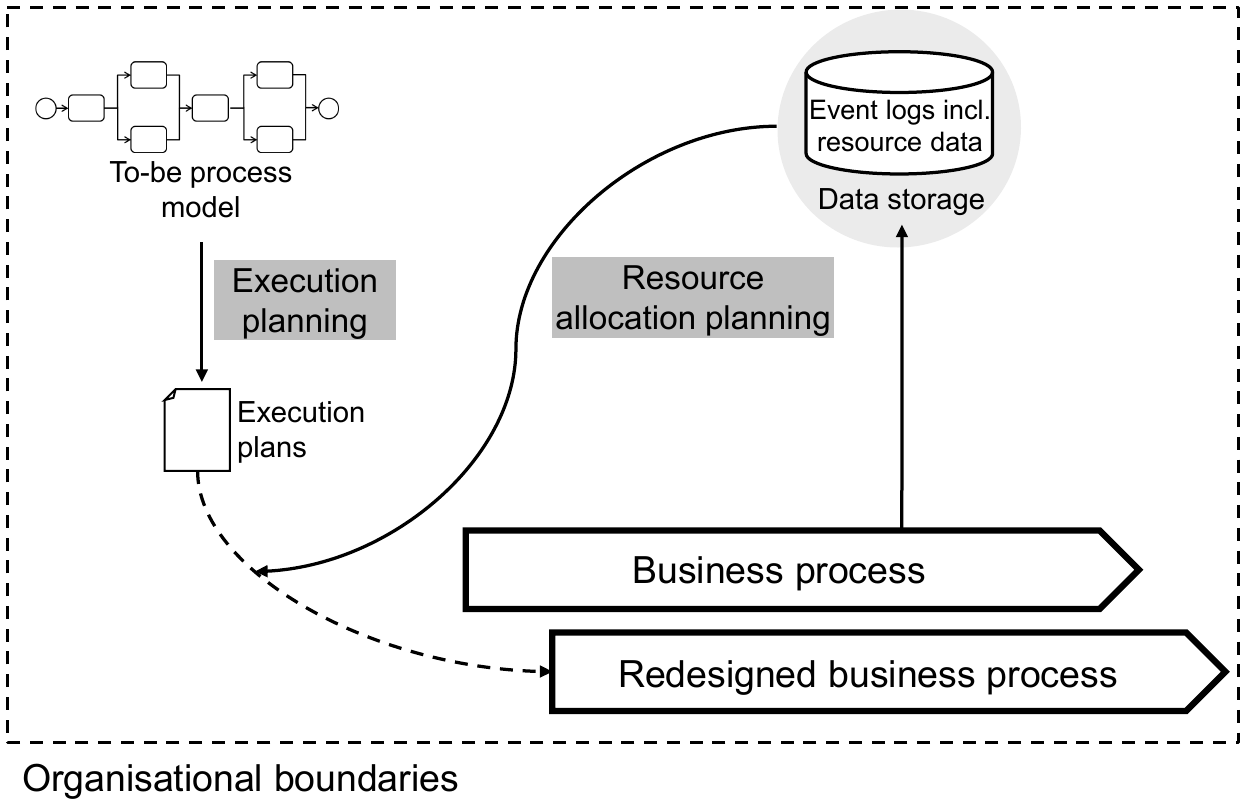}
    \caption{Schematic representation of the tasks in the process implementation phase}
    \label{fig:ProcessImplementation_tasks}
\end{figure}

Redesigned process models must be integrated into an information system before they can be executed. Resources need to be assigned to redesigned tasks, staff needs to be trained, and execution plans need to be rolled out. \Gls{ml}-supported process implementation tasks include creating execution and staffing plans from process models and data and resource allocation planning. Figure~\ref{fig:ProcessImplementation_tasks} shows the tasks of the process implementation phase.

\subsubsection{Execution Planning}
Transferring from a to-be process model to process implementation requires setting up execution plans. Process models, including decision points and process semantics, can be used to create execution plans. This is tedious work, but \gls{ml} applications can approximate an optimal execution plan from a detailed model.
Therefore, \citet{bae2014planning} propose a method to enrich \gls{bpm} structures with alternative process paths and semantics, after which a genetic-based algorithm can find an execution plan that enhances performance. 

\subsubsection{Resource Allocation Planning}
After having set execution plans, resources, such as machines, employees, storage spaces or production lines, must be assigned to cases. Process owners define who or what is employed in which process variant. To make decisions during planning, \gls{ml} applications can provide suggestions that optimise resource utilisation or throughput time. Process owners can then pick a suggested resource allocation plan and adjust it if needed. \gls{ml} applications thereby eliminate trial and error approaches.  
Resource allocation planning is realised based on the prediction of execution routes using a na\"ive Bayes model~\citep{kazakov2018development}, the resource decision mining from events using classification models and heuristics~\citep{senderovich2014mining}, the examination of process instance scheduling using a genetic-based algorithm~\citep{xu2016resource}, or the recommendation of resource allocation candidates using classification models~\citep{liu2012mining}.
Finally, \citet{delcoucq2022resource} apply a bi-dimensional clustering approach for resource allocation planning, where resources are clustered based on similar behaviour in the process, and activities are clustered based on the executing resource.

\subsection{Process Monitoring} 
\begin{figure}[ht]
    \centering
    \includegraphics[scale = 0.5]{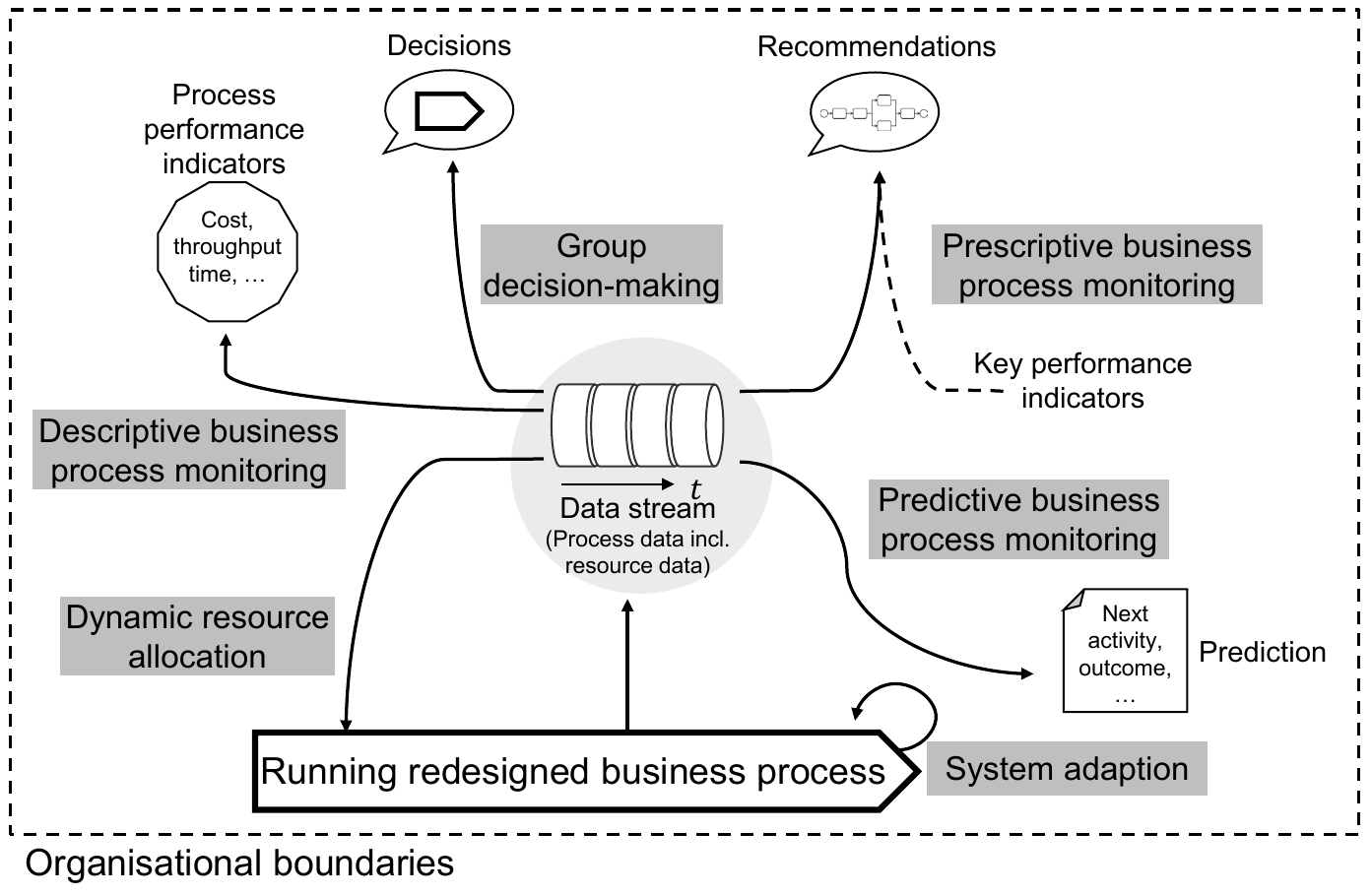}
    \caption{Schematic representation of the tasks in the process monitoring phase}
    \label{fig:ProcessMonitoring_tasks}
\end{figure}

After the redesigned processes are implemented, they are executed, and how well they perform in terms of their performance and objectives must be determined. Process monitoring challenges analysts because parallel cases must be overseen. Additionally, cases, which look fine momentarily, can turn out problematic later on. \gls{ml} applications can recognise patterns in event data that indicate problematic outcomes, even when a case is in a good state. \gls{ml} applications can help analysts identify cases for further analysis in large process data sets. The overarching \gls{ml} applications' role in process monitoring consists of putting an analyst's attention on particular cases and predicting process behaviour learned from historical data. Executing business processes can be optimised regarding \glspl{kpi} or process conformance. \Gls{ml} applications in process monitoring can further automate noticing an appropriate point for starting another \gls{bpm} lifecycle iteration by detecting concept drifts in processes. Figure~\ref{fig:ProcessMonitoring_tasks} provides a schematic overview of the tasks and their relationship.

\subsubsection{Descriptive Business Process Monitoring}
Descriptive business process monitoring determines \glspl{kpi} of running process instances and provides real-time information about reactive actions that could increase process performance. It can also detect concept drifts in processes and identify anomalous behaviour in process instances. \gls{ml} applications can recognise patterns in high-volume data. In doing that, they help to find entry points for reiterating the \gls{bpm} lifecycle to adapt to concept drifts or positive anomalies, or to counteract poor process behaviour. 

\citet{montani2012retrieval} retrieve traces via a k-nearest-neighbour model and cluster them using a hierarchical clustering model to detect changes in event log data. 

Online detection of anomalies is addressed using graph autoencoder models~\citep{huo2021graph} or supervised learning, specifically, a random forest model, extreme gradient-boosting model, or an \gls{lstm} model~\citep{lee2022analysis}. 
Other studies apply a DBSCAN model to cluster process instances into common and anomalous behaviour~\citep{barbon2018framework}, or a DenStream model \citep{cao2006density} to cluster process instances with a graph-based representation including control flow and time~\citep{tavares2019overlapping} and additional attributes~\citep{ceravolo2022real}. 
\citet{krajsic2021variational} create a latent space representation of event data with a variational autoencoder, cluster the representation with a k-means model, and use either an isolation forest, a local outlier factor, or a one-class \gls{svm} model for each cluster to detect anomalies. 
For their part, \citet{nolle2018binet} and \citet{nolle2019binet} determine anomaly scores based on the output of a \gls{gru} model and map them onto types of anomalies, while \citet{lee2021orientation} check process conformance online using a hidden Markov model.

\subsubsection{Predictive Business Process Monitoring}

In contrast to descriptive monitoring, predictive business process monitoring shifts to an ex-ante perspective by predicting different process properties of running process instances. With these predictions, process users can take proactive or corrective actions to mitigate risks or improve the performance of process executions. Predictive business process monitoring pursues multiple objectives including future activities, outcomes, remaining time, or a combination of objectives. Those approaches facilitate early warning systems.

A common prediction task is the next activity prediction, which deep-learning-based techniques address. 
For the next activity prediction, studies employ \gls{lstm} models \citep{evermann2017,pasquadibisceglie2021multi}, \gls{cnn} models~\citep{pasquadibisceglie2019using,mauro2019activity,pasquadibisceglie2020predictive,heinrich2021process}, \gls{mlp} models \citep{mehdiyev2020novel,theis2019decay}, and transformer models \citep{philipp2020predictive,heinrich2021process}.
Some works use deep-learning models and provide explainable next-activity predictions. \citet{wickramanayake2022building} (and their earlier version \citep{sindhgatta2020exploring}) propose \gls{lstm} models that differ in the attention mechanism they use for the model explanation, \citet{hsieh2021dice4el} use an \gls{nn} architecture computing dynamic features through an \gls{lstm} and static features through an \gls{mlp} and generate explanations for predictions via the model-agnostic counterfactual algorithm DiCE~\citep{mothilal2020explaining}, \citet{hanga2020graph} use an \gls{lstm} model and infer from it business process models to explain the model's decision-making process, and \citet{gerlach2022inferring} use a \gls{gru} model and infer from it multi-perspective likelihood graphs for the same purpose.  
A smaller group of studies proposes Bayesian approaches, that is, regularised probabilistic finite automata models \citep{breuker2016comprehensible} or Bayesian network models \citep{pauwels2020bayesian,brunk2021cause}.   
Other studies compute transition probabilities in process models using decision-tree models~\citep{lakshmanan2015markov} or propose factorisation machines to predict next activities~\citep{lee2018predicting}. \citet{pauwels2021incremental} introduce incremental learning strategies for updating next-activity-prediction models.
%
Finally, instead of only predicting the next activity, \citet{bernard2019accurate} address the remaining activity sequence prediction (or suffix prediction) using clustering and \gls{svm} models to provide transparent predictions. 

Another prediction task is outcome prediction, whose outcome can be binary (e.g. the violation of a business rule), multi-class (e.g. multiple conditions), or continuous (e.g. a \gls{kpi} value).  
%
To predict binary process outcomes in the form of conditions, studies use a \gls{cnn} model \citep{pasquadibisceglie2020orange}, an \gls{lstm} model \citep{wang2019outcome} or a random forest or logistic regression model including features created with text models \citep{teinemaa2016predictive}. Other studies with the same aim use a gated \gls{gnn} model, from which attention-based explanations are extracted \citep{harl2020explainable}, a model built with an evolutionary algorithm and decision rules generated to explain the model's predictions \citep{marquez2017run}, or a neuro-fuzzy model, from which explainable rules are directly extractable \citep{pasquadibisceglie2021fox}. 
%
\citet{rizzi2020explainability} predict the violation of business rules with a random forest model and use post-hoc explanations created with LIME~\citep{ribeiro2016should} to manipulate data to improve model accuracy. 
\citet{di2016clustering} use a DBSCAN model as a density or a model-based clustering approach to cluster partial process instances. Then they apply a decision tree or a random forest model per cluster to predict whether a running process instance violates a business rule.
\citet{folino2012discovering} and \citet{bevacqua2013data} predict violations of service-level agreements. \citet{folino2012discovering} first cluster running process instances. Then, with a cluster-dependent finite state machine model, the remaining time is predicted, and a rule is applied to the predictions to determine the violations. \citet{bevacqua2013data} follow a similar approach but apply a rule to remaining-time or remaining-step predictions generated with regression models to determine the violations.
\citet{cuzzocrea2019predictive} predict violations of aggregated performance constraints. First, running process instances of a window are clustered into variants. Afterwards, with a cluster-dependent linear regression or a k-nearest neighbour model, a performance indicator is predicted for each running process instance. Then, a Gaussian-process regression model predicts an aggregated performance indicator over future window slots, and a rule is applied to the predictions to determine the violations. 
%
As multi-class process outcomes, \citet{folino2011mining} predict behavioural classes of running process instances using a decision-tree model. 
Lastly, continuous process outcomes are predicted as customer delays with non-linear regression or regression tree models~\citep{senderovich2015queue}, as delivery times with various regression models and concept-drift handling \citep{baier2020handling}, and as process performance at the process-model level using a \gls{cnn}, \gls{lstm}, or long-term recurrent \gls{cnn} model~\citep{park2020predicting}. 

Prediction of the remaining time is another common prediction task.
The remaining time of running process instances is predicted using shallow \gls{ml} models considering inter-case features~\citep{senderovich2017intra,klijn2020identifying}, shallow \gls{ml} models considering news sentiments as external process context \citep{yeshchenko2018context}, data-aware transition systems annotated with shallow \gls{ml} models without \citep{polato2014data} and with concept-drift adaption \citep{firouzian2019investigation}, Bayesian \gls{nn} models \citep{weytjens2021learning}, \gls{lstm} models \citep{alves2022context}, and \gls{mlp} models with entity-embedding \citep{wahid2019predictive}.

Some studies do not focus on a specific prediction task but address multiple prediction tasks.
\citet{appice2019leveraging} predict next activities, next timestamps, and remaining time using various feature encodings, learning approaches, and shallow \gls{ml} models. 
Extending their earlier work \citep{polato2014data}, \citet{polato2018time} present three methods that rely on na\"ive Bayes and support vector regression models to predict the remaining time, one of which can also be used to predict suffixes.
In a pioneering work using deep learning in the domain, \citet{tax2017predictive} predict next activities, next timestamps, remaining times, and suffixes using a multi-task \gls{lstm} model. 
This approach \citep{tax2017predictive} has been be extended in various ways, such as with embedding layers for efficient computation \citep{camargo2019learning}, linear temporal logic rules in the post-processing for accurate suffix predictions~\citep{di2017eye}, or text models in the pre-processing to compute textual data \citep{pegorarotextaware}. 
\citet{galanti2020explainable} use an \gls{lstm} model to generate remaining time, activity occurrence, and cost predictions, and apply SHAP~\citep{lundberg2017unified} to create post-hoc explanations for the predictions.  
Other deep-learning-based approaches predict process failure and next activities using a \gls{dnn} model with convolutional and recurrent layers \citep{borkowski2019event}, or next activities and suffixes using an \gls{lstm}-based encoder-decoder model with an attention layer \citep{jalayer2020attention}. 
\citet{pfeiffer2021multivariate} encode different event information as gramian angular fields (i.e. 2D-images), learn a representation based on this encoding using a \gls{cnn} model, and fine-tune the trained model towards various prediction tasks. 
\citet{taymouri2020predictive} use a \gls{gan} model to predict next activities and next timestamps, and \citet{taymouri2021deep} combine these with beam search to predict suffixes and remaining time.

Other studies propose solutions for certain prediction tasks --- predicting activity ordering~\citep{verenich2016minimizing}, enterprise social networks~\citep{pham2021process}, or process outcomes based on scarcely labelled event logs \citep{folino2022semi}. 

\subsubsection{Prescriptive Business Process Monitoring}

Prescriptive business process monitoring aims at steering processes towards specific optimisation objectives. To achieve this, such \gls{ml} applications recommend taking action in process instances. Other approaches warn process owners when a process instance requires human attention; for instance, if a negative outcome is expected. \gls{ml} applications point analysts' attention to important process instances and provide instructions to process workers on how they can carry out their work effectively and efficiently.   

One stream of research investigates warnings that are based on process predictions. 
One indicator of a triggered warning can be the prediction of a process outcome, as \citet{teinemaa2018alarm} propose using a random forest model or a gradient-boosted tree model in combination with a cost model. Another indicator can be the prediction of a process performance indicator, as suggested by~\citet{kang2012periodic} using an \gls{svm} model.
\citet{bozorgi2021prescriptive} use an orthogonal random forest model to recommend whether and when to apply an intervention in a process to decrease running process instances' cycle times. 

Another stream of research investigates recommendations for actions that need to be taken in response to process predictions.
The recommendation of the next best actions is addressed by predicting process risk using a decision-tree model and applying integer linear programming \citep{conforti2015recommendation}. 
Next-best-action recommendations are also determined based on next activity predictions generated by an \gls{lstm} model, \gls{kpi} information, and declarative process model simulation~\citep{weinzierl2020prescriptive}. Alternatively, they are based on \gls{kpi} value predictions obtained using a random forest, \gls{svm}, or decision-tree model and a transition-system abstraction \citep{de2020design}.
\citet{khan2021decision} model the next-best-action recommendation as a Markov decision process and apply deep Q-learning to learn the optimal policy for solving this decision problem.  


\subsubsection{Dynamic Resource Allocation}
While resource allocation planning is an important task in process implementation, dynamic resource allocation is part of daily operations. 
Because of changes in order situations or resource availability beyond expectation, organisations need to reallocate their resources to activities during process execution. This task can be formulated as an optimisation problem. Therefore, \gls{ml} applications optimise resource allocation taking run-time constraints into account, and assist process owners in finding optimal resource utilisation.    
Some papers apply shallow \gls{ml} models to map activities to appropriate workers \citep{liu2008semi}, use na\"ive Bayes models to predict the performance of human resources before incoming jobs are assigned to them based on the predictions \citep{wibisono2015fly}, or apply a k-means model to group process instances at critical activities before these are mapped to available resources~\citep{pflug2016application}.
Other works use a DBSCAN model and an ensemble of \gls{mlp} models to allocate human resources based on team faultiness~\citep{zhao2020human}, or an agglomerative hierarchical clustering model and a k-means model to recommend resources' task preferences~\citep{zhao2016entropy}.
Studies applying reinforcement learning, specifically the Q-learning algorithm, allocate resources via policies that consider time and costs~\citep{huang2011reinforcement} or time and workload balancing~\citep{firouzian2019cycle} as optimisation objectives.
\citet{park2019prediction} use an \gls{lstm} model to predict the next activities and processing time and use the predictions to build a bipartite graph, laying the basis for solving a minimum-cost and maximum-flow problem for resource allocation.

\subsubsection{System Adaption}
Process and system adaptations during execution may be required to adjust to changing process environments. Such adjustments do not necessarily need to trigger an iteration of the full \gls{bpm} lifecycle but can be implemented instantaneously. Self-adaptive systems approaches find adaptations, which improve process performance in the monitoring phase. In contrast to prescriptive business process monitoring, self-adaptive systems apply permanent changes to a process instead of improving a process instance. 
Studies address self-adaptive systems that increase process reliability using an ensemble of \gls{lstm} models~\citep{metzger2019proactive} and reinforcement learning~\citep{metzger2020triggering}.
\citet{saraeian2019optimal} use an \gls{mlp} model to estimate uncertain characteristics to control and optimise an autonomous \gls{bpm} system architecture.
\citet{huang2010adaptive} adapt work distributions based on reinforcement learning, consider process performance goals as optimisation objectives, and learn work distribution policies during process condition changes.
\citet{samiri2017toward} combine reinforcement learning and forecasting to adapt workflows automatically.

\subsubsection{Group Decision-Making}
Group decision-making is only addressed by \citet{de2016framework}, who use reinforcement learning to learn the weighting of decision-makers (e.g. heterogeneous experts) for a decision activity (e.g. select a supplier) based on past process executions considering context and business process performances.

\section{Synthesis of Results and Derived Findings}
\label{sec:synthesis_results_findings}

In this section, we synthesise the results presented in the previous section for each \gls{bpm} lifecycle phase. Based on the synthesised results along with the interactive coding table, we derived findings, which we present in Section~\ref{sec:derived_findings}. 

\subsection{Process Identification}
As data input, most \gls{ml} applications in process identification use event logs, including time or resource information. Unlike other \gls{bpm} lifecycle phases, input data types include unstructured raw event, video, and text data for event log creation.

Moreover, \gls{ml} applications that address process representation creation and learning employ language models trained with \glspl{nn} such as word2vec \citep[e.g.][]{koninck2018act2vec} or use embedding layers in \gls{dnn} models \citep[e.g.][]{guzzo2021multi} for feature extraction. 

In process identification, \gls{ml} applications follow the supervised- and unsupervised-learning approach. Additionally, \gls{ml} applications deal with self-supervised learning because \glspl{nn} are employed to create and learn process representations. These \glspl{nn} are designed to extract a label from underlying data automatically \citep[e.g.][]{nguyen2019autoencoders}. While around half of the \gls{ml} applications in this phase adopt deep learning, almost all use \gls{nn} models. This is because of the flexibility of designing \glspl{nn} used in process representation creation and learning, event log creation, and data improvement applications.        

To assess the performance of models in the tasks process representation creation and learning and event log creation, clustering models are trained with the created process data instances, and clustering metrics are calculated for the clustered instances (e.g. rand index or mutual information score). For this, mainly publicly available event logs are used such as the ones from the BPI challenges\footnote{For example, see Business Process Intelligence Challenge: https://www.tf-pm.org/resources/logs}.

\subsection{Process Discovery}
In process discovery, most \gls{ml} applications use event logs that only include control-flow information as data input to discover process models. Additionally, unstructured data, such as text from e-mails or image data, are used because \gls{ml} applications can extract process models from unstructured data. 

\gls{ml} applications addressing process model extraction use language models trained with \glspl{nn}, such as a BERT model \citep{lopez2021declarative} for feature extraction from unstructured data. For other tasks of this phase, like discovery of procedural process models, feature extraction refers to transforming the control flow of event logs into instance types, such as activity sequences or activity graphs. 

Concerning the \gls{ml} paradigm, \gls{ml} applications of this phase follow the supervised-learning and unsupervised-learning approaches at about the same rate. Because of the lack of label information, unsupervised learning, however, is more common. For example, in discovery of procedural process model, process activity relations are mined in an unsupervised manner using the RIPPER algorithm~\citep{muarucster2006rule}. 
Regarding \gls{ml} concepts, deep learning and transfer learning are adopted in process model extraction applications. While deep learning facilitates learning process model patterns from unstructured input data, transfer learning supports learning such patterns through pre-trained models. 
Moreover, clustering, Bayesian, decision-tree, genetic, and rule-based algorithms are used in process discovery.  
Clustering algorithms facilitate generating comprehensible process models for further analysis, as, for example,~\citet{greco2006discovering} do.
Bayesian-based algorithms can induce process structures from data~\citep[e.g.][]{herbst2000machine}. 
Decision-tree-based algorithms are applied to create decision-tree models in decision-mining applications. 
Genetic-based algorithms are used to find optimal process models according to one or more selected process model quality criteria, as, for example,~\citet{vazquez2015prodigen} do.
Rule-based algorithms are applied to learn rules from event logs; such rules represent a declarative process model in process discovery~\citep[e.g.][]{maggi2012efficient}.

The model assessment uses process discovery metrics from process mining, such as fitness, precision or generalisation, indicating how good the \gls{ml}-created process models are. Often, own event logs are used for model assessment, but also publicly available logs.

\subsection{Process Analysis}
Most \Gls{ml} applications use event logs, including time, resource, or data-related context information. Text data or process models are only occasionally used as data input. Some approaches also integrate domain knowledge.

\gls{ml} applications in process analysis propose individual feature engineering techniques for event log data, and commonly used techniques cannot be observed, even when considering single tasks.  

Like in process discovery, the supervised-learning and unsupervised-learning approach is addressed at about the same rate. This ratio can be explained by trace clustering applications, which are prominent in process analysis and address unsupervised-learning problems \citep[e.g.][]{bose2009context}. 
Analysis applications adopt deep, ensemble, active, meta, and multi-view learning. 
Deep learning is used to learn accurate models for process analysis tasks by considering intricate structures in event log data, as \citet{camargo2022learning} do for process simulation. 
Ensemble learning is adopted to detect patterns based on the outcome of models, which are trained with different supervised \gls{ml} algorithms \citep[e.g.][]{cuzzocrea2016multi}.      
Active learning is used to increase \gls{ml} models' accuracy by selecting process instances from an event log based on a metric, as in \citet{weerdt2013active} for trace clustering, or human judgement, as in \citet{cuzzocrea2016robust} for pattern detection. 
A meta-model is learned based on the outcome of several base models to detect patterns in event log data \citep[e.g.][]{cuzzocrea2016multi}.
Multi-view learning is adopted to train multiple shallow \gls{ml} models based on event logs, where each model refers to a view for accurate pattern detection \citep[e.g.][]{cuzzocrea2015multi} or complexity-reduced trace clustering \citep{appice2015co}.
\gls{ml} applications in process analysis use unsupervised clustering algorithms for trace clustering.
Further, as in process discovery, we observe that rule-based algorithms are employed here to learn models as a set of explainable rules from event log data; these rules reveal the cause of delayed processes~\citep[e.g.][]{ferreira2015using}. 

Process discovery metrics (e.g. fitness, precision, or generalisation) or clustering metrics (e.g. cluster set entropy or mutual information score) are calculated to assess the performance of clustering models employed in trace clustering applications. For applications addressing other tasks in process analysis, common supervised \gls{ml} metrics are calculated for model assessment. Mainly publicly available event logs, but sometimes also own event logs are used. 

\subsection{Process Redesign}

\Gls{ml} applications for (semi-)automatic process redesign receive business process designs consisting of activities, connections, and routing decisions \citep[e.g.][]{afflerbach2017design} as data input, while \gls{ml} applications for redesign suggestion identification can compute event log data \citep{Mustansir2022}.
%
In \gls{ml} applications for process redesign, feature engineering is not explicitly described.
%
\gls{ml} applications for process redesign follow the supervised-learning approach as both tasks in this phase, redesign suggestion identification~\citep[e.g.][]{Mustansir2022} and (semi-)automatic process redesign \citep[e.g.][]{afflerbach2017design}, require guidance for model training. 
Concerning \gls{ml} algorithms, \glspl{nn} are used for detecting redesign suggestions, while (semi-)automatic process redesign is addressed using genetic algorithms. 
%
Common supervised \gls{ml} metrics and \glspl{kpi} (e.g. customer satisfaction) are used with publicly available event logs for model assessment.

\subsection{Process Implementation}

\Gls{ml} applications for resource allocation receive event logs with resource information as data input \citep[e.g.][]{liu2012mining}. Other dataset types can also be found in this phase, such as business process problems~\citep[e.g.][]{bae2014planning}.   
%
Regarding feature engineering, \gls{ml} applications for process implementation only selectively describe the extraction of features. However, common techniques for feature extraction cannot be observed. 
%
\gls{ml} applications in this phase follow the supervised- and unsupervised-learning approach. However, there is a tendency towards the supervised-learning approach. Resource allocation planning can explain this ratio \citep[e.g.][]{xu2016resource}, which is addressed equally using supervised- and unsupervised-learning algorithms. Further, compared to other \gls{ml} algorithm groups, genetic-based algorithms tend to be used more for process implementation. 
%
Resource-specific or time-related measures are calculated using mainly publicly available event logs for model assessment of resource allocation planning applications, such as r-precision \citep[e.g.][]{delcoucq2022resource} or processing time \citep[e.g.][]{xu2016resource}.

\subsection{Process Monitoring}

While other \gls{bpm} lifecycle phases work with event logs of fixed sizes, process monitoring applications receive continuous data streams in practice. Still, most of the recent process monitoring approaches are developed based on fixed-sized event logs~\citep[e.g.][]{brunk2021cause, heinrich2021process,evermann2017}. Because of their public availability for research purposes and their static properties (e.g. a static number of activities), event logs are preferred over event streams in research. These event logs are generally not limited to the control flow and include time, resource, or data-related context information. Additionally, some approaches integrate domain knowledge.

Monitoring applications rely on established techniques to extract features from event log data. Examples of such encoding techniques are the boolean, frequency, or index-based technique \citep[e.g.][]{leontjeva2016complex}. 
Monitoring applications incorporate features that are created manually from event log data that describe certain aspects of a business process, facilitating the learning of accurate \gls{ml} models. For instance, temporal features can be created based on timestamps of events, such as the time from the first event of a process instance to the current event of the same process instance \citep[e.g.][]{tax2017predictive}. Another example is creating inter-case features, such as the number of resources that perform a certain activity~\citep[e.g.][]{senderovich2017intra}. 
Feature selection is applied in this phase. For example, to learn accurate models by choosing an event log's most relevant context features~\citep[e.g.][]{alves2022context}.

Regarding model building, monitoring applications cope primarily with supervised learning, perhaps because \gls{ml} is used to learn mappings from process instances or prefixes (i.e. subsequences of process instances) to targets~\citep[e.g.][]{maggi2014predictive}. 
Monitoring applications adopt deep learning, ensemble learning, multi-task learning, online learning, and transfer learning.
In predictive and prescriptive business process monitoring, process instances from event logs are converted into prefixes to create predictions or prescriptions as early as possible, increasing the volume of data in an event log. As deep learning profits from large data volume, \gls{ml} applications in this area often use deep-learning models, such as for predicting the next activity, as \citet{evermann2017} do.
Ensemble learning is adopted for addressing all tasks of the monitoring phase using the outcome of several models, which are trained with supervised-learning algorithms \citep[e.g.][]{metzger2019proactive}.
The concept of multi-task learning is used when several process indicators should be monitored, leading to the use of, for instance, \gls{dnn} models that are designed to approximate multiple learning targets simultaneously \citep[e.g.][]{tax2017predictive}. 
The concept of incremental learning assumes a continuous data stream as input, so the concept is present in the process-monitoring phase. Concept-drift detection, in particular, appears with incremental learning~\citep[e.g.][]{baier2020handling}. Because concept-drift detection aims to identify changes in labels or distributions in event stream data, it is a good match for these approaches. 
Only a few studies use transfer learning and all these studies are realised with \glspl{dnn} that distinguish between representation learning and fine-tuning~\citep[e.g.][]{pfeiffer2021multivariate}. After learning a generic representation from scratch, the resulting model can be fine-tuned for a desired task. A pre-trained model can also be fine-tuned to transfer from the general representation to the task.
Instance, tree, regression, and \gls{nn}-based algorithms are often employed in the process-monitoring phase because they include popular supervised \gls{ml} algorithms that can be used for a variety of prediction tasks (e.g. the tree-based algorithm C4.5).
Bayesian algorithms are also common in predictive business process monitoring. 
This is because transition probabilities are often learned with a Bayesian algorithm for a given process model and a certain prediction task before the process model is augmented with the learned probabilities. In doing so,  the process model is transformed into a predictive model.
Reinforcement-learning algorithms are used for a few process-monitoring tasks, such as next-best-action recommendations in prescriptive business process monitoring, as in \citet{khan2021decision}, or dynamic resource allocation, as in \citet{firouzian2019cycle}. Because of the required mapping between actions and a certain state or change of state in a system or an environment, reinforcement learning is particularly well suited to these tasks.
Clustering algorithms are also used in predictive business process monitoring to train accurate predictive models per cluster, as~\citet{di2016clustering} do.
Additionally, some studies focus on the explanation of \gls{ml} applications, as \citet{galanti2020explainable} do by applying SHAP for predictive business process monitoring. 

For some tasks, such as predictive business process monitoring, the quality of \gls{ml} models can be directly assessed. Because of the availability of ground-truth labels in event logs (e.g. next activities), established metrics from supervised \gls{ml} research can be calculated. 
For other tasks, model assessment must consider metrics other than these standard metrics. 
For example, prescriptive monitoring applications take \glspl{kpi}, often defined based on context features, like cost savings \citep[e.g.][]{bozorgi2021prescriptive} or expected throughput-time~\citep[e.g.][]{weinzierl2020prescriptive} into account for model assessment. 
Also, measuring time-related metrics is important in this phase as models are applied in running business processes. For example, prediction time \citep{zhao2016entropy} or earliness \citep{teinemaa2016predictive} are such metrics.  
For model assessment mainly publicly available event logs are used. Little approaches use own event logs. 
Further, monitoring applications use a split validation strategy more often than a cross-validation strategy compared to other \gls{bpm} lifecycle phases.
This preference towards split validation is due to retaining the process data's natural structure~\citep{weytjens2021creating}.

\subsection{Derived Findings}
\label{sec:derived_findings}
The results described in the previous sections are summarised in Table~\ref{tab:findings} for each \gls{bpm} lifecycle phase and \gls{ml} model development phase. Based on these synthesised results, we derived overarching findings. 

\noindent \textbf{\emph{Finding 1: \gls{ml} applications mainly address the event-data-intensive \gls{bpm} lifecycle phases process discovery, analysis and monitoring.}}
A look into the number of reviewed papers shows that most \gls{ml} applications address process monitoring, analysis and discovery. 
Due to the availability of input data, especially process data in the form of event log data, \gls{ml} applications prevail in these phases. 
Therefore, various prediction and prescription tasks have been addressed in predictive and prescriptive business process monitoring, various pattern detection and trace clustering tasks in process analysis, and various model discovery tasks in process discovery. 
On the contrary, for example, it is more challenging to develop \gls{ml} applications in non-event-data-intensive phases, where process identification relies on documentation, process redesign on redesign heuristics and practices, and process implementation on work plans.   

\noindent \textbf{\emph{Finding 2: \Gls{ml} applications along the \gls{bpm} lifecycle mainly use event logs with limited process context.}} 
Most \gls{ml} applications in \gls{bpm} are developed based on publicly available event logs. These event logs document a business process or its sub-types and include a limited amount of context data. 
However, in organisations, information systems (e.g. ERP systems) can add more context data to event logs than publicly available event logs. That is important for \gls{ml} applications of \gls{bpm} lifecycle phases, where context information plays a role, such as process monitoring or analysis. 

\noindent \textbf{\emph{Finding 3: New forms of process data enable new use cases.}} 
New process data forms have led to the development of \gls{ml} applications that address new \gls{bpm} tasks. However, new \gls{bpm} tasks can already be addressed if an event log includes a certain context attribute, such as the resource attribute for the tasks resource allocation planning in process identification or dynamic resource allocation in process monitoring. 
In the case of unstructured forms of data, event log creation in process identification is possible with video or text data, and process model extraction in process discovery can be addressed with text or image data.
Other \gls{bpm} tasks, such as (semi-)automatic process redesign, are enabled with more specific forms of data like business process designs.

\noindent \textbf{\emph{Finding 4: Integrating domain knowledge can advance \gls{ml} applications.}} 
Domain knowledge in various forms has been integrated into \gls{ml} applications to fulfil different purposes. 
For example, process models are integrated into prescriptive-business-process-monitoring applications to ensure the conformance of next best actions \citep[e.g.][]{weinzierl2020prescriptive}, constraints integrated into trace-clustering applications to control the clustering of traces \citep[e.g.][]{koninck2021expert}, or logical rules integrated into predictive-business-process-monitoring applications to improve the prediction performance \citep[e.g.][]{di2017eye}.

\noindent \textbf{\emph{Finding 5: Common feature extraction techniques are only used for a few \gls{bpm} tasks.}}
Considering feature extraction techniques of \gls{ml} applications across \gls{bpm} tasks, we observe that \gls{ml} applications apply the same feature extraction techniques for a few tasks. In contrast, specific feature extraction techniques were proposed for the other tasks.
For example, \gls{ml} applications for predictive business process monitoring often use the same techniques to realise feature extraction from event log data.
These techniques facilitate the development of predictive-business-process-monitoring applications, as finding an appropriate technique to extract features from a given event log can be challenging. 
Examples of such techniques are boolean, frequency, or index-based sequence encoding \citep{leontjeva2016complex}. 

\noindent \textbf{\emph{Finding 6: Deep learning is the dominant \gls{ml} concept in \gls{bpm}.}}
Deep learning is the most adopted \gls{ml} concept in \gls{bpm}. The main reason for this is that the architecture of \gls{dnn} models can be constructed flexibly depending on the desired tasks \citep{goodfellow2016deep}.
For example, autoencoder models reproducing the input are constructed in process identification to create process representations \citep[e.g.][]{guzzo2021multi}. In contrast, \gls{gnn} models considering event dependencies in the form of edges are constructed in process discovery to detect generalisable process models~\citep[e.g.][]{sommers2021process}.

\noindent \textbf{\emph{Finding 7: \gls{ml} applications mainly focus on improving performance.}}
\gls{ml} applications in \gls{bpm} are developed to achieve a performance improvement. A performance improvement can be a higher prediction or detection performance, a lower training or inference time, or a trade-off of various performance criteria.   

\noindent \textbf{\emph{Finding 8: Explainability of \gls{ml} applications is primarily addressed in process monitoring.}}
Although \gls{bpm} research already develops explainable \gls{ml} applications based on approaches from the field of \gls{xai}\footnote{The focus of this paper is on \gls{ml}, a sub-field of \gls{ai}. However, as some concepts or research areas are branded with the term \textquote{\gls{ai}}, we use \textquote{\gls{ai}} rather than \textquote{\gls{ml}}.}, these \gls{ml} applications are mainly developed for predictive business process monitoring~\citep{stierlexpbpm}.  
Specifically, when \gls{dnn} models are used for predictive business process monitoring, which are perceived as black boxes, \gls{ml} applications aim to make model decisions transparent for process users or other process stakeholders.

\noindent \textbf{\emph{Finding 9: Model assessment of \gls{ml} applications is mainly done with data-based metrics.}}
\gls{ml} applications in \gls{bpm} are typically assessed with data-based metrics. 
Common supervised \gls{ml} metrics, refinements of such, or \glspl{kpi} are used in \gls{bpm} tasks, based on supervised learning.
In contrast, model assessment is often realised indirectly for \gls{bpm} tasks where no supervised learning is addressed. 
That is, a not-supervised-learned \gls{ml} model is applied, and then clustering or process discovery metrics are calculated based on the model output. In doing that, model assessment can be data-based, even if no label information is available.

\begin{landscape}
    
\begin{table}[ht]
\caption{Overview of synthesised results and derived findings}
\label{tab:findings}
\scriptsize

\begin{tabular}{p{2.5cm}p{5cm}p{5cm}p{5cm}p{5cm}}
\toprule
\textbf{}                 
& \multicolumn{1}{c}{\textbf{Data input}}                                                   & \multicolumn{1}{c}{\textbf{Feature engineering}}                                          & \multicolumn{1}{c}{\textbf{Model building}}                                               & \multicolumn{1}{c}{\textbf{Model assessment}}                                                                                                                                        \\ \midrule

\begin{itemize}[left=0pt,topsep=0pt, noitemsep] \item[] \textbf{Process} \item[] \textbf{identification} 
\item[] (12 paper) \end{itemize}

&\begin{itemize}[left=0pt,topsep=0pt, noitemsep] \item Mostly event logs with time- and resource-related context information 

\item Other dataset types for \textit{event log creation} (e.g. raw event, video, and text data) \end{itemize}

&  \begin{itemize}[left=0pt,topsep=0pt, noitemsep] \item For \textit{process representation creation and learning} language models based on artificial \glspl{nn} (e.g. word2vec or doc2vec) or artificial \glspl{nn} with embedding layer \end{itemize}

&  \begin{itemize}[left=0pt,topsep=0pt, noitemsep] \item Besides supervised and unsupervised learning relatively often self-supervised learning for \textit{process representation creation and learning} \item Around half of the \gls{ml} applications adopt deep learning 
\item Almost all \gls{ml} applications use an artificial \gls{nn} \end{itemize}

& \begin{itemize}[left=0pt,topsep=0pt, noitemsep] \item For \textit{process representation creation and learning} and \textit{event log creation} task clustering models (k-Means) and clustering metrics (e.g. rand index or mutual information score) \item Mainly publicly available event logs (e.g. BPI challenge event logs) \end{itemize}                                                                   \\

\begin{itemize}[left=0pt,topsep=0pt, noitemsep] \item[] \textbf{Process} \item[] \textbf{discovery} 
\item[] (45 paper) \end{itemize}

&  \begin{itemize}[left=0pt,topsep=0pt, noitemsep] \item Mostly event logs without context information
\item Sometimes other dataset types for \textit{process model extraction} (e.g. text or image data) \end{itemize}

&\begin{itemize}[left=0pt,topsep=0pt, noitemsep] \item For \textit{process model extraction} language models based on artificial \glspl{nn} (e.g. word2vec or BERT) \item
For other tasks, feature engineering refers to the transformation of control-flow data into certain instance types (e.g. activity sequences or activity graphs) \end{itemize}

& \begin{itemize}[left=0pt,topsep=0pt, noitemsep] \item Supervised and unsupervised at about the same rate; unsupervised more common because of lack of label data (e.g. \textit{discovery of declarative process model})\item Occurrence of deep learning and transfer learning for \textit{process model extraction} more than once \item Relatively often Bayesian-based, clustering-based, rule-based, and genetic-based  \end{itemize}

&\begin{itemize}[left=0pt,topsep=0pt, noitemsep] \item Often process discovery metrics (e.g. precision, generalisation, or fitness) \item Often own event logs, but also publicly available logs \end{itemize}
\\

\begin{itemize}[left=0pt,topsep=0pt, noitemsep] \item[]\textbf{Process} \item[] \textbf{analysis}\item[] (44 paper) \end{itemize}  

&\begin{itemize}[left=0pt,topsep=0pt, noitemsep] \item Mostly event logs with time-, resource-, or data-related context 
\item Rarely other data types (e.g. text data or set of process models)
\item Sometimes domain knowledge integrated
\end{itemize}	\nointerlineskip
& \begin{itemize}[left=0pt,topsep=0pt, noitemsep] \item Different approaches; no standards regarding feature extraction, creation, or selection \end{itemize}
& \begin{itemize}[left=0pt,topsep=0pt, noitemsep] \item Supervised and unsupervised at about the same rate
\item  Deep Learning (various tasks), ensemble learning (\textit{pattern detection}), active learning (\textit{pattern detection and trace clustering}), meta Learning (\textit{pattern detection}), and multi-view learning (\textit{pattern detection}) more than once \item
Often clustering; relatively often rule-based (as in process discovery)	 \end{itemize}

& \begin{itemize}[left=0pt,topsep=0pt, noitemsep] \item Often common performance metrics from supervised \gls{ml} \item
Metrics for \textit{trace clustering} often process discovery metrics (e.g. fitness, precision, or generalisation) or clustering metrics (e.g. cluster set entropy or mutual information score) 
\item Mainly publicly available event logs, but sometimes also own event logs \end{itemize}
\\

\begin{itemize}[left=0pt,topsep=0pt, noitemsep] \item[] \textbf{Process} \item[] \textbf{redesign} \item[] (3 paper) \end{itemize}

&   \begin{itemize}[left=0pt,topsep=0pt, noitemsep] \item Event logs for \textit{redesign suggestion detection} \item 
Business process designs for \textit{(semi-) automatic process redesign}	\end{itemize}

& \begin{itemize}[left=0pt,topsep=0pt, noitemsep] \item Feature engineering generally not explicitly described 	\end{itemize}

& \begin{itemize}[left=0pt,topsep=0pt, noitemsep] \item Only supervised learning \item
Genetic-based for \textit{(semi-) automatic process redesign} and artificial-\gls{nn}-based for \textit{redesign suggestion detection}	\end{itemize}
&	\begin{itemize}[left=0pt,topsep=0pt, noitemsep] \item Common performance metrics from supervised \gls{ml} and KPIs (e.g. customer satisfaction) \item Mainly publicly available event logs\end{itemize}
\\
\bottomrule

\end{tabular}
\end{table}

\end{landscape}

\begin{landscape}
    
\begin{table}[ht]
\scriptsize
\begin{tabular}{p{2.5cm}p{5cm}p{5cm}p{5cm}p{5cm}}

\toprule

\textbf{}    
& \multicolumn{1}{c}{\textbf{Data input}}                                                   & \multicolumn{1}{c}{\textbf{Feature engineering}}                                          & \multicolumn{1}{c}{\textbf{Model building}}                                               & \multicolumn{1}{c}{\textbf{Model assessment}}                                                                                                                                                                                                                                   \\ \midrule
\begin{itemize}[left=0pt,topsep=0pt, noitemsep] \item[]\textbf{Process} \textbf{implementation} \item[](6 paper) \end{itemize}

&   \begin{itemize}[left=0pt,topsep=0pt, noitemsep] \item Event logs with resource information for \textit{resource allocation planning} \item  Other data set types can be found (e.g. set of business process problems) \end{itemize}

&	\begin{itemize}[left=0pt,topsep=0pt, noitemsep] \item Feature extraction sometimes described, but common patterns cannot be observed	\end{itemize}

&	\begin{itemize}[left=0pt,topsep=0pt, noitemsep] \item Only supervised and unsupervised learning with a tendency to supervised learning as \textit{resource allocation planning} is addressed about the same rate with supervised and unsupervised learning	\end{itemize}

&	\begin{itemize}[left=0pt,topsep=0pt, noitemsep] \item Resource-specific measures (e.g. r-precision) or time-specific measures (e.g. processing time) \item Mainly publicly available event logs \end{itemize} \\

\begin{itemize}[left=0pt,topsep=0pt, noitemsep] \item[] \textbf{Process} \item[] \textbf{monitoring} \item[] (90 paper)   \end{itemize}
& \begin{itemize}[left=0pt,topsep=0pt, noitemsep] \item Continuous data streams in theory, but in practice mostly event logs with fix size \item Event log often with time-related, resource-related, or data-related context \item Sometimes domain knowledge integrated\end{itemize}& \begin{itemize}[left=0pt,topsep=0pt, noitemsep] \item Certain techniques for feature extraction from event logs (e.g. event- or sequence-encoding techniques) \item
Manual feature creation from event log data (e.g. temporal or inter-case features) \item
Feature selection sometimes applied (e.g. to select relevant context features) \end{itemize}	&	\begin{itemize}[left=0pt,topsep=0pt, noitemsep] \item Mostly supervised learning \item
Occurrence of deep learning (various tasks), ensemble learning (various tasks), multi-tasking learning (various tasks), incremental learning (various tasks), and transfer learning (\textit{predictive business process monitoring}) more than once \item
Often \gls{nn}-based, instance-based, and tree-based, regression-based, Bayesian-based \gls{ml} applications for \textit{predictive business process monitoring}, reinforcement-learning-based for few process monitoring tasks (e.g. \textit{prescriptive business process monitoring} or \textit{dynamic resource allocation}), and clustering-based for \textit{predictive business process monitoring}\item Explainability of \gls{ml} applications is sometimes focused \end{itemize} & \begin{itemize}[left=0pt,topsep=0pt, noitemsep] \item Often established performance metrics from supervised \gls{ml} research (for some tasks) \item 
Often metrics defined based on context features for \textit{prescriptive business process monitoring} (e.g. cost savings) \item
 Relatively often time-related metrics (e.g. prediction time or earliness) \item
More often split validation than cross-validation compared to other phases \item Mainly publicly available event logs; seldomly own event logs \end{itemize}
\\ \midrule

\begin{itemize}[left=0pt,topsep=0pt, noitemsep] \item[] \textbf{Derived} \item[]  \textbf{findings} \end{itemize}

& \begin{itemize}[left=0pt,topsep=0pt, noitemsep]
\item[] \textbf{F1:} ML applications mainly address the event-data-intensive BPM lifecycle phases process discovery, analysis and monitoring
\item[] \textbf{F2:} ML applications along the BPM lifecycle mainly use event logs with limited process context 
\item[] \textbf{F3:} New forms of process data enable new use cases
\item[] \textbf{F4:} Integrating domain knowledge can advance ML applications
\end{itemize}

& \begin{itemize}[left=0pt,topsep=0pt, noitemsep] \item[] \textbf{F5:} Common feature extraction techniques are used for a few BPM tasks
\end{itemize}

& \begin{itemize}[left=0pt,topsep=0pt, noitemsep]
\item[] \textbf{F6:} Deep learning is the dominant ML concept in BPM
\item[] \textbf{F7:} ML applications mainly focus on improving performance
\item[] \textbf{F8:} Explainability of ML applications is primarily addressed in process monitoring
\end{itemize}

& 
\begin{itemize}[left=0pt,topsep=0pt, noitemsep]
\item[] \textbf{F9:} Model assessment of ML applications is   mainly done with data-based metrics
\item[] \textbf{F10:} Most ML applications along the BPM lifecycle use benchmark event logs
\end{itemize} \\

\bottomrule
\end{tabular}
\end{table}

\end{landscape}

\noindent\textbf{\emph{Finding 10: Most \gls{ml} applications along the \gls{bpm} lifecycle use benchmark event logs.}} 
Our reviewed papers show that \gls{ml} applications are mostly evaluated using benchmark experiments. 
Further, most benchmark experiments use publicly available event logs. 
\Gls{bpm} research, in particular predictive business process monitoring research, has elaborated on many ways to address prediction tasks with those event logs. Therefore, identifying accurate \gls{ml} models for a business process captured in one of the publicly available event logs is consequently quite simple. 
When addressing a different business process, it is hard to identify promising approaches based on those benchmarks, as they only cover a selection of possible process types and characteristics. Consequently, transferring the acquired findings and insights into practice is limited.

\section{Discussion}
\label{sec:discussion}
This section discusses future research directions we derived from our findings, implications of our literature review for research and practice, and limitations of our study.

\subsection{Future Research Directions}
\label{sec:findings}

We propose ten future research directions to advance research on \gls{ml} applications in \gls{bpm}, as summarised in Table~\ref{tab:futureresearch}. We identified these research directions based on the findings from our literature review and structured them along the phases of the \gls{ml} model development process.

\begin{table}[!ht]
\caption{Future research agenda}
\label{tab:futureresearch}
\scriptsize
\begin{tabular}{@{}p{0.5cm}p{14cm}p{1.5cm}@{}}
\toprule
\multicolumn{2}{l}{\textbf{Research direction}}                                                                                                & \textbf{Based on} \\ \midrule
\multicolumn{3}{l}{\textbf{Data input}}\\
\multicolumn{1}{l}{1} & 
\Gls{ml} applications for event-data-intensive \gls{bpm} lifecycle phases (i.e. process discovery, analysis and monitoring) received much attention. Exhibiting unstructured data enables the development of \gls{ml} applications for less event-data-intensive \gls{bpm} lifecycle phases, where especially process data in the form of event log data are not or only to a limited extent available (i.e. process identification, redesign, and implementation). & F1\\
\multicolumn{1}{l}{2} & Most \gls{ml} applications along the \gls{bpm} lifecycle mainly use event logs with limited process context. However, in real-world cases, event logs contain more information about processes; for example, machine parameters in manufacturing. To overcome that, future research should develop \gls{ml} applications considering the enterprise process network and their data.& F2\\
\multicolumn{1}{l}{3} & New forms of process data enable new use cases. As object-thinking is gaining attention in practice and academia, process data is increasingly stored as object-centric event logs. Object-centric event logs include object dependencies and thus enable new use cases. Therefore, future research should investigate the development of \gls{ml} applications for object-centric event logs. & F3\\
\multicolumn{1}{l}{4} & Business processes depend on their execution domain and integrating domain knowledge can advance \gls{ml} applications. Process rules and heuristics may be known in advance. Therefore, in future research, the integration of domain knowledge into \gls{ml} applications should be further investigated. & F4\\ 
\multicolumn{3}{l}{\textbf{Feature engineering}}\\
\multicolumn{1}{l}{5} & 
Most feature extraction techniques used in \gls{ml} applications for \gls{bpm} are created paper for paper and common feature extraction techniques are only used for a few \gls{bpm} tasks. Therefore, the development of general feature extraction techniques and the use of those is a direction for future research. For this, generative \gls{ai} approaches, such as large language models, are promising. & F5\\
\multicolumn{3}{l}{\textbf{Model building}}\\
\multicolumn{1}{l}{6} & Deep learning is the dominant \gls{ml} concept in \gls{bpm} research. However, beyond deep learning, there are \gls{ml} concepts that have received little attention in \gls{bpm} research but are promising to advance \gls{ml} applications and unlock future research. For example, such an \gls{ml} concept is transfer learning in the context of standard business processes. 
& F6 \\
\multicolumn{1}{l}{7} & \gls{ml} applications strongly emphasise improving process performance, which can be unsuitable for successful use in practice. Therefore, incorporating aspects of ethical \gls{ai}, including transparency, justice, fairness, non-maleficence, responsibility and privacy, into \gls{ml} applications is a promising direction for future \gls{ml} research in \gls{bpm}. &  F7 \\
\multicolumn{1}{l}{8} & While explainability is primarily addressed in the development of \gls{ml} applications for process monitoring tasks, future research should consider explainability in novel \gls{ml} applications for other tasks along the \gls{bpm} lifecycle. The use of \gls{xai} approaches and the symbiosis with domain knowledge are important aspects in this future research direction. 
&  F8 \\ 
\multicolumn{3}{l}{\textbf{Model assessment}}\\
\multicolumn{1}{l}{9} & Model assessment in \gls{ml} applications in \gls{bpm} is mostly done using standard data-based metrics. Beyond data-based assessment, human-centric and economic metrics related to the \gls{bpm} domain may fit the model assessment more accurately. Therefore, future research should develop such metrics for model assessment for \gls{bpm} tasks (e.g. regarding efficiency and explainability of \gls{ml} models). & F9 \\ 
\multicolumn{1}{l}{10} & Even though experiments with benchmark event logs are the de facto standard for assessing the utility of \gls{ml} applications in \gls{bpm}, generalizability and transfer of findings and insights can be limited. Therefore, future research should propose new approaches to support assessing the utility of \gls{ml} applications.  
For example, promising approaches include the elaboration of methodological and technical guidance for benchmark experiments or the application of qualitative and quantitative methods to gain further insights from practitioners. 
& F10 \\
\bottomrule
\end{tabular}
\end{table}

\noindent\textbf{\emph{Research direction 1:}} 
According to Finding 1, \gls{ml} applications in \gls{bpm} mainly address tasks in the event-data-intensive \gls{bpm} lifecycle phases of process discovery, analysis and monitoring, where process data in the form of event log data is typically available. 
Therefore, we propose as future research direction the examination of other data sources from different enterprise systems and the investigation of approaches to induce structure into unstructured and semi-structured data.
Consequently, future research can develop novel \gls{ml} applications for the less event-data-intensive \gls{bpm} lifecycle phases (i.e. process identification, redesign, and implementation). 

To make unstructured process data in these lifecycle phases usable, techniques from the \gls{nlp} \citep[e.g.][]{otter2020survey} or computer vision \citep[e.g.][]{voulodimos2018deep} field, which are deeply interconnected with \gls{ml} approaches, could facilitate the development of novel \gls{ml} applications. 
For example, setting up organisational business process landscapes in process identification with \gls{nlp} techniques based on textual documentation~\citep[e.g.][]{van2018challenges}, suggesting and testing redesign opportunities in process redesign with computer vision techniques based on video recordings~\citep[e.g.][]{Mustansir2022}, and supporting the implementation of business processes in suitable information systems with \gls{nlp} techniques based on text-based work plans~\citep[e.g.][]{xu2016resource}. 

\noindent\textbf{\emph{Research direction 2:}} 
In accordance with Finding 2, \gls{ml} applications along the \gls{bpm} lifecycle mainly use event logs with limited process context.
Therefore, we propose as future research direction the development of novel \gls{ml} applications, which consider the enterprise process network instead of isolated business processes~\citep{oberdorf2023predictive}. In doing that, \gls{ml} applications receive input data from multiple data sources, including control-flow information from different business processes and process context information related to the business processes 
As more context is considered in such \gls{ml} applications, the selection of valuable context information via feature selection increases in importance. 
Finally, as business processes can span multiple organisations, process networks are not limited to one organisation. \gls{ml} applications could thus span across multiple organisations, providing insights and recommendations for supply chain processes.  

\noindent\textbf{\emph{Research direction 3:}} 
Based on new forms of process data (Finding 1), including object-centric event logs, future research should explore further use cases to develop novel \gls{ml} applications. 
%
Currently, both process-mining vendors and academia are adjusting to object-thinking in \gls{bpm}~\citep[e.g.][]{li2018extracting}, which originated in artefact-centric workflow modelling~\citep{nigam2003business}. 
Consequently, process data are more and more stored as object-centric event logs.
\textquote{Object-centric} is a process paradigm, after which an instance of a business process is not performed in isolation like in the process-instance-centric process paradigm, but interacts via objects with other instances of business processes \citep{van2019object}
According to the object-centric process paradigm, an object-centric event log stores object dependencies, which are neglected in traditional event logs \citep{vanderAalst2016action}. 
Therefore, object-centric event logs reveal new use cases, such as predicting remaining object interactions in the next 24 hours, detecting incorrect object interactions, or repairing missing object interactions in object-centric event log data.

\noindent\textbf{\emph{Research direction 4:}} 
Derived from Finding 4, integrating domain knowledge into model building can advance \gls{ml} applications. Therefore, the integration of domain knowledge into \gls{ml} applications should be further investigated. For example, this includes i) the integration of multiple entities of domain knowledge, ii) the amount and form of integrated domain knowledge required for solving certain learning problems, and iii) the integration of profound domain knowledge from \gls{bpm}.  

Concerning i), each entity can express its domain knowledge as \textquote{logical rules, constraints, mathematical equations [...], probability distributions, similarity measures, knowledge graphs or ontologies}~\citep[p.~86]{folino2021ai}. These entities can also be related to each other in diverse ways or be structured into various areas or levels.
Predicting future process behaviour could benefit from integrating domain knowledge as employee-related rules and a company-related ontology into \gls{ml} applications.
Concerning ii), \gls{ml} research contains work that proposes approaches that allow controlling which domain knowledge should be integrated into a specific situation into an \gls{ml} algorithm. For example, \citet{maier2019learning} propose operators to incorporate prior knowledge into \gls{ml} algorithms in a controlled manner.
The development of future \gls{ml} applications could include control mechanisms to handle intended exceptional process flows that, for example, mitigate risks. 
Concerning iii), profound domain knowledge from the \gls{bpm} domain could be present as redesign heuristics, or topical and organisational constraints \citep[e.g.][]{dumas2018fundamentals}. The integration of \gls{bpm} knowledge could aid in automatically suggesting improved process models and definitions.

\noindent\textbf{\emph{Research direction 5:}} 
According to Finding 5, common feature extraction techniques are only used for a few \gls{bpm} tasks.
Therefore, future research should propose general feature extraction techniques for \gls{bpm}.  
Particularly interesting are tasks in \gls{bpm} lifecycle phases for which various specific techniques have been proposed (e.g. pattern detection in process analysis). 
The common feature extraction techniques should also consider the extraction of features from dataset types apart from event logs (e.g. text or video)~\citep{kratsch2022shedding}. 

To realise common feature extraction techniques, the use of language models seems to be a promising direction. 
In our reviewed papers, we observed that (large) language models (e.g. word2vec and BERT) are increasingly used in \gls{ml} applications either as primary components to solve \gls{bpm} tasks (e.g. process representation creation and learning \citep[e.g.][]{koninck2018act2vec} or process model extraction \citep[e.g.][]{qian2020approach}) or as secondary components to support other \gls{bpm} tasks through learning precise representations of event log or text data (e.g. predictive business process monitoring \citep[e.g.][]{teinemaa2016predictive} or pattern detection \citep[e.g.][]{junior2020anomaly}). 

When pre-trained large language models (e.g. the generative pre-trained transformer 4 (GPT-4) \citep{openai2023}) are fine-tuned and evolve from the pure \gls{ml} model to a system (e.g. a conversational agent like ChatGPT \citep{guo2023close}) or application (e.g. content generation like search engine optimisation) \citep{feuerriegel2023generative}, they open up new possibilities for \gls{ml}-based \gls{bpm} research \citep{vidgof2023large}.
For example, for predictive business process monitoring, conversational agents can be used during process execution, and process users can ask questions to these agents about different aspects of a running process instance. Another example refers to redesign suggestion identification, for which conversational agents can be asked to reveal innovation opportunities for process improvement. Moreover, for process implementation, large-language-model-based applications solving the problem of routine task automation are a new opportunity for \gls{ml}-based \gls{bpm} research.   
Finally, as large language models refer to a type of model used in the context of generative \gls{ai} \citep{feuerriegel2023generative}, other model types, such as diffusion probabilistic models, \glspl{gan} models, or variational autoencoder models, are promising for developing novel \gls{ml} applications in \gls{bpm}. Additionally, models considering data modalities beyond text (e.g. image, audio, and code) or combinations of them (e.g. image-to-text or audio-to-text) are promising for future \gls{bpm} research on \gls{ml} applications.

\noindent\textbf{\emph{Research direction 6:}} 
In accordance with Finding 6, deep learning is the dominant \gls{ml} concept in \gls{bpm}. However, beyond deep learning, we propose as future research direction the development of novel \gls{ml} applications considering \gls{ml} concepts that have received little attention in \gls{bpm} research but are promising to advance \gls{ml} applications. This includes in particular the \gls{ml} concepts transfer learning, federated learning, causal learning, and neuro-symbolic \gls{ai}.  

The concept of transfer learning \citep[e.g.][]{pan2009survey} is promising for \gls{bpm} research as \gls{bpm} software provides reference models for many standard business processes (e.g. order to cash)~\citep{dongen2007verification}, and such standard processes are similar in many organisations and organisational units.
For example, in such a homogeneous scenario, a model for predicting remaining time could be trained in an organisational unit before this unit could pass the pre-trained model on to other units that do not have enough process data to train models themselves. 
First works in \gls{bpm} research make use of transfer learning to address multiple prediction targets given an event log for comprehensive predictive business process monitoring~\citep[e.g.][]{pfeiffer2021multivariate} or to integrate additional context information for accurate discovery of declarative process models~\citep[e.g.][]{lopez2021declarative}. 

The concept of federated learning \citep[e.g.][]{yang2019federated} is worth investigating for \gls{bpm} research as processes can be distributed and go beyond the boundaries of one organisation. 
For such business processes, federated learning provides promising approaches to training shared \gls{ml} models using data from multiple owners -- while keeping all training data local and private -- to train powerful \gls{ml} models. 
Multi-organisational process optimisation could also be addressed using federated learning, and small companies with only a small amount of event data could federate to learn predictive models for common processes while keeping their data private.

The concept of causal learning (or causal \gls{ml}) is promising for \gls{bpm} research as it aids in better understanding business process improvement by formalising the data-generation process as a structural causal model and enabling to reason about the effects of changes to this process (intervention) and what would happened in hindsight (counterfactuals) \citep{kaddour2022causal}. 
Some research in process analysis~\citep[e.g.][]{bozorgi2020process} and process monitoring~\citep[e.g.][]{bozorgi2021prescriptive} already demonstrates the benefit of using causal supervised~\gls{ml}.
However, the high number of additional ideas in causal \gls{ml}, including causal generative modelling, causal explanations, causal fairness, and causal reinforcement learning \citep{kaddour2022causal}, provide promising avenues for future research.

The concept of neuro-symbolic \gls{ai} (or neuro-symbolic computation) is worth investigating for \gls{bpm} research as it provides new principles, concepts, and methods for integrating domain knowledge into \gls{ml} applications. 
In fact, neuro-symbolic \gls{ai} emerged in \gls{ai} research to overcome the challenge of integrating learning and reasoning~\citep{garcez2015neural}. 
For this purpose, neuro-symbolic \gls{ai} combines robust connectionist machines (e.g. neural networks) with sound, logical abstractions (e.g. logical rules)~\citep{garcez2022neural}.  
For example, \citet{pasquadibisceglie2021fox} do so as one of the first with a neuro-fuzzy model in predictive business process monitoring. 

\noindent\textbf{\emph{Research direction 7:}} 
According to Finding 7, \gls{ml} applications mainly focus on improving performance. 
However, a pure consideration of performance can be insufficient for successful use in practice, and future \gls{bpm} research should address the development of novel \gls{ml} applications that incorporate aspects of ethical \gls{ai}.
The field of \gls{ai} ethics, often called trustworthy \gls{ai}, has emerged in response to growing concerns about the impact of \gls{ml} applications among other issues~\citep{kazim2021high}.
\gls{ai} ethics comprises such aspects as transparency, justice, fairness, non-maleficence, responsibility, and privacy~\citep{jobin2019global,van2017responsible,feuerriegel2020fair}, and its application regularises the impact of \gls{ml} applications. 

However, despite \gls{ai} ethics' importance for \gls{ml} applications on the individual level, \gls{bpm} research has discussed \gls{ai} ethics only in theory~\citep{mendling2018machine}.
\gls{bpm} research addresses fairness-aware process mining~\citep{qafari2019fairness} and demonstrates how process mining can be used to analyse and ensure \gls{ai}'s ethical compliance~\citep{pery2022trustworthy}, but those approaches consider fairness and ethical compliance from a process-instance perspective in specific situations.
Aspects of \gls{ai} ethics on the individual level are not yet incorporated in developing \gls{ml} applications for \gls{bpm}.  

One way to address this gap could be to consider fairness in developing an \gls{ml} application predicting process outcomes, which can be robust to biases. For example, a credit application might be rejected because of discrimination in a loan application process. 
Another example refers to privacy, which should be considered in developing an \gls{ml} application for detecting deviations. Deviations may present an incorrect behaviour of a specific worker to every process analyst without restrictions.  

\noindent\textbf{\emph{Research direction 8:}} 
In accordance with Finding 8, explainability is primarily addressed in the development of \gls{ml} applications for process monitoring tasks. As explainability is useful and not restricted to process monitoring tasks, we propose as a future research direction the consideration of it in the development of novel \gls{ml} applications for other tasks along the \gls{bpm} lifecycle, such as trace clustering, to determine why certain traces are assigned to clusters. \citet{koninck2017explaining} already propose one of the first approaches in this direction.

Following \citet{koj2023anomaly}, explainability is also relevant for anomaly detection, as explanations aid in understanding why a detected instance is an anomaly.  
Organisations can also be interested to understand why an \gls{ml} model for (semi-)automatic redesign decides that a process should change.
As an additional point of this research direction, we subscribe to \citet{bauer2021expl}'s recommendation of merging \gls{xai} with domain knowledge using human-in-the-loop approaches \citep{wu2022survey,zanzotto2019human}, as first studies in the \gls{bpm} domain demonstrate~\citep[e.g.][]{barbon2018framework}.

\noindent\textbf{\emph{Research direction 9:}} 
According to Finding 9, model assessment of \gls{ml} applications is mainly done with data-based metrics. Therefore, we propose as future research direction the development of additional human-centric and economic metrics that measure \gls{bpm}-specific aspects of \gls{ml} applications, that are relevant for process stakeholders but cannot be directly obtained from the underlying data. 

One of these aspects is the effectiveness of \gls{ml} applications, that is, how useful the output of an \gls{ml} application is for process users and other process stakeholders. 
For example, in prescriptive business process monitoring, some approaches recommend the next best actions and report \gls{kpi} improvements (e.g. cost or time savings). However, these approaches do not consider how useful the recommended actions are for process users or other stakeholders \citep[e.g.][]{weinzierl2020predictiveWi}.   

Another aspect is the explainability of \gls{ml} applications, that is, how explainable \gls{ml} applications are for process users or other stakeholders. 
For predictive business process monitoring, where explainability of \gls{ml} applications is mostly addressed in \gls{bpm} research, the quality of explanation is often assessed via a demonstration of created explanations without any reference to process users or other stakeholders \citep{stierlexpbpm}. 
However, the focus of the explanations should be on the humans who execute the processes rather than those who develop them. This focus is important as developers may lack the knowledge to understand such explanations~\citep{bauer2021expl}. 
Therefore, metrics are required to measure the explainability of \gls{ml} applications for process users or other stakeholders. 

\noindent\textbf{\emph{Research direction 10:}} 
According to Finding 10, most \gls{ml} applications along the \gls{bpm} lifecycle use benchmark event logs for assessing the utility.  
However, as generalizability and transfer of findings and insights can be limited using benchmark event logs, future research should investigate new approaches to support assessing the utility of \gls{ml} applications in~\gls{bpm}. 

One approach could be elaborating methodological and technical guidance for benchmark experiments, including event logs, which represent different process types and characteristics. While there is such guidance at pre-print stages for general \gls{ml} benchmarks, including a framework and data sets~\citep[e.g.][]{romano2021pmlb}, \gls{bpm} requires such methodological approaches for developing \gls{ml} applications.

Another approach could be assessing the utility of \gls{ml} applications qualitatively or quantitatively. Involving practitioners in early-stage problem formulation and in the evaluation of an \gls{ml} application's utility would improve both model assessment and transfer to practice. For example, the utility of an \gls{ml} application for an organisation can be evaluated in case studies, like \citet{stierle2021technique} do in the context of process analysis.

\subsection{Implications}
Our literature review has implications for both research and practice. 
For research, our literature review provides a knowledge base of \gls{ml} applications structured into \gls{bpm} lifecycle phases and \gls{bpm} tasks. The phases and tasks of the knowledge base are described and defined in the view of \gls{ml} applications. With this knowledge base, we structure the discipline of \gls{ml} applications in \gls{bpm} and show how, where, and when \gls{ml} applications can be used in \gls{bpm}. While researchers from the \gls{bpm} domain can use and extend the knowledge base for their research purposes, researchers new to the field can use it as an entry point. Finally, with an interactive coding table, we give researchers a tool that supports them in analysing \gls{ml} applications in \gls{bpm} according to concepts, which are structured into the \gls{ml} model development process phases. 

We stimulate more \gls{ml} research outside the predominant \gls{bpm} lifecycle phases. In doing that, we contribute to \gls{ml} applications developed and used in the field of \gls{bpm}. Nevertheless, this review covers \gls{ml} applications from all \gls{bpm} lifecycle phases, as we found that a comparison of \gls{ml} applications across all lifecycle phases is essential, as the obtained insights from predominant phases can be transferred to less investigated phases. 

With our findings and research directions addressing the utility of \gls{ml} applications, we propose shifting the focus from prediction-performance-oriented \gls{ml} research in \gls{bpm} to more business-value-oriented \gls{ml} research. In doing that, we aim to improve the transfer of developed \gls{ml} applications, concepts, or ideas from research into practice.   

Our literature review also has implications for practice. We show practitioners various ways and potentials to develop and use \gls{ml} applications to improve business processes and ultimately create value. For example, \gls{ml} can be used in process identification to create event logs, which are then analysed with process mining applications to detect bottlenecks in business processes. As another example, \gls{ml} can be used in process monitoring to predict the outcome of running business processes. This enables process users to intervene in running business processes if the predicted process outcomes cause problems. These examples illustrate the benefits, which \gls{ml} applications from \gls{bpm} research can have for practice. 

Moreover, we recorded the existence of implementations of \gls{ml} applications in our coding table. Therefore, practitioners can use our coding table to find the source code of an \gls{ml} application that addresses a desired \gls{bpm} task. As the source code from a paper can be used as a basis for implementation, practitioners can build on existing knowledge.

\subsection{Limitations}
\label{sec:limitations}

Like all studies, this research project has limitations.
First, as with any literature review, a natural limitation is the investigated time period. Thus, derived findings may change over time and new findings can evolve. To omit this issue, we provide the interactive coding table alongside this publication. The content can be updated periodically and new findings may be derived. 

Second, the number of \gls{bpm}-related and \gls{ml}-related keywords in our search string is unbalanced, as many papers do not use the keyword \textquote{machine learning} itself but the name of a specific \gls{ml} algorithm or \gls{ml} concept. Including more specific \gls{bpm}-related keywords (e.g. the \gls{bpm} lifecycle phase process monitoring) resulted in many research papers retrieved.  

Third, because of the high number of papers examined in this review, the papers were divided among three researchers for reading and coding. 
By applying an iterative approach with discussions, as described in Section~\ref{sec:research}, we tried to ensure that the researchers had a common understanding of the concepts to be coded.
Our inter-coder reliability analysis results (\ref{sec:intercoderReliability}) confirm that the researchers used a common understanding in coding the concepts. Still, we cannot guarantee that the coding table is free of inconsistencies.

Fourth, we suppose that some papers do not report all of the concepts we are interested in even when those concepts are part of the underlying \gls{ml} application. 

Fifth, assigning each paper to a \gls{bpm} lifecycle phase was not trivial. Identifying a clear assignment to a \gls{bpm} lifecycle phase was challenging when the authors of a paper did not mention a particular phase. In addition, a few papers address more than one phase, and we assigned those papers based on the papers' primary focus.

Finally, we used the \gls{ml} model development process as a basis on which to structure \gls{ml} applications, but some \gls{ml} applications use more than one \gls{ml} model, which we considered in our coding. However, the composition of \gls{ml} models in or across \gls{ml} applications goes beyond the scope of this review. 

\newpage
\appendix
\section{Search Process Specifications}
\label{sec:searchprocessspecifications}
The search process consisted of three steps: First, we developed a \textit{search string} (\ref{sec:searchString}). Then, we conducted the collection of papers based on different search activities depending on the used databases (\ref{sec:searchactivities}). Lastly, we derived the final set of papers in the \textit{assessment phase}, applying various \textit{quality and exclusion criteria} (\ref{sec:criteria}). 

\subsection{Search String Development}
\label{sec:searchString}

The search string was constructed from two perspectives: the \textit{\gls{bpm} perspective} and the \textit{\gls{ml} perspective}.

To cover the \textit{\gls{bpm} perspective}, we used the search terms \textit{\textquote{business process management}} and \textit{\textquote{process mining}}\citep{vanderAalst2016action}, as well as \textit{\textquote{workflow management}} and \textit{\textquote{workflow mining}} for a general view. Then, we used the phases of the \gls{bpm} lifecycle by \citet{dumas2018fundamentals}, combined with the term \textit{\textquote{business}} to strictly consider business processes (e.g. \textit{\textquote{business process monitoring}} or \textit{\textquote{business process discovery}}). We also included related concepts like \textit{\textquote{process intelligence}}\citep{vanderAalst2016action}. Lastly, we combined the keywords \textit{\textquote{data-driven}} and \textit{\textquote{business process}} to include publications that address business processes from a data-driven perspective.

For creating the \gls{ml} perspective of our search string, we first screened fundamental works on \gls{ml} to get a comprehensive overview of \gls{ml} approaches. However, none of the investigated works provided an overview with a suitable degree of detail for our work. 
Consequently, we set out to develop a new overview following fundamental works. 
First, we defined levels of the overview, that is, \gls{ml} paradigm, \gls{ml} concept, and \gls{ml} algorithm group following \citet{janiesch2021machine}. Second, we initially defined the categories for the level \gls{ml} paradigm following \citet{mohri2018foundations} and refined these based on further works (see listed works under \gls{ml} paradigm in Table \ref{tab:ml-approaches}).  
Third, we initially defined the categories for the level \gls{ml} concept following \citet{goodfellow2016deep}, \citet{murphy2012machine} and \citet{ mohri2018foundations} and refined these based on further works (see listed works under \gls{ml} concept in Table \ref{tab:ml-approaches}). Fourth, we initially defined the categories for the level \gls{ml} algorithm group and the corresponding representatives for each category following \citet{mitchell.1997}, \citet{bishop2006pattern} and \citet{mohri2018foundations} and refined these based on further works (see listed works under \gls{ml} algorithm group in Table \ref{tab:ml-approaches}).

%
Further, concerning the \gls{ml} perspective of our search string, we included, where applicable keywords using alternate spellings (e.g. \textit{\textquote{semi-supervised learning}}, \textit{\textquote{semi-supervised learning}}) or keywords and their abbreviations (e.g. \textit{\textquote{support vector machine}} and \textit{\textquote{SVM}}).
For most keywords, we used the keyword as a fixed term, but for some, we used the word stem to cover a broader spectrum of papers (e.g. \textquote{regress*} to find results for \textit{\textquote{regression}} and \textit{\textquote{regressor}} among other).

We combined the search strings using Boolean operators. The keywords were combined with \textquote{OR} within one perspective, whereas two perspectives were combined with \textquote{AND} to ensure that both \gls{ml} and \gls{bpm} are covered in the results.
The search was conducted for the title of the publication, its abstract, and its keywords.

\subsection{Conducted Search Activities per Database}
\label{sec:searchactivities}
To conduct the search and retrieve the relevant documents, three different databases including a wide range of academic publications were used. 

\begin{description}
\item[Scopus] 
For the identification of all relevant records in Scopus\footnote{\url{https://www.scopus.com/search/form.uri?display=advanced}}, the search string as depicted in Figure~\ref{fig:searchstring} was used in the \textquote{Advanced document search}. After the documents were found, we refined the search by only including records in English. 
The results were then exported as a csv file including all bibliometric information.
\item[IEEE Xplore] 
For the database IEEE Xplore\footnote{\url{https://ieeexplore.ieee.org/search/advanced/command}}, the search string in Figure~\ref{fig:searchstring} was used with the \textquote{command search} function where only the data fields were updated to fit the syntax. As there is no possibility to search for a paper's title, abstract and keywords in one search, three individual searches were conducted. 
Once all records were found for one of the three searches, the items per page was set to the maximum of 100 records. All results per page were then selected using the \textquote{select all on page} function. The results and the bibliometric information were exported as a csv file. The same was done for the remaining pages until all records were exported. The search procedure was then repeated for all three data fields (title, abstract and keywords). The individual csv files were consolidated in one file and all duplicates were removed. 
\item[Web of Science]
For Web of Science \footnote{\url{https://www.webofscience.com/wos/woscc/advanced-search}} the \textquote{advanced search} was used for the search string shown in Figure~\ref{fig:searchstring}. To search for title, abstract and keywords, the \textquote{topic} field was applied. After the search was conducted, the records were filtered to only include English publications. The results were then exported with the full records as an Excel file. 
\end{description}
All individual files retrieved from the three databases were then consolidated using MS Excel to all follow the same structure. Lastly, a unique identifier was assigned to each entry of the consolidated list. This list was also used to filter duplicates and to keep track of the papers during the screening phases.

\subsection{Quality and Exclusion Criteria}
\label{sec:criteria}
In line with \citet{okoli2015guide} and \citet{vom2015standing}, we created quality and exclusion criteria for the full-text assessment of the papers, summarised in Table~\ref{tab:criteria}.

\begin{table}[ht]
\caption{Quality and exclusion criteria applied in the search process}
\centering
\scriptsize
\begin{tabular}{@{}lp{16cm}@{}}
\toprule
\textbf{ID} & \textbf{Criterion}                 \\\midrule
Q1 & The paper meets a length requirement of four pages to ensure a profound contribution.\\
Q2 & The paper is accessible to enable further screening and potential coding.\\
Q3 & The paper has been published at a conference or journal (except predatory journals).\\
Q4& The paper meets the citation of at least two citations per year on average for papers published in 2021 or earlier, while younger papers were included regardlessly. \\ \midrule
E1 & The paper does not treat a business process, e.g. social network analyses.
\\
E2 & The paper presents an \gls{ml} application for addressing a domain-specific problem, e.g. the prediction of a certain disease in the healthcare domain.
\\

E3 & The paper only considers optimisation algorithms (i.e. solvers) without a reference to optimising parameters of an \gls{ml} model, e.g. linear programming.                                                                                                 \\
E4 & The paper uses classical \gls{ai} approaches, e.g. logic programming.                                                                                                                      \\
E5 & The paper uses a clustering algorithm without an automated improvement procedure, e.g. a rule-based clustering approach for process model discovery.
\\
E6 & The paper uses \gls{ml} solely for an evaluation purpose, e.g. linear regression used in a quantitative study. 
\\
E7 & The paper does not provide a novel approach, but has a comparative nature, e.g. comparison of different approaches or is a literature review.
\\

\bottomrule

\end{tabular}
\label{tab:criteria}
\end{table}

\section{Inter-coder Reliability Analysis}
\label{sec:intercoderReliability}

Table~\ref{tab:intercoder} shows the results of the inter-coder reliability analysis for all binary and categorical dimensions. The dimensions that required free text were all checked by one coder at the end of the coding phase to ensure consistency, for which we used a random sample of 25 papers (approximately 12.5\% of the set of papers). The coding was done independently by the three coders based on the concept matrix that we developed during the coding process. 

We used three types of metrics to measure the inter-coder reliability. First, we calculated the percentage agreement representing the relative number of papers, for which all coders set the same code~\citep{lombard2002}. 
We also calculated Krippendorff's $\alpha$, which is appropriate for two or more coders and can deal with diverse types of data and missing values~\citep{krippendorff2018content}.
Lastly, we calculated Fleiss' $\kappa$ \citep{fleiss1973equivalence}, which is a common metric for three or more coders. We used R (version 4.2.1) with the \emph{DescTools} library for the calculation.
%

The percentage agreement among the coders is high for all dimensions and concepts (greater than 0.8); only one dimension (Number of data sets) is below 0.8, so a single coder coded all papers for this dimension again to ensure consistency in the coding.

The percentage agreement does not account for agreements by chance \citep{lombard2002}, so we calculated Krippendorff's $\alpha$ and Fleiss' $\kappa$, following \citet{neuendorf2017content}. Both metrics correct for the probability of agreement by chance \citep{landis1977application}.
These metrics reveal that almost all concepts are within a very good range, as most values are above 0.8, which resembles a near-perfect agreement \citep{landis1977application}, or are between 0.61 and 0.8, resembling substantial agreement \citep{landis1977application}.

We observed divergent behaviour for only three concepts: The values for \textit{Feature creation}, \textit{Self-supervised learning}, and \textit{Online/ incremental learning} are between 0.45 and 0.5. The results for \textit{Rule-based} and \textit{Other} in the algorithm group section, are not representative due to strongly imbalanced data. However, to ensure the reliability of our coding, one of the researchers checked at the end again the complete coding.    

\begin{table}[htbp!]
\centering
\caption{Inter-coder reliability analysis results}
\label{tab:intercoder}
\scriptsize
\begin{tabular}{@{}llrrr@{}}

\toprule
\textbf{Dimension}          & \textbf{}                               & \textbf{\% Agreement} & \textbf{Krippendorff's $\alpha$} & \textbf{Fleiss’ $\kappa$} \\ \midrule
\textbf{Overall}         &   & \textbf{0.954}                                \textbf{}             &                               &                        \\ \midrule
\textbf{Evaluation}    & \textbf{Overall}                        & \textbf{0.867}             &                           &                        \\
                            & Reference                               &      0.958                 &     0.785                          & 0.787                       \\
                            & Data type                               &      0.917                 &      0.884                         & 0.883                       \\
                            & Number of data sets                      & 0.625                      &       0.686                        & 0.684                       \\
                            & Implementation available              &    0.833                   &      0.774                         & 0.773                       \\ \midrule
\textbf{BPM lifecycle phase}      & \textbf{Overall}                        & \textbf{0.972}             &                               &                        \\
\end{tabular}

\label{tab:my-table}
\end{table}


\newpage

\pdfpagewidth=17in \pdfpageheight=11in

\setlength{\textwidth}{2.3\textwidth}



\section{Coding Table}
\label{sec:codingtable}

\setlength \arrayrulewidth{1pt}
\setstretch{0.85} 

\begin{tiny}

\begin{longtable}{@{}p{1.8cm}|p{1.5cm}p{0.05cm}p{2.7cm}p{1.5cm}p{1.5cm}p{0.05cm}|p{0.05cm}p{0.05cm}p{0.05cm}p{0.05cm}p{0.05cm}p{0.05cm}!{\vrule width 0.01pt}p{2.5cm}|p{1cm}p{1cm}|p{1.9cm}p{0.05cm}p{0.05cm}|p{0.05cm}p{0.05cm}p{0.05cm}p{0.05cm}p{0.05cm}p{0.05cm}!{\vrule width 0.01pt}p{0.05cm}p{0.05cm}p{0.05cm}p{0.05cm}p{0.05cm}p{0.05cm}p{0.05cm}p{0.05cm}p{0.05cm}p{0.05cm}!{\vrule width 0.01pt}p{0.05cm}p{0.05cm}p{0.05cm}p{0.05cm}p{0.05cm}p{0.05cm}p{0.05cm}p{0.05cm}p{0.05cm}p{0.05cm}|p{2.5cm}l@{}}

\caption{Coding table for ML applications in BPM}\\

\hline

\multicolumn{1}{c|}{\textbf{Paper}} & 
\multicolumn{6}{c|}{\textbf{Evaluation}} & 
\multicolumn{6}{c!{\vrule width 0.01pt}}{\textbf{BPM lifecycle phase}}&
\multicolumn{1}{c|}{\textbf{BPM task category}} &
\multicolumn{2}{c|}{\textbf{Data input}} & 
\multicolumn{3}{c|}{\textbf{Feature engineering}} &
\multicolumn{26}{c|}{\textbf{Model building}}&                   
\multicolumn{2}{c}{\textbf{Model assessment}}  
\\

\multicolumn{1}{c|}{} & 
\multicolumn{6}{c|}{} & 
\multicolumn{6}{c!{\vrule width 0.01pt}}{\textbf{}}&
\multicolumn{1}{c|}{\textbf{}} &
\multicolumn{2}{c|}{\textbf{}} & 
\multicolumn{3}{c|}{\textbf{}} &
\multicolumn{6}{c!{\vrule width 0.01pt}}{\textbf{ML paradigm}}&                   \multicolumn{10}{c!{\vrule width 0.01pt}}{\textbf{ML concept}}&
\multicolumn{10}{c|}{\textbf{ML algorithm group}} &
\multicolumn{2}{c}{\textbf{}}   
\\

                        
&\rotatebox[origin=l]{90}{Data type}\centering 
&\rotatebox[origin=l]{90}{Number of datasets}\centering
&\rotatebox[origin=l]{90}{Name of datasets}\centering
&\rotatebox[origin=l]{90}{Statistical method}\centering
&\rotatebox[origin=l]{90}{Further}\centering
&\rotatebox[origin=l]{90}{Implementation available}\centering


&\rotatebox[origin=l]{90}{Process identification}\centering 
&\rotatebox[origin=l]{90}{Process discovery}\centering 
&\rotatebox[origin=l]{90}{Process analysis}\centering
&\rotatebox[origin=l]{90}{Process redesign}\centering
&\rotatebox[origin=l]{90}{Process implementation}\centering
&\rotatebox[origin=l]{90}{Process monitoring}\centering

&\rotatebox[origin=l]{90}{BPM task category}\centering

&\rotatebox[origin=l]{90}{Dataset type}\centering 
&\rotatebox[origin=l]{90}{Rel. context information}\centering

&\rotatebox[origin=l]{90}{Feature extraction}\centering
&\rotatebox[origin=l]{90}{Feature creation}\centering
&\rotatebox[origin=l]{90}{Feature selection} \centering

&\rotatebox[origin=l]{90}{Supervised learning}\centering 
&\rotatebox[origin=l]{90}{Semi-supervised learning}\centering 
&\rotatebox[origin=l]{90}{Reinforcement learning}\centering 
&\rotatebox[origin=l]{90}{Self-supervised learning}\centering
&\rotatebox[origin=l]{90}{Unsupervised learning}\centering 
&\rotatebox[origin=l]{90}{Other}\centering

&\rotatebox[origin=l]{90}{Multi-task learning} \centering
&\rotatebox[origin=l]{90}{Active learning}\centering
&\rotatebox[origin=l]{90}{Online/ increm. learning}\centering
&\rotatebox[origin=l]{90}{Transfer learning}\centering 
&\rotatebox[origin=l]{90}{Ensemble learning}\centering 
&\rotatebox[origin=l]{90}{Deep learning}\centering
&\rotatebox[origin=l]{90}{Meta learning}\centering
&\rotatebox[origin=l]{90}{Federated learning}\centering
&\rotatebox[origin=l]{90}{Multi-view learning}\centering 
&\rotatebox[origin=l]{90}{Other}\centering

&\rotatebox[origin=l]{90}{Instance-based}\centering
&\rotatebox[origin=l]{90}{Decision tree-based}\centering
&\rotatebox[origin=l]{90}{Clustering-based}\centering
&\rotatebox[origin=l]{90}{Artificial-NN-based}\centering
&\rotatebox[origin=l]{90}{Regression-based}\centering
&\rotatebox[origin=l]{90}{Bayesian-based}\centering   
&\rotatebox[origin=l]{90}{Rule-based}\centering
&\rotatebox[origin=l]{90}{Reinf.-learn.-based}\centering
&\rotatebox[origin=l]{90}{Genetic-based}\centering 
&\rotatebox[origin=l]{90}{Other}\centering

&\rotatebox[origin=l]{90}{Metrics}\centering
&\rotatebox[origin=l]{90}{Validation strategy}\centering


\endfirsthead

\hline

\multicolumn{1}{c|}{\textbf{Paper}} & 
\multicolumn{6}{c|}{\textbf{Evaluation}} & 
\multicolumn{6}{c!{\vrule width 0.01pt}}{\textbf{BPM lifecycle phase}}&
\multicolumn{1}{c|}{\textbf{BPM task category}} &
\multicolumn{2}{c|}{\textbf{Data input}} & 
\multicolumn{3}{c|}{\textbf{Feature engineering}} &
\multicolumn{26}{c|}{\textbf{Model building}}&                   
\multicolumn{2}{c}{\textbf{Model assessment}}   \\ 

\multicolumn{1}{c|}{} & 
\multicolumn{6}{c|}{} & 
\multicolumn{6}{c!{\vrule width 0.01pt}}{\textbf{}}&
\multicolumn{1}{c|}{\textbf{}} &
\multicolumn{2}{c|}{\textbf{}} & 
\multicolumn{3}{c|}{\textbf{}} &
\multicolumn{6}{c!{\vrule width 0.01pt}}{\textbf{ML paradigm}}&                   \multicolumn{10}{c!{\vrule width 0.01pt}}{\textbf{ML concept}}&
\multicolumn{10}{c|}{\textbf{ML algorithm group}} &
\multicolumn{2}{c}{\textbf{}}   \\
\hline

\end{longtable}

\end{tiny}


\newpage
\pdfpagewidth=8.5in \pdfpageheight=11in
\setlength{\textwidth}{0.435\textwidth}

 \bibliography{cas-refs}

\begin{thebibliography}{311}
\expandafter\ifx\csname natexlab\endcsname\relax\def\natexlab#1{#1}\fi
\providecommand{\url}[1]{\texttt{#1}}
\providecommand{\href}[2]{#2}
\providecommand{\path}[1]{#1}
\providecommand{\DOIprefix}{doi:}
\providecommand{\ArXivprefix}{arXiv:}
\providecommand{\URLprefix}{URL: }
\providecommand{\Pubmedprefix}{pmid:}
\providecommand{\doi}[1]{\href{http://dx.doi.org/#1}{\path{#1}}}
\providecommand{\Pubmed}[1]{\href{pmid:#1}{\path{#1}}}
\providecommand{\bibinfo}[2]{#2}
\ifx\xfnm\relax \def\xfnm[#1]{\unskip,\space#1}\fi
\bibitem[{van~der Aa et~al.(2018)van~der Aa, Carmona~Vargas, Leopold, Mendling \& Padr{\'o}}]{van2018challenges}
\bibinfo{author}{van~der Aa, H.}, \bibinfo{author}{Carmona~Vargas, J.}, \bibinfo{author}{Leopold, H.}, \bibinfo{author}{Mendling, J.}, \& \bibinfo{author}{Padr{\'o}, L.} (\bibinfo{year}{2018}).
\newblock \bibinfo{title}{Challenges and opportunities of applying natural language processing in business process management}.
\newblock In {\it \bibinfo{booktitle}{Proceedings of the 27th International Conference on Computational Linguistics}\/} (pp. \bibinfo{pages}{2791--2801}).
\newblock \bibinfo{organization}{ACL}.
\bibitem[{van~der Aalst(2019)}]{van2019object}
\bibinfo{author}{van~der Aalst, W.} (\bibinfo{year}{2019}).
\newblock \bibinfo{title}{Object-centric process mining: Dealing with divergence and convergence in event data}.
\newblock In {\it \bibinfo{booktitle}{Proceedings of the 17th International Conference on Software Engineering and Formal Methods}\/} (pp. \bibinfo{pages}{3--25}).
\newblock \bibinfo{organization}{Springer}.
\bibitem[{van~der Aalst(2016)}]{vanderAalst2016action}
\bibinfo{author}{van~der Aalst, W. M.~P.} (\bibinfo{year}{2016}).
\newblock {\it \bibinfo{title}{{Process Mining}}\/}.
\newblock (\bibinfo{edition}{2nd} ed.).
\newblock \bibinfo{publisher}{Springer}.
\bibitem[{van~der Aalst et~al.(2011)van~der Aalst, Adriansyah, De~Medeiros, Arcieri, Baier, Blickle, Bose, Van Den~Brand, Brandtjen, Buijs et~al.}]{van.2011}
\bibinfo{author}{van~der Aalst, W. M.~P.}, \bibinfo{author}{Adriansyah, A.}, \bibinfo{author}{De~Medeiros, A. K.~A.}, \bibinfo{author}{Arcieri, F.}, \bibinfo{author}{Baier, T.}, \bibinfo{author}{Blickle, T.}, \bibinfo{author}{Bose, R. J.~C.}, \bibinfo{author}{Van Den~Brand, P.}, \bibinfo{author}{Brandtjen, R.}, \bibinfo{author}{Buijs, J.} et~al. (\bibinfo{year}{2011}).
\newblock \bibinfo{title}{Process mining manifesto}.
\newblock In {\it \bibinfo{booktitle}{{Proceedings of the 9th International Conference of Business Process Management (Workshops)}}\/} (pp. \bibinfo{pages}{169--194}).
\newblock \bibinfo{publisher}{Springer}.
\bibitem[{van~der Aalst et~al.(2017)van~der Aalst, Bichler \& Heinzl}]{van2017responsible}
\bibinfo{author}{van~der Aalst, W. M.~P.}, \bibinfo{author}{Bichler, M.}, \& \bibinfo{author}{Heinzl, A.} (\bibinfo{year}{2017}).
\newblock \bibinfo{title}{Responsible data science}.
\newblock {\it \bibinfo{journal}{Business \& Information Systems Engineering}\/},  {\it \bibinfo{volume}{59}\/}, \bibinfo{pages}{311--313}.
\bibitem[{van~der Aalst et~al.(2005)van~der Aalst, De~Medeiros \& Weijters}]{van2005genetic}
\bibinfo{author}{van~der Aalst, W. M.~P.}, \bibinfo{author}{De~Medeiros, A. K.~A.}, \& \bibinfo{author}{Weijters, A.~J.} (\bibinfo{year}{2005}).
\newblock \bibinfo{title}{Genetic process mining}.
\newblock In {\it \bibinfo{booktitle}{{Proceedings of the 26th International Conference on Application and Theory of Petri Nets}}\/} (pp. \bibinfo{pages}{48--69}).
\newblock \bibinfo{publisher}{Springer}.
\bibitem[{Afflerbach et~al.(2017)Afflerbach, Hohendorf \& Manderscheid}]{afflerbach2017design}
\bibinfo{author}{Afflerbach, P.}, \bibinfo{author}{Hohendorf, M.}, \& \bibinfo{author}{Manderscheid, J.} (\bibinfo{year}{2017}).
\newblock \bibinfo{title}{{Design it like Darwin -- A value-based application of evolutionary algorithms for proper and unambiguous business process redesign}}.
\newblock {\it \bibinfo{journal}{Information Systems Frontiers}\/},  {\it \bibinfo{volume}{19}\/}, \bibinfo{pages}{1101--1121}.
\bibitem[{Alpaydin(2014)}]{Alpaydin2014}
\bibinfo{author}{Alpaydin, E.} (\bibinfo{year}{2014}).
\newblock {\it \bibinfo{title}{{Introduction to Machine Learning}}\/}.
\newblock (\bibinfo{edition}{3rd} ed.).
\newblock \bibinfo{publisher}{MIT Press}.
\bibitem[{Alves et~al.(2022)Alves, Barbieri, Stroeh, Peres \& Madeira}]{alves2022context}
\bibinfo{author}{Alves, R.~M.}, \bibinfo{author}{Barbieri, L.}, \bibinfo{author}{Stroeh, K.}, \bibinfo{author}{Peres, S.~M.}, \& \bibinfo{author}{Madeira, E. R.~M.} (\bibinfo{year}{2022}).
\newblock \bibinfo{title}{{Context-aware completion time prediction for business process monitoring}}.
\newblock In {\it \bibinfo{booktitle}{{Proceedings of the 2022 World Conference on Information Systems and Technologies}}\/} (pp. \bibinfo{pages}{355--365}).
\newblock \bibinfo{publisher}{Springer}.
\bibitem[{Appice et~al.(2019)Appice, Di~Mauro \& Malerba}]{appice2019leveraging}
\bibinfo{author}{Appice, A.}, \bibinfo{author}{Di~Mauro, N.}, \& \bibinfo{author}{Malerba, D.} (\bibinfo{year}{2019}).
\newblock \bibinfo{title}{Leveraging shallow machine learning to predict business process behavior}.
\newblock In {\it \bibinfo{booktitle}{{Proceedings of the 2019 IEEE International Conference on Services Computing}}\/} (pp. \bibinfo{pages}{184--188}).
\newblock \bibinfo{publisher}{IEEE}.
\bibitem[{Appice \& Malerba(2015)}]{appice2015co}
\bibinfo{author}{Appice, A.}, \& \bibinfo{author}{Malerba, D.} (\bibinfo{year}{2015}).
\newblock \bibinfo{title}{A co-training strategy for multiple view clustering in process mining}.
\newblock {\it \bibinfo{journal}{{IEEE Transactions on Services Computing}}\/},  {\it \bibinfo{volume}{9}\/}, \bibinfo{pages}{832--845}.
\bibitem[{Bae et~al.(2014)Bae, Lee \& Moon}]{bae2014planning}
\bibinfo{author}{Bae, H.}, \bibinfo{author}{Lee, S.}, \& \bibinfo{author}{Moon, I.} (\bibinfo{year}{2014}).
\newblock \bibinfo{title}{Planning of business process execution in business process management environments}.
\newblock {\it \bibinfo{journal}{Information Sciences}\/},  {\it \bibinfo{volume}{268}\/}, \bibinfo{pages}{357--369}.
\bibitem[{Baier et~al.(2020)Baier, Reimold \& K{\"u}hl}]{baier2020handling}
\bibinfo{author}{Baier, L.}, \bibinfo{author}{Reimold, J.}, \& \bibinfo{author}{K{\"u}hl, N.} (\bibinfo{year}{2020}).
\newblock \bibinfo{title}{Handling concept drift for predictions in business process mining}.
\newblock In {\it \bibinfo{booktitle}{{Proceedings of the 22nd IEEE Conference on Business Informatics}}\/} (pp. \bibinfo{pages}{76--83}).
\newblock \bibinfo{publisher}{IEEE}.
\bibitem[{Bauer et~al.(2021)Bauer, Hinz, van~der Aalst \& Weinhardt}]{bauer2021expl}
\bibinfo{author}{Bauer, K.}, \bibinfo{author}{Hinz, O.}, \bibinfo{author}{van~der Aalst, W.}, \& \bibinfo{author}{Weinhardt, C.} (\bibinfo{year}{2021}).
\newblock \bibinfo{title}{{Expl(AI)n it to me -- Explainable AI and information systems research}}.
\newblock {\it \bibinfo{journal}{Business \& Information Systems Engineering}\/},  {\it \bibinfo{volume}{63}\/}, \bibinfo{pages}{79--82}.
\bibitem[{Bazhenova et~al.(2016)Bazhenova, Buelow \& Weske}]{bazhenova2016discovering}
\bibinfo{author}{Bazhenova, E.}, \bibinfo{author}{Buelow, S.}, \& \bibinfo{author}{Weske, M.} (\bibinfo{year}{2016}).
\newblock \bibinfo{title}{Discovering decision models from event logs}.
\newblock In {\it \bibinfo{booktitle}{{Proceedings of the 19th International Conference on Business Information Systems}}\/} (pp. \bibinfo{pages}{237--251}).
\newblock \bibinfo{publisher}{Springer}.
\bibitem[{Beese et~al.(2019)Beese, Haki, Aier \& Winter}]{beese2019simulation}
\bibinfo{author}{Beese, J.}, \bibinfo{author}{Haki, M.~K.}, \bibinfo{author}{Aier, S.}, \& \bibinfo{author}{Winter, R.} (\bibinfo{year}{2019}).
\newblock \bibinfo{title}{Simulation-based research in information systems}.
\newblock {\it \bibinfo{journal}{Business \& Information Systems Engineering}\/},  {\it \bibinfo{volume}{61}\/}, \bibinfo{pages}{503--521}.
\bibitem[{Berkenstadt et~al.(2020)Berkenstadt, Gal, Senderovich, Shraga \& Weidlich}]{Berkenstadt2020}
\bibinfo{author}{Berkenstadt, G.}, \bibinfo{author}{Gal, A.}, \bibinfo{author}{Senderovich, A.}, \bibinfo{author}{Shraga, R.}, \& \bibinfo{author}{Weidlich, M.} (\bibinfo{year}{2020}).
\newblock \bibinfo{title}{Queueing inference for process performance analysis with missing life-cycle data}.
\newblock In {\it \bibinfo{booktitle}{{Proceedings of the 2nd International Conference on Process Mining}}\/} (pp. \bibinfo{pages}{57--64}).
\newblock \bibinfo{publisher}{IEEE}.
\newblock \DOIprefix\doi{10.1109/ICPM49681.2020.00019}.
\bibitem[{Bernard \& Andritsos(2019)}]{bernard2019accurate}
\bibinfo{author}{Bernard, G.}, \& \bibinfo{author}{Andritsos, P.} (\bibinfo{year}{2019}).
\newblock \bibinfo{title}{Accurate and transparent path prediction using process mining}.
\newblock In {\it \bibinfo{booktitle}{{Proceedings of the 23rd European Conference on Advances in Databases and Information Systems}}\/} (pp. \bibinfo{pages}{235--250}).
\newblock \bibinfo{publisher}{Springer}.
\bibitem[{Bevacqua et~al.(2013)Bevacqua, Carnuccio, Folino, Guarascio \& Pontieri}]{bevacqua2013data}
\bibinfo{author}{Bevacqua, A.}, \bibinfo{author}{Carnuccio, M.}, \bibinfo{author}{Folino, F.}, \bibinfo{author}{Guarascio, M.}, \& \bibinfo{author}{Pontieri, L.} (\bibinfo{year}{2013}).
\newblock \bibinfo{title}{A data-driven prediction framework for analyzing and monitoring business process performances}.
\newblock In {\it \bibinfo{booktitle}{{Proceedings of the 15th International Conference on Enterprise Information Systems}}\/} (pp. \bibinfo{pages}{100--117}).
\newblock \bibinfo{publisher}{Springer}.
\bibitem[{Beverungen et~al.(2021)Beverungen, Buijs, Becker, Di~Ciccio, van~der Aalst, Bartelheimer, vom Brocke, Comuzzi, Kraume, Leopold, Matzner, Mendling, Ogonek, Post, Resinas, Revoredo, del Río-Ortega, La~Rosa, Santoro, Solti, Song, Stein, Stierle \& Wolf}]{beverungen2021seven}
\bibinfo{author}{Beverungen, D.}, \bibinfo{author}{Buijs, J.}, \bibinfo{author}{Becker, J.}, \bibinfo{author}{Di~Ciccio, C.}, \bibinfo{author}{van~der Aalst, W. M.~P.}, \bibinfo{author}{Bartelheimer, C.}, \bibinfo{author}{vom Brocke, J.}, \bibinfo{author}{Comuzzi, M.}, \bibinfo{author}{Kraume, K.}, \bibinfo{author}{Leopold, H.}, \bibinfo{author}{Matzner, M.}, \bibinfo{author}{Mendling, J.}, \bibinfo{author}{Ogonek, N.}, \bibinfo{author}{Post, T.}, \bibinfo{author}{Resinas, M.}, \bibinfo{author}{Revoredo, K.}, \bibinfo{author}{del Río-Ortega, A.}, \bibinfo{author}{La~Rosa, M.}, \bibinfo{author}{Santoro, F.~M.}, \bibinfo{author}{Solti, A.}, \bibinfo{author}{Song, M.}, \bibinfo{author}{Stein, A.}, \bibinfo{author}{Stierle, M.}, \& \bibinfo{author}{Wolf, V.} (\bibinfo{year}{2021}).
\newblock \bibinfo{title}{Seven paradoxes of business process management in a hyper-connected world}.
\newblock {\it \bibinfo{journal}{Business \& Information Systems Engineering}\/},  {\it \bibinfo{volume}{63}\/}, \bibinfo{pages}{145--156}. \DOIprefix\doi{10.1007/s12599-020-00646-z}.
\bibitem[{Bezerra et~al.(2009)Bezerra, Wainer \& van~der Aalst}]{bezerra2009anomaly}
\bibinfo{author}{Bezerra, F.}, \bibinfo{author}{Wainer, J.}, \& \bibinfo{author}{van~der Aalst, W. M.~P.} (\bibinfo{year}{2009}).
\newblock \bibinfo{title}{Anomaly detection using process mining}.
\newblock In {\it \bibinfo{booktitle}{{Proceedings of the 14th International Conference on Enterprise, Business-Process and Information Systems Modeling}}\/} (pp. \bibinfo{pages}{149--161}).
\newblock \bibinfo{publisher}{Springer}.
\bibitem[{Bishop(2006)}]{bishop2006pattern}
\bibinfo{author}{Bishop, C.~M.} (\bibinfo{year}{2006}).
\newblock {\it \bibinfo{title}{{Pattern Recognition and Machine Learning}}\/}.
\newblock \bibinfo{publisher}{Springer}.
\bibitem[{Blum(1998)}]{blum1998line}
\bibinfo{author}{Blum, A.} (\bibinfo{year}{1998}).
\newblock \bibinfo{title}{On-line algorithms in machine learning}.
\newblock In {\it \bibinfo{booktitle}{{Online Algorithms -- State of the Art}}\/} (pp. \bibinfo{pages}{306--325}).
\newblock \bibinfo{publisher}{Springer}.
\bibitem[{Boltenhagen et~al.(2019)Boltenhagen, Chatain \& Carmona}]{boltenhagen2019generalized}
\bibinfo{author}{Boltenhagen, M.}, \bibinfo{author}{Chatain, T.}, \& \bibinfo{author}{Carmona, J.} (\bibinfo{year}{2019}).
\newblock \bibinfo{title}{Generalized alignment-based trace clustering of process behavior}.
\newblock In {\it \bibinfo{booktitle}{Proceedings of the 40th International Conference on Application and Theory of Petri Nets and Concurrency}\/} (pp. \bibinfo{pages}{237--257}).
\newblock \bibinfo{organization}{Springer}.
\bibitem[{Borkowski et~al.(2019)Borkowski, Fdhila, Nardelli, Rinderle-Ma \& Schulte}]{borkowski2019event}
\bibinfo{author}{Borkowski, M.}, \bibinfo{author}{Fdhila, W.}, \bibinfo{author}{Nardelli, M.}, \bibinfo{author}{Rinderle-Ma, S.}, \& \bibinfo{author}{Schulte, S.} (\bibinfo{year}{2019}).
\newblock \bibinfo{title}{Event-based failure prediction in distributed business processes}.
\newblock {\it \bibinfo{journal}{Information Systems}\/},  {\it \bibinfo{volume}{81}\/}, \bibinfo{pages}{220--235}.
\bibitem[{Bose \& van~der Aalst(2009)}]{bose2009context}
\bibinfo{author}{Bose, R. J.~C.}, \& \bibinfo{author}{van~der Aalst, W. M.~P.} (\bibinfo{year}{2009}).
\newblock \bibinfo{title}{Context aware trace clustering: Towards improving process mining results}.
\newblock In {\it \bibinfo{booktitle}{{Proceedings of the 2009 SIAM International Conference on Data Mining}}\/} (pp. \bibinfo{pages}{401--412}).
\newblock \bibinfo{publisher}{SIAM}.
\bibitem[{Bose et~al.(2013)Bose, Mans \& van~der Aalst}]{bose2013wanna}
\bibinfo{author}{Bose, R. J.~C.}, \bibinfo{author}{Mans, R.~S.}, \& \bibinfo{author}{van~der Aalst, W. M.~P.} (\bibinfo{year}{2013}).
\newblock \bibinfo{title}{Wanna improve process mining results?}
\newblock In {\it \bibinfo{booktitle}{{Proceedings of the 2013 IEEE Symposium on Computational Intelligence and Data Mining}}\/} (pp. \bibinfo{pages}{127--134}).
\newblock \bibinfo{publisher}{IEEE}.
\bibitem[{Bozorgi et~al.(2020)Bozorgi, Teinemaa, Dumas, La~Rosa \& Polyvyanyy}]{bozorgi2020process}
\bibinfo{author}{Bozorgi, Z.~D.}, \bibinfo{author}{Teinemaa, I.}, \bibinfo{author}{Dumas, M.}, \bibinfo{author}{La~Rosa, M.}, \& \bibinfo{author}{Polyvyanyy, A.} (\bibinfo{year}{2020}).
\newblock \bibinfo{title}{Process mining meets causal machine learning: Discovering causal rules from event logs}.
\newblock In {\it \bibinfo{booktitle}{{Proceedings of the 2nd International Conference on Process Mining}}\/} (pp. \bibinfo{pages}{129--136}).
\newblock \bibinfo{publisher}{IEEE}.
\bibitem[{Bozorgi et~al.(2021)Bozorgi, Teinemaa, Dumas, La~Rosa \& Polyvyanyy}]{bozorgi2021prescriptive}
\bibinfo{author}{Bozorgi, Z.~D.}, \bibinfo{author}{Teinemaa, I.}, \bibinfo{author}{Dumas, M.}, \bibinfo{author}{La~Rosa, M.}, \& \bibinfo{author}{Polyvyanyy, A.} (\bibinfo{year}{2021}).
\newblock \bibinfo{title}{Prescriptive process monitoring for cost-aware cycle time reduction}.
\newblock In {\it \bibinfo{booktitle}{{Proceedings of the 3rd International Conference on Process Mining}}\/} (pp. \bibinfo{pages}{96--103}).
\newblock \bibinfo{publisher}{IEEE}.
\bibitem[{Breiman et~al.(1984)Breiman, Friedman, Olshen \& Stone}]{breiman1984classification}
\bibinfo{author}{Breiman, L.}, \bibinfo{author}{Friedman, J.~H.}, \bibinfo{author}{Olshen, R.}, \& \bibinfo{author}{Stone, C.} (\bibinfo{year}{1984}).
\newblock {\it \bibinfo{title}{{Classification and Regression Trees}}\/}.
\newblock \bibinfo{publisher}{Wadsworth.}
\bibitem[{Breuker et~al.(2016)Breuker, Matzner, Delfmann \& Becker}]{breuker2016comprehensible}
\bibinfo{author}{Breuker, D.}, \bibinfo{author}{Matzner, M.}, \bibinfo{author}{Delfmann, P.}, \& \bibinfo{author}{Becker, J.} (\bibinfo{year}{2016}).
\newblock \bibinfo{title}{Comprehensible predictive models for business processes}.
\newblock {\it \bibinfo{journal}{MIS Quarterly}\/},  {\it \bibinfo{volume}{40}\/}, \bibinfo{pages}{1009--1034}. \DOIprefix\doi{10.25300/MISQ/2016/40.4.10}.
\bibitem[{vom Brocke et~al.(2015)vom Brocke, Simons, Riemer, Niehaves, Plattfaut \& Cleven}]{vom2015standing}
\bibinfo{author}{vom Brocke, J.}, \bibinfo{author}{Simons, A.}, \bibinfo{author}{Riemer, K.}, \bibinfo{author}{Niehaves, B.}, \bibinfo{author}{Plattfaut, R.}, \& \bibinfo{author}{Cleven, A.} (\bibinfo{year}{2015}).
\newblock \bibinfo{title}{Standing on the shoulders of giants: Challenges and recommendations of literature search in information systems research}.
\newblock {\it \bibinfo{journal}{{Communications of the Association for Information Systems}}\/},  {\it \bibinfo{volume}{37}\/}, \bibinfo{pages}{205 – 224}.
\bibitem[{Brunk et~al.(2021)Brunk, Stierle, Papke, Revoredo, Matzner \& Becker}]{brunk2021cause}
\bibinfo{author}{Brunk, J.}, \bibinfo{author}{Stierle, M.}, \bibinfo{author}{Papke, L.}, \bibinfo{author}{Revoredo, K.}, \bibinfo{author}{Matzner, M.}, \& \bibinfo{author}{Becker, J.} (\bibinfo{year}{2021}).
\newblock \bibinfo{title}{Cause vs. effect in context-sensitive prediction of business process instances}.
\newblock {\it \bibinfo{journal}{Information Systems}\/},  {\it \bibinfo{volume}{95}\/}, \bibinfo{pages}{101635}.
\bibitem[{Brynjolfsson \& McAfee(2017)}]{Brynjolfsson2017}
\bibinfo{author}{Brynjolfsson, E.}, \& \bibinfo{author}{McAfee, A.} (\bibinfo{year}{2017}).
\newblock \bibinfo{title}{{The business of artificial intelligence}}.
\newblock \URLprefix \url{https://hbr.org/2017/07/the-business-of-artificial-intelligence}.
\bibitem[{Buijs et~al.(2012)Buijs, van Dongen \& van~der Aalst}]{buijs2012genetic}
\bibinfo{author}{Buijs, J.}, \bibinfo{author}{van Dongen, B.~F.}, \& \bibinfo{author}{van~der Aalst, W. M.~P.} (\bibinfo{year}{2012}).
\newblock \bibinfo{title}{A genetic algorithm for discovering process trees}.
\newblock In {\it \bibinfo{booktitle}{{Proceedings of the 2012 IEEE Congress on Evolutionary Computation}}\/} (pp. \bibinfo{pages}{1--8}).
\newblock \bibinfo{publisher}{IEEE}.
\bibitem[{Buijs et~al.(2013)Buijs, van Dongen \& van~der Aalst}]{buijs2013mining}
\bibinfo{author}{Buijs, J.}, \bibinfo{author}{van Dongen, B.~F.}, \& \bibinfo{author}{van~der Aalst, W. M.~P.} (\bibinfo{year}{2013}).
\newblock \bibinfo{title}{Mining configurable process models from collections of event logs}.
\newblock In {\it \bibinfo{booktitle}{{Proceedings of the 11th International Conference on Business Process Management}}\/} (pp. \bibinfo{pages}{33--48}).
\newblock \bibinfo{publisher}{Springer}.
\bibitem[{Camargo et~al.(2019)Camargo, Dumas \& Gonz{\'a}lez-Rojas}]{camargo2019learning}
\bibinfo{author}{Camargo, M.}, \bibinfo{author}{Dumas, M.}, \& \bibinfo{author}{Gonz{\'a}lez-Rojas, O.} (\bibinfo{year}{2019}).
\newblock \bibinfo{title}{Learning accurate {LSTM} models of business processes}.
\newblock In {\it \bibinfo{booktitle}{{Proceedings of the 17th International Conference on Business Process Management}}\/} (pp. \bibinfo{pages}{286--302}).
\newblock \bibinfo{publisher}{Springer}.
\bibitem[{Camargo et~al.(2022)Camargo, Dumas \& Gonz{\'a}lez-Rojas}]{camargo2022learning}
\bibinfo{author}{Camargo, M.}, \bibinfo{author}{Dumas, M.}, \& \bibinfo{author}{Gonz{\'a}lez-Rojas, O.} (\bibinfo{year}{2022}).
\newblock \bibinfo{title}{Learning accurate business process simulation models from event logs via automated process discovery and deep learning}.
\newblock In {\it \bibinfo{booktitle}{{Proceedings of the 34th International Conference on Advanced Information Systems Engineering}}\/} (pp. \bibinfo{pages}{55--71}).
\newblock \bibinfo{publisher}{Springer}.
\bibitem[{Cao et~al.(2006)Cao, Estert, Qian \& Zhou}]{cao2006density}
\bibinfo{author}{Cao, F.}, \bibinfo{author}{Estert, M.}, \bibinfo{author}{Qian, W.}, \& \bibinfo{author}{Zhou, A.} (\bibinfo{year}{2006}).
\newblock \bibinfo{title}{Density-based clustering over an evolving data stream with noise}.
\newblock In {\it \bibinfo{booktitle}{{Proceedings of the 2006 SIAM International Conference on Data Mining}}\/} (pp. \bibinfo{pages}{328--339}).
\newblock \bibinfo{publisher}{SIAM}.
\bibitem[{Ceravolo et~al.(2022)Ceravolo, Damiani, Schepis \& Tavares}]{ceravolo2022real}
\bibinfo{author}{Ceravolo, P.}, \bibinfo{author}{Damiani, E.}, \bibinfo{author}{Schepis, E.~F.}, \& \bibinfo{author}{Tavares, G.~M.} (\bibinfo{year}{2022}).
\newblock \bibinfo{title}{Real-time probing of control-flow and data-flow in event logs}.
\newblock {\it \bibinfo{journal}{Procedia Computer Science}\/},  {\it \bibinfo{volume}{197}\/}, \bibinfo{pages}{751--758}.
\bibitem[{Chesani et~al.(2009)Chesani, Lamma, Mello, Montali, Riguzzi \& Storari}]{chesani2009exploiting}
\bibinfo{author}{Chesani, F.}, \bibinfo{author}{Lamma, E.}, \bibinfo{author}{Mello, P.}, \bibinfo{author}{Montali, M.}, \bibinfo{author}{Riguzzi, F.}, \& \bibinfo{author}{Storari, S.} (\bibinfo{year}{2009}).
\newblock \bibinfo{title}{Exploiting inductive logic programming techniques for declarative process mining}.
\newblock {\it \bibinfo{journal}{Transactions on Petri Nets and Other Models of Concurrency II: Special Issue on Concurrency in Process-Aware Information Systems}\/},  (pp. \bibinfo{pages}{278--295}).
\bibitem[{Ciasullo et~al.(2018)Ciasullo, Fenza, Loia, Orciuoli, Troisi \& Herrera-Viedma}]{ciasullo2018business}
\bibinfo{author}{Ciasullo, M.~V.}, \bibinfo{author}{Fenza, G.}, \bibinfo{author}{Loia, V.}, \bibinfo{author}{Orciuoli, F.}, \bibinfo{author}{Troisi, O.}, \& \bibinfo{author}{Herrera-Viedma, E.} (\bibinfo{year}{2018}).
\newblock \bibinfo{title}{Business process outsourcing enhanced by fuzzy linguistic consensus model}.
\newblock {\it \bibinfo{journal}{Applied Soft Computing}\/},  {\it \bibinfo{volume}{64}\/}, \bibinfo{pages}{436--444}.
\bibitem[{Cohen(1995)}]{COHEN1995115}
\bibinfo{author}{Cohen, W.~W.} (\bibinfo{year}{1995}).
\newblock \bibinfo{title}{Fast effective rule induction}.
\newblock In {\it \bibinfo{booktitle}{{Proceedings of the 12th International Conference on Machine Learning}}\/} (pp. \bibinfo{pages}{115--123}).
\newblock \bibinfo{publisher}{Elsevier}.
\newblock \DOIprefix\doi{10.1016/B978-1-55860-377-6.50023-2}.
\bibitem[{Conforti et~al.(2015)Conforti, de~Leoni, La~Rosa, van~der Aalst \& ter Hofstede}]{conforti2015recommendation}
\bibinfo{author}{Conforti, R.}, \bibinfo{author}{de~Leoni, M.}, \bibinfo{author}{La~Rosa, M.}, \bibinfo{author}{van~der Aalst, W. M.~P.}, \& \bibinfo{author}{ter Hofstede, A.~H.} (\bibinfo{year}{2015}).
\newblock \bibinfo{title}{A recommendation system for predicting risks across multiple business process instances}.
\newblock {\it \bibinfo{journal}{Decision Support Systems}\/},  {\it \bibinfo{volume}{69}\/}, \bibinfo{pages}{1--19}.
\bibitem[{Cooper(1988)}]{cooper1988organizing}
\bibinfo{author}{Cooper, H.~M.} (\bibinfo{year}{1988}).
\newblock \bibinfo{title}{Organizing knowledge syntheses: A taxonomy of literature reviews}.
\newblock {\it \bibinfo{journal}{Knowledge in Society}\/},  {\it \bibinfo{volume}{1}\/}, \bibinfo{pages}{104--126}.
\bibitem[{Cortes \& Vapnik(1995)}]{vapnik1995support}
\bibinfo{author}{Cortes, C.}, \& \bibinfo{author}{Vapnik, V.} (\bibinfo{year}{1995}).
\newblock \bibinfo{title}{Support-vector networks}.
\newblock {\it \bibinfo{journal}{Machine Learning}\/},  {\it \bibinfo{volume}{20}\/}, \bibinfo{pages}{273--297}. \DOIprefix\doi{10.1007/BF00994018}.
\bibitem[{Cuzzocrea et~al.(2015)Cuzzocrea, Folino, Guarascio \& Pontieri}]{cuzzocrea2015multi}
\bibinfo{author}{Cuzzocrea, A.}, \bibinfo{author}{Folino, F.}, \bibinfo{author}{Guarascio, M.}, \& \bibinfo{author}{Pontieri, L.} (\bibinfo{year}{2015}).
\newblock \bibinfo{title}{A multi-view learning approach to the discovery of deviant process instances}.
\newblock In {\it \bibinfo{booktitle}{{Proceedings of the 2015 OTM Confederated International Conferences "On the Move to Meaningful Internet Systems"}}\/} (pp. \bibinfo{pages}{146--165}).
\newblock \bibinfo{publisher}{Springer}.
\bibitem[{Cuzzocrea et~al.(2016{\natexlab{a}})Cuzzocrea, Folino, Guarascio \& Pontieri}]{cuzzocrea2016multi}
\bibinfo{author}{Cuzzocrea, A.}, \bibinfo{author}{Folino, F.}, \bibinfo{author}{Guarascio, M.}, \& \bibinfo{author}{Pontieri, L.} (\bibinfo{year}{2016}{\natexlab{a}}).
\newblock \bibinfo{title}{A multi-view multi-dimensional ensemble learning approach to mining business process deviances}.
\newblock In {\it \bibinfo{booktitle}{{Proceedings of the 2016 International Joint Conference on Neural Networks}}\/} (pp. \bibinfo{pages}{3809--3816}).
\newblock \bibinfo{publisher}{IEEE}.
\bibitem[{Cuzzocrea et~al.(2016{\natexlab{b}})Cuzzocrea, Folino, Guarascio \& Pontieri}]{cuzzocrea2016robust}
\bibinfo{author}{Cuzzocrea, A.}, \bibinfo{author}{Folino, F.}, \bibinfo{author}{Guarascio, M.}, \& \bibinfo{author}{Pontieri, L.} (\bibinfo{year}{2016}{\natexlab{b}}).
\newblock \bibinfo{title}{A robust and versatile multi-view learning framework for the detection of deviant business process instances}.
\newblock {\it \bibinfo{journal}{International Journal of Cooperative Information Systems}\/},  {\it \bibinfo{volume}{25}\/}, \bibinfo{pages}{1740003}.
\bibitem[{Cuzzocrea et~al.(2019)Cuzzocrea, Folino, Guarascio \& Pontieri}]{cuzzocrea2019predictive}
\bibinfo{author}{Cuzzocrea, A.}, \bibinfo{author}{Folino, F.}, \bibinfo{author}{Guarascio, M.}, \& \bibinfo{author}{Pontieri, L.} (\bibinfo{year}{2019}).
\newblock \bibinfo{title}{Predictive monitoring of temporally-aggregated performance indicators of business processes against low-level streaming events}.
\newblock {\it \bibinfo{journal}{Information Systems}\/},  {\it \bibinfo{volume}{81}\/}, \bibinfo{pages}{236--266}.
\bibitem[{De~Jong et~al.(1993)De~Jong, Spears \& Gordon}]{de1993using}
\bibinfo{author}{De~Jong, K.~A.}, \bibinfo{author}{Spears, W.~M.}, \& \bibinfo{author}{Gordon, D.~F.} (\bibinfo{year}{1993}).
\newblock \bibinfo{title}{Using genetic algorithms for concept learning}.
\newblock {\it \bibinfo{journal}{{Machine Learning}}\/},  {\it \bibinfo{volume}{13}\/}, \bibinfo{pages}{161--188}.
\bibitem[{De~Koninck et~al.(2018)De~Koninck, vanden Broucke \& De~Weerdt}]{koninck2018act2vec}
\bibinfo{author}{De~Koninck, P.}, \bibinfo{author}{vanden Broucke, S.}, \& \bibinfo{author}{De~Weerdt, J.} (\bibinfo{year}{2018}).
\newblock \bibinfo{title}{{Act2vec, trace2vec, log2vec, and model2vec: Representation learning for business processes}}.
\newblock In {\it \bibinfo{booktitle}{{Proceedings of the 16th International Conference on Business Process Management}}\/} (pp. \bibinfo{pages}{305--321}).
\newblock \bibinfo{publisher}{Springer}.
\bibitem[{De~Koninck \& De~Weerdt(2019)}]{koninck2019scalable}
\bibinfo{author}{De~Koninck, P.}, \& \bibinfo{author}{De~Weerdt, J.} (\bibinfo{year}{2019}).
\newblock \bibinfo{title}{Scalable mixed-paradigm trace clustering using super-instances}.
\newblock In {\it \bibinfo{booktitle}{{Proceedings of the 1st International Conference on Process Mining}}\/} (pp. \bibinfo{pages}{17--24}).
\newblock \bibinfo{publisher}{IEEE}.
\bibitem[{De~Koninck et~al.(2017{\natexlab{a}})De~Koninck, De~Weerdt \& vanden Broucke}]{koninck2017explaining}
\bibinfo{author}{De~Koninck, P.}, \bibinfo{author}{De~Weerdt, J.}, \& \bibinfo{author}{vanden Broucke, S.} (\bibinfo{year}{2017}{\natexlab{a}}).
\newblock \bibinfo{title}{Explaining clusterings of process instances}.
\newblock {\it \bibinfo{journal}{{Data Mining and Knowledge Discovery}}\/},  {\it \bibinfo{volume}{31}\/}, \bibinfo{pages}{774--808}.
\bibitem[{De~Koninck et~al.(2017{\natexlab{b}})De~Koninck, Nelissen, Baesens, Snoeck \& De~Weerdt}]{koninck2017approach}
\bibinfo{author}{De~Koninck, P.}, \bibinfo{author}{Nelissen, K.}, \bibinfo{author}{Baesens, B.}, \bibinfo{author}{Snoeck, M.}, \& \bibinfo{author}{De~Weerdt, J.} (\bibinfo{year}{2017}{\natexlab{b}}).
\newblock \bibinfo{title}{An approach for incorporating expert knowledge in trace clustering}.
\newblock In {\it \bibinfo{booktitle}{{Proceedings of the 29th International Conference on Advanced Information Systems Engineering}}\/} (pp. \bibinfo{pages}{561--576}).
\newblock \bibinfo{publisher}{Springer}.
\bibitem[{De~Koninck et~al.(2021)De~Koninck, Nelissen, Baesens, Snoeck \& De~Weerdt}]{koninck2021expert}
\bibinfo{author}{De~Koninck, P.}, \bibinfo{author}{Nelissen, K.}, \bibinfo{author}{Baesens, B.}, \bibinfo{author}{Snoeck, M.}, \& \bibinfo{author}{De~Weerdt, J.} (\bibinfo{year}{2021}).
\newblock \bibinfo{title}{Expert-driven trace clustering with instance-level constraints}.
\newblock {\it \bibinfo{journal}{Knowledge and Information Systems}\/},  {\it \bibinfo{volume}{63}\/}, \bibinfo{pages}{1197--1220}.
\bibitem[{De~Maio et~al.(2016)De~Maio, Fenza, Loia, Orciuoli \& Herrera-Viedma}]{de2016framework}
\bibinfo{author}{De~Maio, C.}, \bibinfo{author}{Fenza, G.}, \bibinfo{author}{Loia, V.}, \bibinfo{author}{Orciuoli, F.}, \& \bibinfo{author}{Herrera-Viedma, E.} (\bibinfo{year}{2016}).
\newblock \bibinfo{title}{A framework for context-aware heterogeneous group decision making in business processes}.
\newblock {\it \bibinfo{journal}{Knowledge-Based Systems}\/},  {\it \bibinfo{volume}{102}\/}, \bibinfo{pages}{39--50}.
\bibitem[{De~Morais \& Kazan(2014)}]{DeMorais2014}
\bibinfo{author}{De~Morais, R.~M.}, \& \bibinfo{author}{Kazan, S.} (\bibinfo{year}{2014}).
\newblock \bibinfo{title}{An analysis of {BPM} lifecycles: From a literature review to a framework proposal}.
\newblock {\it \bibinfo{journal}{Business Process Management Journal}\/},  {\it \bibinfo{volume}{20}\/}, \bibinfo{pages}{1463--7154}. \DOIprefix\doi{10.1108/BPMJ-03-2013-0035}.
\bibitem[{De~Weerdt et~al.(2012)De~Weerdt, vanden Broucke, Vanthienen \& Baesens}]{weerdt2012leveraging}
\bibinfo{author}{De~Weerdt, J.}, \bibinfo{author}{vanden Broucke, S.}, \bibinfo{author}{Vanthienen, J.}, \& \bibinfo{author}{Baesens, B.} (\bibinfo{year}{2012}).
\newblock \bibinfo{title}{Leveraging process discovery with trace clustering and text mining for intelligent analysis of incident management processes}.
\newblock In {\it \bibinfo{booktitle}{{Proceedings of the 2012 IEEE Congress on Evolutionary Computation}}\/} (pp. \bibinfo{pages}{1--8}).
\newblock \bibinfo{publisher}{IEEE}.
\bibitem[{De~Weerdt et~al.(2013)De~Weerdt, vanden Broucke, Vanthienen \& Baesens}]{weerdt2013active}
\bibinfo{author}{De~Weerdt, J.}, \bibinfo{author}{vanden Broucke, S.}, \bibinfo{author}{Vanthienen, J.}, \& \bibinfo{author}{Baesens, B.} (\bibinfo{year}{2013}).
\newblock \bibinfo{title}{Active trace clustering for improved process discovery}.
\newblock {\it \bibinfo{journal}{IEEE Transactions on Knowledge and Data Engineering}\/},  {\it \bibinfo{volume}{25}\/}, \bibinfo{pages}{2708--2720}.
\bibitem[{Delcoucq et~al.(2022)Delcoucq, Dupiereux-Fettweis, Lecron \& Fortemps}]{delcoucq2022resource}
\bibinfo{author}{Delcoucq, L.}, \bibinfo{author}{Dupiereux-Fettweis, T.}, \bibinfo{author}{Lecron, F.}, \& \bibinfo{author}{Fortemps, P.} (\bibinfo{year}{2022}).
\newblock \bibinfo{title}{Resource and activity clustering based on a hierarchical cell formation algorithm}.
\newblock {\it \bibinfo{journal}{Applied Intelligence}\/},  (pp. \bibinfo{pages}{1--10}).
\bibitem[{Delias et~al.(2019)Delias, Doumpos, Grigoroudis \& Matsatsinis}]{delias2019non}
\bibinfo{author}{Delias, P.}, \bibinfo{author}{Doumpos, M.}, \bibinfo{author}{Grigoroudis, E.}, \& \bibinfo{author}{Matsatsinis, N.} (\bibinfo{year}{2019}).
\newblock \bibinfo{title}{A non-compensatory approach for trace clustering}.
\newblock {\it \bibinfo{journal}{International Transactions in Operational Research}\/},  {\it \bibinfo{volume}{26}\/}, \bibinfo{pages}{1828--1846}.
\bibitem[{Di~Francescomarino et~al.(2016)Di~Francescomarino, Dumas, Maggi \& Teinemaa}]{di2016clustering}
\bibinfo{author}{Di~Francescomarino, C.}, \bibinfo{author}{Dumas, M.}, \bibinfo{author}{Maggi, F.~M.}, \& \bibinfo{author}{Teinemaa, I.} (\bibinfo{year}{2016}).
\newblock \bibinfo{title}{Clustering-based predictive process monitoring}.
\newblock {\it \bibinfo{journal}{IEEE Transactions on Services Computing}\/},  {\it \bibinfo{volume}{12}\/}, \bibinfo{pages}{896--909}.
\bibitem[{Di~Francescomarino et~al.(2018)Di~Francescomarino, Ghidini, Maggi \& Milani}]{di2018predictive}
\bibinfo{author}{Di~Francescomarino, C.}, \bibinfo{author}{Ghidini, C.}, \bibinfo{author}{Maggi, F.~M.}, \& \bibinfo{author}{Milani, F.} (\bibinfo{year}{2018}).
\newblock \bibinfo{title}{Predictive process monitoring methods: Which one suits me best?}
\newblock In {\it \bibinfo{booktitle}{{Proceedings of 16th International Conference on Business Process Management}}\/} (pp. \bibinfo{pages}{462--479}).
\newblock \bibinfo{publisher}{Springer}.
\bibitem[{Di~Francescomarino et~al.(2017)Di~Francescomarino, Ghidini, Maggi, Petrucci \& Yeshchenko}]{di2017eye}
\bibinfo{author}{Di~Francescomarino, C.}, \bibinfo{author}{Ghidini, C.}, \bibinfo{author}{Maggi, F.~M.}, \bibinfo{author}{Petrucci, G.}, \& \bibinfo{author}{Yeshchenko, A.} (\bibinfo{year}{2017}).
\newblock \bibinfo{title}{{An eye into the future: Leveraging a-priori knowledge in predictive business process monitoring}}.
\newblock In {\it \bibinfo{booktitle}{{Proceedings of the 15th International Conference on Business Process Management}}\/} (pp. \bibinfo{pages}{252--268}).
\newblock \bibinfo{publisher}{Springer}.
\bibitem[{Di~Mauro et~al.(2019)Di~Mauro, Appice \& Basile}]{mauro2019activity}
\bibinfo{author}{Di~Mauro, N.}, \bibinfo{author}{Appice, A.}, \& \bibinfo{author}{Basile, T.} (\bibinfo{year}{2019}).
\newblock \bibinfo{title}{Activity prediction of business process instances with inception cnn models}.
\newblock In {\it \bibinfo{booktitle}{{Proceedings of the 18th International Conference of the Italian Association for Artificial Intelligence}}\/} (pp. \bibinfo{pages}{348--361}).
\newblock \bibinfo{publisher}{Springer}.
\bibitem[{Diamantini et~al.(2016)Diamantini, Genga \& Potena}]{diamantini2016behavioral}
\bibinfo{author}{Diamantini, C.}, \bibinfo{author}{Genga, L.}, \& \bibinfo{author}{Potena, D.} (\bibinfo{year}{2016}).
\newblock \bibinfo{title}{Behavioral process mining for unstructured processes}.
\newblock {\it \bibinfo{journal}{Journal of Intelligent Information Systems}\/},  {\it \bibinfo{volume}{47}\/}, \bibinfo{pages}{5--32}.
\bibitem[{Dietterich(2002)}]{dietterich2002ensemble}
\bibinfo{author}{Dietterich, T.~G.} (\bibinfo{year}{2002}).
\newblock \bibinfo{title}{Ensemble learning}.
\newblock In {\it \bibinfo{booktitle}{{The Handbook of Brain Theory and Neural Networks}}\/} (pp. \bibinfo{pages}{110--125}).
\newblock \bibinfo{publisher}{MIT Press}. (\bibinfo{edition}{2nd} ed.).
\bibitem[{van Dongen et~al.(2007)van Dongen, Jansen-Vullers, Verbeek \& van~der Aalst}]{dongen2007verification}
\bibinfo{author}{van Dongen, B.~F.}, \bibinfo{author}{Jansen-Vullers, M.~H.}, \bibinfo{author}{Verbeek, H.}, \& \bibinfo{author}{van~der Aalst, W.~M.} (\bibinfo{year}{2007}).
\newblock \bibinfo{title}{Verification of the sap reference models using epc reduction, state-space analysis, and invariants}.
\newblock {\it \bibinfo{journal}{Computers in Industry}\/},  {\it \bibinfo{volume}{58}\/}, \bibinfo{pages}{578--601}.
\bibitem[{Dumas et~al.(2018)Dumas, La~Rosa, Mendling \& Reijers}]{dumas2018fundamentals}
\bibinfo{author}{Dumas, M.}, \bibinfo{author}{La~Rosa, M.}, \bibinfo{author}{Mendling, J.}, \& \bibinfo{author}{Reijers, H.~A.} (\bibinfo{year}{2018}).
\newblock {\it \bibinfo{title}{{Fundamentals of Business Process Management}}\/}.
\newblock (\bibinfo{edition}{2nd} ed.).
\newblock \bibinfo{publisher}{Springer}.
\newblock \DOIprefix\doi{10.1007/978-3-642-33143-5}.
\bibitem[{Effendi \& Sarno(2017)}]{effendi2017discovering}
\bibinfo{author}{Effendi, Y.~A.}, \& \bibinfo{author}{Sarno, R.} (\bibinfo{year}{2017}).
\newblock \bibinfo{title}{Discovering process model from event logs by considering overlapping rules}.
\newblock In {\it \bibinfo{booktitle}{{Proceedings of the 4th International Conference on Electrical Engineering, Computer Science and Informatics}}\/} (pp. \bibinfo{pages}{1--6}).
\newblock \bibinfo{publisher}{IEEE}.
\bibitem[{Es-Soufi et~al.(2016)Es-Soufi, Yahia \& Roucoules}]{Es-Soufi2016}
\bibinfo{author}{Es-Soufi, W.}, \bibinfo{author}{Yahia, E.}, \& \bibinfo{author}{Roucoules, L.} (\bibinfo{year}{2016}).
\newblock \bibinfo{title}{On the use of process mining and machine learning to support decision making in systems design}.
\newblock In {\it \bibinfo{booktitle}{{Proceedings of the 13th IFIP International Conference on Product Lifecycle Management}}\/} (pp. \bibinfo{pages}{56--66}).
\newblock \bibinfo{publisher}{Springer}.
\newblock \DOIprefix\doi{10.1007/978-3-319-54660-5_6}.
\bibitem[{Evermann et~al.(2017)Evermann, Rehse \& Fettke}]{evermann2017}
\bibinfo{author}{Evermann, J.}, \bibinfo{author}{Rehse, J.-R.}, \& \bibinfo{author}{Fettke, P.} (\bibinfo{year}{2017}).
\newblock \bibinfo{title}{Predicting process behaviour using deep learning}.
\newblock {\it \bibinfo{journal}{Decision Support Systems}\/},  {\it \bibinfo{volume}{100}\/}, \bibinfo{pages}{129--140}. \DOIprefix\doi{10.1016/j.dss.2017.04.003}.
\bibitem[{Ferreira \& Gillblad(2009)}]{ferreira2009discovering}
\bibinfo{author}{Ferreira, D.~R.}, \& \bibinfo{author}{Gillblad, D.} (\bibinfo{year}{2009}).
\newblock \bibinfo{title}{Discovering process models from unlabelled event logs}.
\newblock In {\it \bibinfo{booktitle}{Proceedings of the 7th International Conference on Business Process Management}\/} (pp. \bibinfo{pages}{143--158}).
\newblock \bibinfo{organization}{Springer}.
\bibitem[{Ferreira et~al.(2013)Ferreira, Szimanski \& Ralha}]{ferreira2013mining}
\bibinfo{author}{Ferreira, D.~R.}, \bibinfo{author}{Szimanski, F.}, \& \bibinfo{author}{Ralha, C.~G.} (\bibinfo{year}{2013}).
\newblock \bibinfo{title}{Mining the low-level behaviour of agents in high-level business processes}.
\newblock {\it \bibinfo{journal}{International Journal of Business Process Integration and Management}\/},  {\it \bibinfo{volume}{6}\/}, \bibinfo{pages}{146--166}.
\bibitem[{Ferreira \& Vasilyev(2015)}]{ferreira2015using}
\bibinfo{author}{Ferreira, D.~R.}, \& \bibinfo{author}{Vasilyev, E.} (\bibinfo{year}{2015}).
\newblock \bibinfo{title}{Using logical decision trees to discover the cause of process delays from event logs}.
\newblock {\it \bibinfo{journal}{Computers in Industry}\/},  {\it \bibinfo{volume}{70}\/}, \bibinfo{pages}{194--207}.
\bibitem[{Ferreira et~al.(2007)Ferreira, Zacarias, Malheiros \& Ferreira}]{ferreira2007approaching}
\bibinfo{author}{Ferreira, D.~R.}, \bibinfo{author}{Zacarias, M.}, \bibinfo{author}{Malheiros, M.}, \& \bibinfo{author}{Ferreira, P.} (\bibinfo{year}{2007}).
\newblock \bibinfo{title}{Approaching process mining with sequence clustering: Experiments and findings}.
\newblock In {\it \bibinfo{booktitle}{{Proceedings of the 5th International Conference on Business Process Management}}\/} (pp. \bibinfo{pages}{360--374}).
\newblock \bibinfo{publisher}{Springer}.
\bibitem[{Ferri et~al.(2009)Ferri, Hern{\'a}ndez-Orallo \& Modroiu}]{ferri2009experimental}
\bibinfo{author}{Ferri, C.}, \bibinfo{author}{Hern{\'a}ndez-Orallo, J.}, \& \bibinfo{author}{Modroiu, R.} (\bibinfo{year}{2009}).
\newblock \bibinfo{title}{An experimental comparison of performance measures for classification}.
\newblock {\it \bibinfo{journal}{Pattern Recognition Letters}\/},  {\it \bibinfo{volume}{30}\/}, \bibinfo{pages}{27--38}.
\bibitem[{Feuerriegel et~al.(2020)Feuerriegel, Dolata \& Schwabe}]{feuerriegel2020fair}
\bibinfo{author}{Feuerriegel, S.}, \bibinfo{author}{Dolata, M.}, \& \bibinfo{author}{Schwabe, G.} (\bibinfo{year}{2020}).
\newblock \bibinfo{title}{{Fair AI}}.
\newblock {\it \bibinfo{journal}{Business \& Information Systems Engineering}\/},  {\it \bibinfo{volume}{62}\/}, \bibinfo{pages}{379--384}.
\bibitem[{Feuerriegel et~al.(2023)Feuerriegel, Hartmann, Janiesch \& Zschech}]{feuerriegel2023generative}
\bibinfo{author}{Feuerriegel, S.}, \bibinfo{author}{Hartmann, J.}, \bibinfo{author}{Janiesch, C.}, \& \bibinfo{author}{Zschech, P.} (\bibinfo{year}{2023}).
\newblock \bibinfo{title}{Generative ai}.
\newblock {\it \bibinfo{journal}{Business \& Information Systems Engineering}\/}, .
\newblock \bibinfo{note}{Forthcoming}.
\bibitem[{Firouzian et~al.(2019{\natexlab{a}})Firouzian, Zahedi \& Hassanpour}]{firouzian2019cycle}
\bibinfo{author}{Firouzian, I.}, \bibinfo{author}{Zahedi, M.}, \& \bibinfo{author}{Hassanpour, H.} (\bibinfo{year}{2019}{\natexlab{a}}).
\newblock \bibinfo{title}{Cycle time optimization of processes using an entropy-based learning for task allocation}.
\newblock {\it \bibinfo{journal}{International Journal of Engineering}\/},  {\it \bibinfo{volume}{32}\/}, \bibinfo{pages}{1090--1100}.
\bibitem[{Firouzian et~al.(2019{\natexlab{b}})Firouzian, Zahedi \& Hassanpour}]{firouzian2019investigation}
\bibinfo{author}{Firouzian, I.}, \bibinfo{author}{Zahedi, M.}, \& \bibinfo{author}{Hassanpour, H.} (\bibinfo{year}{2019}{\natexlab{b}}).
\newblock \bibinfo{title}{Investigation of the effect of concept drift on data-aware remaining time prediction of business processes}.
\newblock {\it \bibinfo{journal}{International Journal of Nonlinear Analysis and Applications}\/},  {\it \bibinfo{volume}{10}\/}, \bibinfo{pages}{153--166}.
\bibitem[{Fleiss \& Cohen(1973)}]{fleiss1973equivalence}
\bibinfo{author}{Fleiss, J.~L.}, \& \bibinfo{author}{Cohen, J.} (\bibinfo{year}{1973}).
\newblock \bibinfo{title}{The equivalence of weighted kappa and the intraclass correlation coefficient as measures of reliability}.
\newblock {\it \bibinfo{journal}{{Educational and Psychological Measurement}}\/},  {\it \bibinfo{volume}{33}\/}, \bibinfo{pages}{613--619}.
\bibitem[{Folino et~al.(2022)Folino, Folino, Guarascio \& Pontieri}]{folino2022semi}
\bibinfo{author}{Folino, F.}, \bibinfo{author}{Folino, G.}, \bibinfo{author}{Guarascio, M.}, \& \bibinfo{author}{Pontieri, L.} (\bibinfo{year}{2022}).
\newblock \bibinfo{title}{{Semi-supervised discovery of DNN-based outcome predictors from scarcely-labeled process logs}}.
\newblock {\it \bibinfo{journal}{Business \& Information Systems Engineering}\/},  (pp. \bibinfo{pages}{1--21}).
\bibitem[{Folino et~al.(2011)Folino, Greco, Guzzo \& Pontieri}]{folino2011mining}
\bibinfo{author}{Folino, F.}, \bibinfo{author}{Greco, G.}, \bibinfo{author}{Guzzo, A.}, \& \bibinfo{author}{Pontieri, L.} (\bibinfo{year}{2011}).
\newblock \bibinfo{title}{Mining usage scenarios in business processes: Outlier-aware discovery and run-time prediction}.
\newblock {\it \bibinfo{journal}{Data \& Knowledge Engineering}\/},  {\it \bibinfo{volume}{70}\/}, \bibinfo{pages}{1005--1029}.
\bibitem[{Folino et~al.(2012)Folino, Guarascio \& Pontieri}]{folino2012discovering}
\bibinfo{author}{Folino, F.}, \bibinfo{author}{Guarascio, M.}, \& \bibinfo{author}{Pontieri, L.} (\bibinfo{year}{2012}).
\newblock \bibinfo{title}{Discovering context-aware models for predicting business process performances}.
\newblock In {\it \bibinfo{booktitle}{{Proceedings of the 2012 OTM Confederated International Conferences "On the Move to Meaningful Internet Systems"}}\/} (pp. \bibinfo{pages}{287--304}).
\newblock \bibinfo{publisher}{Springer}.
\bibitem[{Folino et~al.(2015)Folino, Guarascio \& Pontieri}]{folino2015mining}
\bibinfo{author}{Folino, F.}, \bibinfo{author}{Guarascio, M.}, \& \bibinfo{author}{Pontieri, L.} (\bibinfo{year}{2015}).
\newblock \bibinfo{title}{Mining multi-variant process models from low-level logs}.
\newblock In {\it \bibinfo{booktitle}{{Proceedings of the 18th International Conference on Business Information Systems}}\/} (pp. \bibinfo{pages}{165--177}).
\newblock \bibinfo{publisher}{Springer}.
\bibitem[{Folino \& Pontieri(2021)}]{folino2021ai}
\bibinfo{author}{Folino, F.}, \& \bibinfo{author}{Pontieri, L.} (\bibinfo{year}{2021}).
\newblock \bibinfo{title}{{AI-empowered process mining for complex application scenarios: Survey and discussion}}.
\newblock {\it \bibinfo{journal}{Journal on Data Semantics}\/},  {\it \bibinfo{volume}{10}\/}, \bibinfo{pages}{77--106}. \DOIprefix\doi{10.1007/s13740-021-00121-2}.
\bibitem[{Forgy(1965)}]{forgy1965cluster}
\bibinfo{author}{Forgy, E.~W.} (\bibinfo{year}{1965}).
\newblock \bibinfo{title}{{Cluster analysis of multivariate data: Efficiency versus interpretability of classifications}}.
\newblock {\it \bibinfo{journal}{Biometrics}\/},  {\it \bibinfo{volume}{21}\/}, \bibinfo{pages}{768--769}.
\bibitem[{F{\"u}rnkranz et~al.(2012)F{\"u}rnkranz, Gamberger \& Lavra{\v{c}}}]{furnkranz2012foundations}
\bibinfo{author}{F{\"u}rnkranz, J.}, \bibinfo{author}{Gamberger, D.}, \& \bibinfo{author}{Lavra{\v{c}}, N.} (\bibinfo{year}{2012}).
\newblock {\it \bibinfo{title}{Foundations of rule learning}\/}.
\newblock \bibinfo{publisher}{Springer}.
\bibitem[{Galanti et~al.(2020)Galanti, Coma-Puig, de~Leoni, Carmona \& Navarin}]{galanti2020explainable}
\bibinfo{author}{Galanti, R.}, \bibinfo{author}{Coma-Puig, B.}, \bibinfo{author}{de~Leoni, M.}, \bibinfo{author}{Carmona, J.}, \& \bibinfo{author}{Navarin, N.} (\bibinfo{year}{2020}).
\newblock \bibinfo{title}{Explainable predictive process monitoring}.
\newblock In {\it \bibinfo{booktitle}{{Proceedings of the 2nd International Conference on Process Mining}}\/} (pp. \bibinfo{pages}{1--8}).
\newblock \bibinfo{publisher}{IEEE}.
\bibitem[{Garcez et~al.(2022)Garcez, Bader, Bowman, Lamb, de~Penning, Illuminoo, Poon \& Gerson~Zaverucha}]{garcez2022neural}
\bibinfo{author}{Garcez, A.~d.}, \bibinfo{author}{Bader, S.}, \bibinfo{author}{Bowman, H.}, \bibinfo{author}{Lamb, L.~C.}, \bibinfo{author}{de~Penning, L.}, \bibinfo{author}{Illuminoo, B.}, \bibinfo{author}{Poon, H.}, \& \bibinfo{author}{Gerson~Zaverucha, C.} (\bibinfo{year}{2022}).
\newblock \bibinfo{title}{Neural-symbolic learning and reasoning: A survey and interpretation}.
\newblock In {\it \bibinfo{booktitle}{Neuro-Symbolic Artificial Intelligence: The State of the Art}\/} (pp. \bibinfo{pages}{1--51}).
\newblock \bibinfo{publisher}{IOS Press}.
\bibitem[{Garcez et~al.(2015)Garcez, Besold, De~Raedt, F{\"o}ldiak, Hitzler, Icard, K{\"u}hnberger, Lamb, Miikkulainen \& Silver}]{garcez2015neural}
\bibinfo{author}{Garcez, A.~d.}, \bibinfo{author}{Besold, T.~R.}, \bibinfo{author}{De~Raedt, L.}, \bibinfo{author}{F{\"o}ldiak, P.}, \bibinfo{author}{Hitzler, P.}, \bibinfo{author}{Icard, T.}, \bibinfo{author}{K{\"u}hnberger, K.-U.}, \bibinfo{author}{Lamb, L.~C.}, \bibinfo{author}{Miikkulainen, R.}, \& \bibinfo{author}{Silver, D.~L.} (\bibinfo{year}{2015}).
\newblock \bibinfo{title}{{Neural-symbolic learning and reasoning: Contributions and challenges}}.
\newblock In {\it \bibinfo{booktitle}{{Proceedings of the 2015 AAAI Spring Symposium -- Knowledge Representation and Reasoning: Integrating Symbolic and Neural Approaches}}\/} (pp. \bibinfo{pages}{18--21}).
\newblock \bibinfo{publisher}{AI Access Foundation}.
\bibitem[{Garc{\'\i}a-Ba{\~n}uelos et~al.(2014)Garc{\'\i}a-Ba{\~n}uelos, Dumas, La~Rosa, De~Weerdt \& Ekanayake}]{garcia2014controlled}
\bibinfo{author}{Garc{\'\i}a-Ba{\~n}uelos, L.}, \bibinfo{author}{Dumas, M.}, \bibinfo{author}{La~Rosa, M.}, \bibinfo{author}{De~Weerdt, J.}, \& \bibinfo{author}{Ekanayake, C.~C.} (\bibinfo{year}{2014}).
\newblock \bibinfo{title}{Controlled automated discovery of collections of business process models}.
\newblock {\it \bibinfo{journal}{Information Systems}\/},  {\it \bibinfo{volume}{46}\/}, \bibinfo{pages}{85--101}.
\bibitem[{Gerlach et~al.(2022)Gerlach, Seeliger, Nolle \& Mühlhäuser}]{gerlach2022inferring}
\bibinfo{author}{Gerlach, Y.}, \bibinfo{author}{Seeliger, A.}, \bibinfo{author}{Nolle, T.}, \& \bibinfo{author}{Mühlhäuser, M.} (\bibinfo{year}{2022}).
\newblock \bibinfo{title}{Inferring a multi-perspective likelihood graph from black-box next event predictors}.
\newblock In {\it \bibinfo{booktitle}{{Proceedings of the 34th International Conference on Advanced Information Systems Engineering}}\/} (pp. \bibinfo{pages}{19--35}).
\newblock \bibinfo{publisher}{Springer}.
\bibitem[{Goodfellow et~al.(2016)Goodfellow, Bengio \& Courville}]{goodfellow2016deep}
\bibinfo{author}{Goodfellow, I.}, \bibinfo{author}{Bengio, Y.}, \& \bibinfo{author}{Courville, A.} (\bibinfo{year}{2016}).
\newblock {\it \bibinfo{title}{{Deep Learning}}\/}.
\newblock \bibinfo{publisher}{MIT Press}.
\bibitem[{Greco et~al.(2004)Greco, Guzzo, Pontieri \& Sacca}]{greco2004mining}
\bibinfo{author}{Greco, G.}, \bibinfo{author}{Guzzo, A.}, \bibinfo{author}{Pontieri, L.}, \& \bibinfo{author}{Sacca, D.} (\bibinfo{year}{2004}).
\newblock \bibinfo{title}{Mining expressive process models by clustering workflow traces}.
\newblock In {\it \bibinfo{booktitle}{{Proceedings of the 8th Pacific-Asia Conference on Knowledge Discovery and Data Mining}}\/} (pp. \bibinfo{pages}{52--62}).
\newblock \bibinfo{publisher}{Springer}.
\bibitem[{Greco et~al.(2006)Greco, Guzzo, Pontieri \& Sacca}]{greco2006discovering}
\bibinfo{author}{Greco, G.}, \bibinfo{author}{Guzzo, A.}, \bibinfo{author}{Pontieri, L.}, \& \bibinfo{author}{Sacca, D.} (\bibinfo{year}{2006}).
\newblock \bibinfo{title}{Discovering expressive process models by clustering log traces}.
\newblock {\it \bibinfo{journal}{{IEEE Transactions on Knowledge and Data Engineering}}\/},  {\it \bibinfo{volume}{18}\/}, \bibinfo{pages}{1010--1027}.
\bibitem[{Grigori et~al.(2004)Grigori, Casati, Castellanos, Dayal, Sayal \& Shan}]{grigori.2004}
\bibinfo{author}{Grigori, D.}, \bibinfo{author}{Casati, F.}, \bibinfo{author}{Castellanos, M.}, \bibinfo{author}{Dayal, U.}, \bibinfo{author}{Sayal, M.}, \& \bibinfo{author}{Shan, M.-C.} (\bibinfo{year}{2004}).
\newblock \bibinfo{title}{Business process intelligence}.
\newblock {\it \bibinfo{journal}{Computers in Industry}\/},  {\it \bibinfo{volume}{53}\/}, \bibinfo{pages}{321--343}. \DOIprefix\doi{0.1016/j.compind.2003.10.007}.
\bibitem[{Guo et~al.(2023)Guo, Zhang, Wang, Jiang, Nie, Ding, Yue \& Wu}]{guo2023close}
\bibinfo{author}{Guo, B.}, \bibinfo{author}{Zhang, X.}, \bibinfo{author}{Wang, Z.}, \bibinfo{author}{Jiang, M.}, \bibinfo{author}{Nie, J.}, \bibinfo{author}{Ding, Y.}, \bibinfo{author}{Yue, J.}, \& \bibinfo{author}{Wu, Y.} (\bibinfo{year}{2023}).
\newblock \bibinfo{title}{How close is chatgpt to human experts? comparison corpus, evaluation, and detection}.
\newblock {\it \bibinfo{journal}{arXiv preprint arXiv:2301.07597}\/}, . \href{http://arxiv.org/abs/2301.07597}{\tt arXiv:2301.07597}.
\bibitem[{Guzzo et~al.(2021)Guzzo, Joaristi, Rullo \& Serra}]{guzzo2021multi}
\bibinfo{author}{Guzzo, A.}, \bibinfo{author}{Joaristi, M.}, \bibinfo{author}{Rullo, A.}, \& \bibinfo{author}{Serra, E.} (\bibinfo{year}{2021}).
\newblock \bibinfo{title}{A multi-perspective approach for the analysis of complex business processes behavior}.
\newblock {\it \bibinfo{journal}{Expert Systems with Applications}\/},  {\it \bibinfo{volume}{177}\/}, \bibinfo{pages}{114934}.
\bibitem[{Ha et~al.(2016)Ha, Bui \& Nguyen}]{ha2016trace}
\bibinfo{author}{Ha, Q.-T.}, \bibinfo{author}{Bui, H.-N.}, \& \bibinfo{author}{Nguyen, T.-T.} (\bibinfo{year}{2016}).
\newblock \bibinfo{title}{A trace clustering solution based on using the distance graph model}.
\newblock In {\it \bibinfo{booktitle}{{Proceedings of the 8th International Conference on Computational Collective Intelligence}}\/} (pp. \bibinfo{pages}{313--322}).
\newblock \bibinfo{publisher}{Springer}.
\bibitem[{Hanga et~al.(2020)Hanga, Kovalchuk \& Gaber}]{hanga2020graph}
\bibinfo{author}{Hanga, K.~M.}, \bibinfo{author}{Kovalchuk, Y.}, \& \bibinfo{author}{Gaber, M.~M.} (\bibinfo{year}{2020}).
\newblock \bibinfo{title}{A graph-based approach to interpreting recurrent neural networks in process mining}.
\newblock {\it \bibinfo{journal}{IEEE Access}\/},  {\it \bibinfo{volume}{8}\/}, \bibinfo{pages}{172923--172938}.
\bibitem[{Harl et~al.(2020)Harl, Weinzierl, Stierle \& Matzner}]{harl2020explainable}
\bibinfo{author}{Harl, M.}, \bibinfo{author}{Weinzierl, S.}, \bibinfo{author}{Stierle, M.}, \& \bibinfo{author}{Matzner, M.} (\bibinfo{year}{2020}).
\newblock \bibinfo{title}{Explainable predictive business process monitoring using gated graph neural networks}.
\newblock {\it \bibinfo{journal}{Journal of Decision Systems}\/},  {\it \bibinfo{volume}{29}\/}, \bibinfo{pages}{312--327}.
\bibitem[{Hastie et~al.(2009)Hastie, Tibshirani \& Friedman}]{hastie2009elements}
\bibinfo{author}{Hastie, T.}, \bibinfo{author}{Tibshirani, R.}, \& \bibinfo{author}{Friedman, J.~H.} (\bibinfo{year}{2009}).
\newblock {\it \bibinfo{title}{{The Elements of Statistical Learning: Data mining, Inference, and Prediction}}\/} volume~\bibinfo{volume}{2}.
\newblock \bibinfo{publisher}{Springer}.
\bibitem[{Hayes-Roth(1985)}]{hayes1985rule}
\bibinfo{author}{Hayes-Roth, F.} (\bibinfo{year}{1985}).
\newblock \bibinfo{title}{Rule-based systems}.
\newblock {\it \bibinfo{journal}{Communications of the ACM}\/},  {\it \bibinfo{volume}{28}\/}, \bibinfo{pages}{921--932}.
\bibitem[{Heinrich et~al.(2021)Heinrich, Zschech, Janiesch \& Bonin}]{heinrich2021process}
\bibinfo{author}{Heinrich, K.}, \bibinfo{author}{Zschech, P.}, \bibinfo{author}{Janiesch, C.}, \& \bibinfo{author}{Bonin, M.} (\bibinfo{year}{2021}).
\newblock \bibinfo{title}{{Process data properties matter: Introducing gated convolutional neural networks (GCNN) and key-value-predict attention networks (KVP) for next event prediction with deep learning}}.
\newblock {\it \bibinfo{journal}{Decision Support Systems}\/},  {\it \bibinfo{volume}{143}\/}, \bibinfo{pages}{113494}. \DOIprefix\doi{10.1016/j.dss.2021.113494}.
\bibitem[{Herbst(2000)}]{herbst2000machine}
\bibinfo{author}{Herbst, J.} (\bibinfo{year}{2000}).
\newblock \bibinfo{title}{A machine learning approach to workflow management}.
\newblock In {\it \bibinfo{booktitle}{{Proceedings of the 11th European Conference on Machine Learning}}\/} (pp. \bibinfo{pages}{183--194}).
\newblock \bibinfo{publisher}{Springer}.
\bibitem[{Herbst \& Karagiannis(2000)}]{herbst2000integrating}
\bibinfo{author}{Herbst, J.}, \& \bibinfo{author}{Karagiannis, D.} (\bibinfo{year}{2000}).
\newblock \bibinfo{title}{Integrating machine learning and workflow management to support acquisition and adaptation of workflow models}.
\newblock {\it \bibinfo{journal}{Intelligent Systems in Accounting, Finance \& Management}\/},  {\it \bibinfo{volume}{9}\/}, \bibinfo{pages}{67--92}.
\bibitem[{Herbst \& Karagiannis(2004)}]{herbst2004workflow}
\bibinfo{author}{Herbst, J.}, \& \bibinfo{author}{Karagiannis, D.} (\bibinfo{year}{2004}).
\newblock \bibinfo{title}{{Workflow mining with InWoLvE}}.
\newblock {\it \bibinfo{journal}{Computers in Industry}\/},  {\it \bibinfo{volume}{53}\/}, \bibinfo{pages}{245--264}.
\bibitem[{Herm et~al.(2021)Herm, Janiesch, Reijers \& Seubert}]{herm2021symbolic}
\bibinfo{author}{Herm, L.-V.}, \bibinfo{author}{Janiesch, C.}, \bibinfo{author}{Reijers, H.~A.}, \& \bibinfo{author}{Seubert, F.} (\bibinfo{year}{2021}).
\newblock \bibinfo{title}{{From symbolic RPA to intelligent RPA: Challenges for developing and operating intelligent software robots}}.
\newblock In {\it \bibinfo{booktitle}{{Proceedings of the 19th International Conference on Business Process Management}}\/} (pp. \bibinfo{pages}{289--305}).
\newblock \bibinfo{publisher}{Springer}.
\bibitem[{Houy et~al.(2010)Houy, Fettke \& Loos}]{Houy2010}
\bibinfo{author}{Houy, C.}, \bibinfo{author}{Fettke, P.}, \& \bibinfo{author}{Loos, P.} (\bibinfo{year}{2010}).
\newblock \bibinfo{title}{{Empirical research in business process management –- Analysis of an emerging field of research}}.
\newblock {\it \bibinfo{journal}{Business Process Management Journal}\/},  {\it \bibinfo{volume}{16}\/}, \bibinfo{pages}{619--661}. \DOIprefix\doi{10.1108/14637151011065946}.
\bibitem[{Hruschka et~al.(2004)Hruschka, Schwartz, St.~John, Picone-Decaro, Jenkins \& Carey}]{hruschka2004reliability}
\bibinfo{author}{Hruschka, D.~J.}, \bibinfo{author}{Schwartz, D.}, \bibinfo{author}{St.~John, D.~C.}, \bibinfo{author}{Picone-Decaro, E.}, \bibinfo{author}{Jenkins, R.~A.}, \& \bibinfo{author}{Carey, J.~W.} (\bibinfo{year}{2004}).
\newblock \bibinfo{title}{Reliability in coding open-ended data: Lessons learned from {HIV} behavioral research}.
\newblock {\it \bibinfo{journal}{Field methods}\/},  {\it \bibinfo{volume}{16}\/}, \bibinfo{pages}{307--331}.
\bibitem[{Hsieh et~al.(2021)Hsieh, Moreira \& Ouyang}]{hsieh2021dice4el}
\bibinfo{author}{Hsieh, C.}, \bibinfo{author}{Moreira, C.}, \& \bibinfo{author}{Ouyang, C.} (\bibinfo{year}{2021}).
\newblock \bibinfo{title}{{DICE4EL: Interpreting process predictions using a milestone-aware counterfactual approach}}.
\newblock In {\it \bibinfo{booktitle}{{Proceedings of the 3rd International Conference on Process Mining}}\/} (pp. \bibinfo{pages}{88--95}).
\newblock \bibinfo{publisher}{IEEE}.
\bibitem[{Huang et~al.(2010)Huang, van~der Aalst, Lu \& Duan}]{huang2010adaptive}
\bibinfo{author}{Huang, Z.}, \bibinfo{author}{van~der Aalst, W. M.~P.}, \bibinfo{author}{Lu, X.}, \& \bibinfo{author}{Duan, H.} (\bibinfo{year}{2010}).
\newblock \bibinfo{title}{An adaptive work distribution mechanism based on reinforcement learning}.
\newblock {\it \bibinfo{journal}{Expert Systems with Applications}\/},  {\it \bibinfo{volume}{37}\/}, \bibinfo{pages}{7533--7541}.
\bibitem[{Huang et~al.(2011)Huang, van~der Aalst, Lu \& Duan}]{huang2011reinforcement}
\bibinfo{author}{Huang, Z.}, \bibinfo{author}{van~der Aalst, W. M.~P.}, \bibinfo{author}{Lu, X.}, \& \bibinfo{author}{Duan, H.} (\bibinfo{year}{2011}).
\newblock \bibinfo{title}{Reinforcement learning based resource allocation in business process management}.
\newblock {\it \bibinfo{journal}{Data \& Knowledge Engineering}\/},  {\it \bibinfo{volume}{70}\/}, \bibinfo{pages}{127--145}.
\bibitem[{Huo et~al.(2021)Huo, V{\"o}lzer, Reddy, Agarwal, Isahagian \& Muthusamy}]{huo2021graph}
\bibinfo{author}{Huo, S.}, \bibinfo{author}{V{\"o}lzer, H.}, \bibinfo{author}{Reddy, P.}, \bibinfo{author}{Agarwal, P.}, \bibinfo{author}{Isahagian, V.}, \& \bibinfo{author}{Muthusamy, V.} (\bibinfo{year}{2021}).
\newblock \bibinfo{title}{Graph autoencoders for business process anomaly detection}.
\newblock In {\it \bibinfo{booktitle}{{Proceedings of the 19th International Conference on Business Process Management}}\/} (pp. \bibinfo{pages}{417--433}).
\newblock \bibinfo{publisher}{Springer}.
\bibitem[{Jaiswal et~al.(2021)Jaiswal, Babu, Zadeh, Banerjee \& Makedon}]{jaiswal2021survey}
\bibinfo{author}{Jaiswal, A.}, \bibinfo{author}{Babu, A.~R.}, \bibinfo{author}{Zadeh, M.~Z.}, \bibinfo{author}{Banerjee, D.}, \& \bibinfo{author}{Makedon, F.} (\bibinfo{year}{2021}).
\newblock \bibinfo{title}{A survey on contrastive self-supervised learning}.
\newblock {\it \bibinfo{journal}{Technologies}\/},  {\it \bibinfo{volume}{9}\/}, \bibinfo{pages}{1--22}. \DOIprefix\doi{10.3390/technologies9010002}.
\bibitem[{Jalayer et~al.(2020)Jalayer, Kahani, Beheshti, Pourmasoumi \& Motahari-Nezhad}]{jalayer2020attention}
\bibinfo{author}{Jalayer, A.}, \bibinfo{author}{Kahani, M.}, \bibinfo{author}{Beheshti, A.}, \bibinfo{author}{Pourmasoumi, A.}, \& \bibinfo{author}{Motahari-Nezhad, H.~R.} (\bibinfo{year}{2020}).
\newblock \bibinfo{title}{Attention mechanism in predictive business process monitoring}.
\newblock In {\it \bibinfo{booktitle}{{Proceedings of the 24th IEEE International Enterprise Distributed Object Computing Conference}}\/} (pp. \bibinfo{pages}{181--186}).
\newblock \bibinfo{publisher}{IEEE}.
\bibitem[{Janiesch et~al.(2021)Janiesch, Zschech \& Heinrich}]{janiesch2021machine}
\bibinfo{author}{Janiesch, C.}, \bibinfo{author}{Zschech, P.}, \& \bibinfo{author}{Heinrich, K.} (\bibinfo{year}{2021}).
\newblock \bibinfo{title}{Machine learning and deep learning}.
\newblock {\it \bibinfo{journal}{Electronic Markets}\/},  {\it \bibinfo{volume}{31}\/}, \bibinfo{pages}{685--695}. \DOIprefix\doi{10.1007/s12525-021-00475-2}.
\bibitem[{Jing \& Tian(2021)}]{Longlong2021}
\bibinfo{author}{Jing, L.}, \& \bibinfo{author}{Tian, Y.} (\bibinfo{year}{2021}).
\newblock \bibinfo{title}{Self-supervised visual feature learning with deep neural networks: A survey}.
\newblock {\it \bibinfo{journal}{IEEE Transactions on Pattern Analysis and Machine Intelligence}\/},  {\it \bibinfo{volume}{43}\/}, \bibinfo{pages}{4037--4058}. \DOIprefix\doi{10.1109/TPAMI.2020.2992393}.
\bibitem[{Jlailaty et~al.(2017)Jlailaty, Grigori \& Belhajjame}]{jlailaty2017business}
\bibinfo{author}{Jlailaty, D.}, \bibinfo{author}{Grigori, D.}, \& \bibinfo{author}{Belhajjame, K.} (\bibinfo{year}{2017}).
\newblock \bibinfo{title}{Business process instances discovery from email logs}.
\newblock In {\it \bibinfo{booktitle}{{Proceeding of the 14th International Conference on Services Computing}}\/} (pp. \bibinfo{pages}{19--26}).
\newblock \bibinfo{publisher}{IEEE}.
\bibitem[{Jobin et~al.(2019)Jobin, Ienca \& Vayena}]{jobin2019global}
\bibinfo{author}{Jobin, A.}, \bibinfo{author}{Ienca, M.}, \& \bibinfo{author}{Vayena, E.} (\bibinfo{year}{2019}).
\newblock \bibinfo{title}{{The global landscape of AI ethics guidelines}}.
\newblock {\it \bibinfo{journal}{Nature Machine Intelligence}\/},  {\it \bibinfo{volume}{1}\/}, \bibinfo{pages}{389--399}.
\bibitem[{Jordan \& Mitchell(2015)}]{jordan2015machine}
\bibinfo{author}{Jordan, M.~I.}, \& \bibinfo{author}{Mitchell, T.~M.} (\bibinfo{year}{2015}).
\newblock \bibinfo{title}{Machine learning: Trends, perspectives, and prospects}.
\newblock {\it \bibinfo{journal}{Science}\/},  {\it \bibinfo{volume}{349}\/}, \bibinfo{pages}{255--260}.
\bibitem[{Jung et~al.(2008)Jung, Bae \& Liu}]{jung2008hierarchical}
\bibinfo{author}{Jung, J.-Y.}, \bibinfo{author}{Bae, J.}, \& \bibinfo{author}{Liu, L.} (\bibinfo{year}{2008}).
\newblock \bibinfo{title}{Hierarchical business process clustering}.
\newblock In {\it \bibinfo{booktitle}{{Proceedings of the 2008 IEEE International Conference on Services Computing}}\/} (pp. \bibinfo{pages}{613--616}).
\newblock \bibinfo{publisher}{IEEE}.
\bibitem[{Jung et~al.(2009)Jung, Bae \& Liu}]{jung2009hierarchical}
\bibinfo{author}{Jung, J.-Y.}, \bibinfo{author}{Bae, J.}, \& \bibinfo{author}{Liu, L.} (\bibinfo{year}{2009}).
\newblock \bibinfo{title}{Hierarchical clustering of business process models}.
\newblock {\it \bibinfo{journal}{International Journal of Innovative Computing, Information and Control}\/},  {\it \bibinfo{volume}{5}\/}, \bibinfo{pages}{1349--4198}.
\bibitem[{Junior et~al.(2020)Junior, Ceravolo, Damiani, Omori \& Tavares}]{junior2020anomaly}
\bibinfo{author}{Junior, S.~B.}, \bibinfo{author}{Ceravolo, P.}, \bibinfo{author}{Damiani, E.}, \bibinfo{author}{Omori, N.~J.}, \& \bibinfo{author}{Tavares, G.~M.} (\bibinfo{year}{2020}).
\newblock \bibinfo{title}{Anomaly detection on event logs with a scarcity of labels}.
\newblock In {\it \bibinfo{booktitle}{{Proceedings of the 2nd International Conference on Process Mining}}\/} (pp. \bibinfo{pages}{161--168}).
\newblock \bibinfo{publisher}{IEEE}.
\bibitem[{Junior et~al.(2018)Junior, Tavares, da~Costa, Ceravolo \& Damiani}]{barbon2018framework}
\bibinfo{author}{Junior, S.~B.}, \bibinfo{author}{Tavares, G.~M.}, \bibinfo{author}{da~Costa, V. G.~T.}, \bibinfo{author}{Ceravolo, P.}, \& \bibinfo{author}{Damiani, E.} (\bibinfo{year}{2018}).
\newblock \bibinfo{title}{A framework for human-in-the-loop monitoring of concept-drift detection in event log stream}.
\newblock In {\it \bibinfo{booktitle}{{Proceedings of the The Web Conference 2018}}\/} (pp. \bibinfo{pages}{319--326}).
\bibitem[{Kaddour et~al.(2022)Kaddour, Lynch, Liu, Kusner \& Silva}]{kaddour2022causal}
\bibinfo{author}{Kaddour, J.}, \bibinfo{author}{Lynch, A.}, \bibinfo{author}{Liu, Q.}, \bibinfo{author}{Kusner, M.~J.}, \& \bibinfo{author}{Silva, R.} (\bibinfo{year}{2022}).
\newblock \bibinfo{title}{{Causal machine learning: A survey and open problems}}.
\newblock {\it \bibinfo{journal}{arXiv preprint arXiv:2206.15475}\/},  (pp. \bibinfo{pages}{1--188}).
\bibitem[{Kaelbling et~al.(1996)Kaelbling, Littman \& Moore}]{Kaelbling1996}
\bibinfo{author}{Kaelbling, L.~P.}, \bibinfo{author}{Littman, M.~L.}, \& \bibinfo{author}{Moore, A.~W.} (\bibinfo{year}{1996}).
\newblock \bibinfo{title}{{Reinforcement learning: A survey}}.
\newblock {\it \bibinfo{journal}{Journal of Artificial Intelligence Research}\/},  {\it \bibinfo{volume}{4}\/}, \bibinfo{pages}{237--285}. \DOIprefix\doi{10.1613/jair.301}.
\bibitem[{Kang et~al.(2012)Kang, Kim \& Kang}]{kang2012periodic}
\bibinfo{author}{Kang, B.}, \bibinfo{author}{Kim, D.}, \& \bibinfo{author}{Kang, S.-H.} (\bibinfo{year}{2012}).
\newblock \bibinfo{title}{Periodic performance prediction for real-time business process monitoring}.
\newblock {\it \bibinfo{journal}{Industrial Management \& Data Systems}\/},  {\it \bibinfo{volume}{112}\/}, \bibinfo{pages}{4--23}.
\bibitem[{Kazakov et~al.(2018)Kazakov, Novikov, Kulagina \& Shlapakova}]{kazakov2018development}
\bibinfo{author}{Kazakov, O.~D.}, \bibinfo{author}{Novikov, S.~P.}, \bibinfo{author}{Kulagina, N.~A.}, \& \bibinfo{author}{Shlapakova, S.~N.} (\bibinfo{year}{2018}).
\newblock \bibinfo{title}{Development of the concept of management of economic systems processes through construction and calling of machine learning models}.
\newblock In {\it \bibinfo{booktitle}{{Proceedings of the 2018 IEEE International Conference "Quality Management, Transport and Information Security, Information Technologies"}}\/} (pp. \bibinfo{pages}{316--321}).
\newblock \bibinfo{publisher}{IEEE}.
\bibitem[{Kazim \& Koshiyama(2021)}]{kazim2021high}
\bibinfo{author}{Kazim, E.}, \& \bibinfo{author}{Koshiyama, A.~S.} (\bibinfo{year}{2021}).
\newblock \bibinfo{title}{{A high-level overview of AI ethics}}.
\newblock {\it \bibinfo{journal}{Patterns}\/},  {\it \bibinfo{volume}{2}\/}, \bibinfo{pages}{100314}.
\bibitem[{Khan et~al.(2021)Khan, Ghose \& Dam}]{khan2021decision}
\bibinfo{author}{Khan, A.}, \bibinfo{author}{Ghose, A.}, \& \bibinfo{author}{Dam, H.} (\bibinfo{year}{2021}).
\newblock \bibinfo{title}{{Decision support for knowledge intensive processes using RL based recommendations}}.
\newblock In {\it \bibinfo{booktitle}{{Proceedings of the 19th International Conference of Business Process Management (Forum)}}\/} (pp. \bibinfo{pages}{246--262}).
\newblock \bibinfo{publisher}{Springer}.
\bibitem[{Khodyrev \& Popova(2014)}]{khodyrev2014discrete}
\bibinfo{author}{Khodyrev, I.}, \& \bibinfo{author}{Popova, S.} (\bibinfo{year}{2014}).
\newblock \bibinfo{title}{Discrete modeling and simulation of business processes using event logs}.
\newblock {\it \bibinfo{journal}{Procedia Computer Science}\/},  {\it \bibinfo{volume}{29}\/}, \bibinfo{pages}{322--331}.
\bibitem[{Kim et~al.(2002)Kim, Suh \& Lee}]{kim2002document}
\bibinfo{author}{Kim, J.}, \bibinfo{author}{Suh, W.}, \& \bibinfo{author}{Lee, H.} (\bibinfo{year}{2002}).
\newblock \bibinfo{title}{{Document-based workflow modeling: A case-based reasoning approach}}.
\newblock {\it \bibinfo{journal}{{Expert Systems with Applications}}\/},  {\it \bibinfo{volume}{23}\/}, \bibinfo{pages}{77--93}.
\bibitem[{Klijn \& Fahland(2020)}]{klijn2020identifying}
\bibinfo{author}{Klijn, E.~L.}, \& \bibinfo{author}{Fahland, D.} (\bibinfo{year}{2020}).
\newblock \bibinfo{title}{Identifying and reducing errors in remaining time prediction due to inter-case dynamics}.
\newblock In {\it \bibinfo{booktitle}{Proceedings of the 2nd International Conference on Process Mining}\/} (pp. \bibinfo{pages}{25--32}).
\newblock \bibinfo{organization}{IEEE}.
\bibitem[{Ko \& Comuzzi(2023)}]{koj2023anomaly}
\bibinfo{author}{Ko, J.}, \& \bibinfo{author}{Comuzzi, M.} (\bibinfo{year}{2023}).
\newblock \bibinfo{title}{A systematic review of anomaly detection for business process event logs}.
\newblock {\it \bibinfo{journal}{Business \& Information Systems Engineering}\/}, .
\bibitem[{Kotsiantis et~al.(2006)Kotsiantis, Zaharakis \& Pintelas}]{Kotsiantis2006}
\bibinfo{author}{Kotsiantis, S.~B.}, \bibinfo{author}{Zaharakis, I.~D.}, \& \bibinfo{author}{Pintelas, P.~E.} (\bibinfo{year}{2006}).
\newblock \bibinfo{title}{{Machine learning: A review of classification and combining techniques}}.
\newblock {\it \bibinfo{journal}{Artificial Intelligence Review}\/},  {\it \bibinfo{volume}{26}\/}, \bibinfo{pages}{159--190}. \DOIprefix\doi{10.1007/s10462-007-9052-3}.
\bibitem[{Krajsic \& Franczyk(2020)}]{krajsic2020lambda}
\bibinfo{author}{Krajsic, P.}, \& \bibinfo{author}{Franczyk, B.} (\bibinfo{year}{2020}).
\newblock \bibinfo{title}{{Lambda architecture for anomaly detection in online process mining using autoencoders}}.
\newblock In {\it \bibinfo{booktitle}{{Proceeding of the 12th International Conference on Computational Collective Intelligence}}\/} (pp. \bibinfo{pages}{579--589}).
\newblock \bibinfo{publisher}{Springer}.
\bibitem[{Krajsic \& Franczyk(2021)}]{krajsic2021variational}
\bibinfo{author}{Krajsic, P.}, \& \bibinfo{author}{Franczyk, B.} (\bibinfo{year}{2021}).
\newblock \bibinfo{title}{{Variational autoencoder for anomaly detection in event data in online process mining}}.
\newblock In {\it \bibinfo{booktitle}{{Proceedings of the 23rd International Conferences on Enterprise Information Systems}}\/} (pp. \bibinfo{pages}{567--574}).
\newblock \bibinfo{publisher}{SciTePress}.
\bibitem[{Kratsch et~al.(2022)Kratsch, K{\"o}nig \& R{\"o}glinger}]{kratsch2022shedding}
\bibinfo{author}{Kratsch, W.}, \bibinfo{author}{K{\"o}nig, F.}, \& \bibinfo{author}{R{\"o}glinger, M.} (\bibinfo{year}{2022}).
\newblock \bibinfo{title}{{Shedding light on blind spots -- Developing a reference architecture to leverage video data for process mining}}.
\newblock {\it \bibinfo{journal}{Decision Support Systems}\/},  {\it \bibinfo{volume}{158}\/}, \bibinfo{pages}{113794}.
\bibitem[{Krippendorff(2018)}]{krippendorff2018content}
\bibinfo{author}{Krippendorff, K.} (\bibinfo{year}{2018}).
\newblock {\it \bibinfo{title}{{Content Analysis: An Introduction to its Methodology}}\/}.
\newblock \bibinfo{publisher}{Sage Publications}.
\bibitem[{Lakshmanan et~al.(2015)Lakshmanan, Shamsi, Doganata, Unuvar \& Khalaf}]{lakshmanan2015markov}
\bibinfo{author}{Lakshmanan, G.~T.}, \bibinfo{author}{Shamsi, D.}, \bibinfo{author}{Doganata, Y.~N.}, \bibinfo{author}{Unuvar, M.}, \& \bibinfo{author}{Khalaf, R.} (\bibinfo{year}{2015}).
\newblock \bibinfo{title}{A markov prediction model for data-driven semi-structured business processes}.
\newblock {\it \bibinfo{journal}{Knowledge and Information Systems}\/},  {\it \bibinfo{volume}{42}\/}, \bibinfo{pages}{97--126}.
\bibitem[{Lamma et~al.(2007{\natexlab{a}})Lamma, Mello, Montali, Riguzzi \& Storari}]{lamma2007inducing}
\bibinfo{author}{Lamma, E.}, \bibinfo{author}{Mello, P.}, \bibinfo{author}{Montali, M.}, \bibinfo{author}{Riguzzi, F.}, \& \bibinfo{author}{Storari, S.} (\bibinfo{year}{2007}{\natexlab{a}}).
\newblock \bibinfo{title}{Inducing declarative logic-based models from labeled traces}.
\newblock In {\it \bibinfo{booktitle}{{Proceedings of the 5th International Conference on Business Process Management}}\/} (pp. \bibinfo{pages}{344--359}).
\newblock \bibinfo{publisher}{Springer}.
\bibitem[{Lamma et~al.(2007{\natexlab{b}})Lamma, Mello, Riguzzi \& Storari}]{lamma2007applying}
\bibinfo{author}{Lamma, E.}, \bibinfo{author}{Mello, P.}, \bibinfo{author}{Riguzzi, F.}, \& \bibinfo{author}{Storari, S.} (\bibinfo{year}{2007}{\natexlab{b}}).
\newblock \bibinfo{title}{Applying inductive logic programming to process mining}.
\newblock In {\it \bibinfo{booktitle}{{Proceedings of the 17th International Conference on Inductive Logic Programming}}\/} (pp. \bibinfo{pages}{132--146}).
\newblock \bibinfo{publisher}{Springer}.
\bibitem[{Landis \& Koch(1977)}]{landis1977application}
\bibinfo{author}{Landis, J.~R.}, \& \bibinfo{author}{Koch, G.~G.} (\bibinfo{year}{1977}).
\newblock \bibinfo{title}{An application of hierarchical kappa-type statistics in the assessment of majority agreement among multiple observers}.
\newblock {\it \bibinfo{journal}{Biometrics}\/},  (pp. \bibinfo{pages}{363--374}).
\bibitem[{LeCun et~al.(2015)LeCun, Bengio \& Hinton}]{lecun2015deep}
\bibinfo{author}{LeCun, Y.}, \bibinfo{author}{Bengio, Y.}, \& \bibinfo{author}{Hinton, G.} (\bibinfo{year}{2015}).
\newblock \bibinfo{title}{Deep learning}.
\newblock {\it \bibinfo{journal}{Nature}\/},  {\it \bibinfo{volume}{521}\/}, \bibinfo{pages}{436--444}. \DOIprefix\doi{10.1038/nature14539}.
\bibitem[{Lee et~al.(2022)Lee, Lu \& Reijers}]{lee2022analysis}
\bibinfo{author}{Lee, S.}, \bibinfo{author}{Lu, X.}, \& \bibinfo{author}{Reijers, H.~A.} (\bibinfo{year}{2022}).
\newblock \bibinfo{title}{{The analysis of online event streams: Predicting the next activity for anomaly detection}}.
\newblock In {\it \bibinfo{booktitle}{{Proceeding of the 16th International Conference on Research Challenges in Information Science}}\/} (pp. \bibinfo{pages}{248--264}).
\newblock \bibinfo{publisher}{Springer}.
\bibitem[{Lee et~al.(2021)Lee, Burattin, Munoz-Gama \& Sepulveda}]{lee2021orientation}
\bibinfo{author}{Lee, W. L.~J.}, \bibinfo{author}{Burattin, A.}, \bibinfo{author}{Munoz-Gama, J.}, \& \bibinfo{author}{Sepulveda, M.} (\bibinfo{year}{2021}).
\newblock \bibinfo{title}{{Orientation and conformance: A HMM-based approach to online conformance checking}}.
\newblock {\it \bibinfo{journal}{Information Systems}\/},  {\it \bibinfo{volume}{102}\/}, \bibinfo{pages}{101674}. \DOIprefix\doi{10.1016/j.is.2020.101674}.
\bibitem[{Lee et~al.(2018)Lee, Parra, Munoz-Gama \& Sepulveda}]{lee2018predicting}
\bibinfo{author}{Lee, W. L.~J.}, \bibinfo{author}{Parra, D.}, \bibinfo{author}{Munoz-Gama, J.}, \& \bibinfo{author}{Sepulveda, M.} (\bibinfo{year}{2018}).
\newblock \bibinfo{title}{Predicting process behavior meets factorization machines}.
\newblock {\it \bibinfo{journal}{Expert Systems with Applications}\/},  {\it \bibinfo{volume}{112}\/}, \bibinfo{pages}{87--98}.
\bibitem[{Leno et~al.(2018)Leno, Dumas \& Maggi}]{leno2018correlating}
\bibinfo{author}{Leno, V.}, \bibinfo{author}{Dumas, M.}, \& \bibinfo{author}{Maggi, F.~M.} (\bibinfo{year}{2018}).
\newblock \bibinfo{title}{Correlating activation and target conditions in data-aware declarative process discovery}.
\newblock In {\it \bibinfo{booktitle}{{Proceedings of the 16th International Conference on Business Process Management}}\/} (pp. \bibinfo{pages}{176--193}).
\newblock \bibinfo{publisher}{Springer}.
\bibitem[{Leno et~al.(2020)Leno, Dumas, Maggi, La~Rosa \& Polyvyanyy}]{leno2020automated}
\bibinfo{author}{Leno, V.}, \bibinfo{author}{Dumas, M.}, \bibinfo{author}{Maggi, F.~M.}, \bibinfo{author}{La~Rosa, M.}, \& \bibinfo{author}{Polyvyanyy, A.} (\bibinfo{year}{2020}).
\newblock \bibinfo{title}{Automated discovery of declarative process models with correlated data conditions}.
\newblock {\it \bibinfo{journal}{Information Systems}\/},  {\it \bibinfo{volume}{89}\/}, \bibinfo{pages}{101482}.
\bibitem[{de~Leoni et~al.(2014)de~Leoni, van~der Aalst \& Dees}]{leoni2014general}
\bibinfo{author}{de~Leoni, M.}, \bibinfo{author}{van~der Aalst, W. M.~P.}, \& \bibinfo{author}{Dees, M.} (\bibinfo{year}{2014}).
\newblock \bibinfo{title}{A general framework for correlating business process characteristics}.
\newblock In {\it \bibinfo{booktitle}{{Proceedings of the 12th International Conference on Business Process Management}}\/} (pp. \bibinfo{pages}{250--266}).
\newblock \bibinfo{publisher}{Springer}.
\bibitem[{de~Leoni et~al.(2016)de~Leoni, van~der Aalst \& Dees}]{leoni2016general}
\bibinfo{author}{de~Leoni, M.}, \bibinfo{author}{van~der Aalst, W. M.~P.}, \& \bibinfo{author}{Dees, M.} (\bibinfo{year}{2016}).
\newblock \bibinfo{title}{A general process mining framework for correlating, predicting and clustering dynamic behavior based on event logs}.
\newblock {\it \bibinfo{journal}{Information Systems}\/},  {\it \bibinfo{volume}{56}\/}, \bibinfo{pages}{235--257}.
\bibitem[{de~Leoni et~al.(2020)de~Leoni, Dees \& Reulink}]{de2020design}
\bibinfo{author}{de~Leoni, M.}, \bibinfo{author}{Dees, M.}, \& \bibinfo{author}{Reulink, L.} (\bibinfo{year}{2020}).
\newblock \bibinfo{title}{Design and evaluation of a process-aware recommender system based on prescriptive analytics}.
\newblock In {\it \bibinfo{booktitle}{{Proceedings of the 2nd International Conference on Process Mining}}\/} (pp. \bibinfo{pages}{9--16}).
\newblock \bibinfo{publisher}{IEEE}.
\bibitem[{de~Leoni et~al.(2013)de~Leoni, Dumas \& Garc{\'\i}a-Ba{\~n}uelos}]{leoni2013discovering}
\bibinfo{author}{de~Leoni, M.}, \bibinfo{author}{Dumas, M.}, \& \bibinfo{author}{Garc{\'\i}a-Ba{\~n}uelos, L.} (\bibinfo{year}{2013}).
\newblock \bibinfo{title}{Discovering branching conditions from business process execution logs}.
\newblock In {\it \bibinfo{booktitle}{{Proceedings of the 16th International Conference on Fundamental Approaches to Software Engineering}}\/} (pp. \bibinfo{pages}{114--129}).
\newblock \bibinfo{publisher}{Springer}.
\bibitem[{Leontjeva et~al.(2016)Leontjeva, Conforti, Di~Francescomarino, Dumas \& Maggi}]{leontjeva2016complex}
\bibinfo{author}{Leontjeva, A.}, \bibinfo{author}{Conforti, R.}, \bibinfo{author}{Di~Francescomarino, C.}, \bibinfo{author}{Dumas, M.}, \& \bibinfo{author}{Maggi, F.~M.} (\bibinfo{year}{2016}).
\newblock \bibinfo{title}{Complex symbolic sequence encodings for predictive monitoring of business processes}.
\newblock In {\it \bibinfo{booktitle}{{Proceedings of the 14th International Conference on Business Process Management}}\/} (pp. \bibinfo{pages}{297--313}).
\newblock \bibinfo{publisher}{Springer}.
\bibitem[{Lewis(1998)}]{lewis1998naive}
\bibinfo{author}{Lewis, D.~D.} (\bibinfo{year}{1998}).
\newblock \bibinfo{title}{{Naive (Bayes) at forty: The independence assumption in information retrieval}}.
\newblock In {\it \bibinfo{booktitle}{{Proceedings of the 10th European Conference on Machine Learning}}\/} (pp. \bibinfo{pages}{4--15}).
\newblock \bibinfo{publisher}{Springer}.
\bibitem[{Li et~al.(2010)Li, Reichert \& Wombacher}]{li2010minadept}
\bibinfo{author}{Li, C.}, \bibinfo{author}{Reichert, M.}, \& \bibinfo{author}{Wombacher, A.} (\bibinfo{year}{2010}).
\newblock \bibinfo{title}{The minadept clustering approach for discovering reference process models out of process variants}.
\newblock {\it \bibinfo{journal}{International Journal of Cooperative Information Systems}\/},  {\it \bibinfo{volume}{19}\/}, \bibinfo{pages}{159--203}.
\bibitem[{Li et~al.(2018)Li, de~Murillas, de~Carvalho \& van~der Aalst}]{li2018extracting}
\bibinfo{author}{Li, G.}, \bibinfo{author}{de~Murillas, E. G.~L.}, \bibinfo{author}{de~Carvalho, R.~M.}, \& \bibinfo{author}{van~der Aalst, W. M.~P.} (\bibinfo{year}{2018}).
\newblock \bibinfo{title}{{Extracting object-centric event logs to support process mining on databases}}.
\newblock In {\it \bibinfo{booktitle}{{Proceedings of the 30th International Conference on Advanced Information Systems Engineering (Forum)}}\/} (pp. \bibinfo{pages}{182--199}).
\newblock \bibinfo{publisher}{Springer}.
\bibitem[{Liu \& Motoda(1998)}]{liu1998feature}
\bibinfo{author}{Liu, H.}, \& \bibinfo{author}{Motoda, H.} (\bibinfo{year}{1998}).
\newblock {\it \bibinfo{title}{{Feature Extraction, Construction and Selection: A Data Mining Perspective}}\/} volume \bibinfo{volume}{453}.
\newblock \bibinfo{publisher}{Springer}.
\bibitem[{Liu et~al.(2012)Liu, Cheng \& Ni}]{liu2012mining}
\bibinfo{author}{Liu, T.}, \bibinfo{author}{Cheng, Y.}, \& \bibinfo{author}{Ni, Z.} (\bibinfo{year}{2012}).
\newblock \bibinfo{title}{Mining event logs to support workflow resource allocation}.
\newblock {\it \bibinfo{journal}{Knowledge-Based Systems}\/},  {\it \bibinfo{volume}{35}\/}, \bibinfo{pages}{320--331}.
\bibitem[{Liu et~al.(2008)Liu, Wang, Yang \& Sun}]{liu2008semi}
\bibinfo{author}{Liu, Y.}, \bibinfo{author}{Wang, J.}, \bibinfo{author}{Yang, Y.}, \& \bibinfo{author}{Sun, J.} (\bibinfo{year}{2008}).
\newblock \bibinfo{title}{A semi-automatic approach for workflow staff assignment}.
\newblock {\it \bibinfo{journal}{Computers in Industry}\/},  {\it \bibinfo{volume}{59}\/}, \bibinfo{pages}{463--476}.
\bibitem[{Lombard et~al.(2006)Lombard, Snyder-Duch \& Bracken}]{lombard2002}
\bibinfo{author}{Lombard, M.}, \bibinfo{author}{Snyder-Duch, J.}, \& \bibinfo{author}{Bracken, C.~C.} (\bibinfo{year}{2006}).
\newblock \bibinfo{title}{{Content analysis in mass communication: Assessment and reporting of intercoder reliability}}.
\newblock {\it \bibinfo{journal}{Human Communication Research}\/},  {\it \bibinfo{volume}{28}\/}, \bibinfo{pages}{587--604}. \DOIprefix\doi{10.1111/j.1468-2958.2002.tb00826.x}. \href{http://arxiv.org/abs/https://academic.oup.com/hcr/article-pdf/28/4/587/22337849/jhumcom0587.pdf}{\tt arXiv:https://academic.oup.com/hcr/article-pdf/28/4/587/22337849/jhumcom0587.pdf}.
\bibitem[{L{\'o}pez et~al.(2021)L{\'o}pez, Str{\o}msted, Niyodusenga \& Marquard}]{lopez2021declarative}
\bibinfo{author}{L{\'o}pez, H.~A.}, \bibinfo{author}{Str{\o}msted, R.}, \bibinfo{author}{Niyodusenga, J.-M.}, \& \bibinfo{author}{Marquard, M.} (\bibinfo{year}{2021}).
\newblock \bibinfo{title}{Declarative process discovery: Linking process and textual views}.
\newblock In {\it \bibinfo{booktitle}{{Proceedings of the 33rd International Conference on Advanced Information Systems Engineering (Forum)}}\/} (pp. \bibinfo{pages}{109--117}).
\newblock \bibinfo{publisher}{Springer}.
\bibitem[{Lu et~al.(2016)Lu, Zeng \& Duan}]{lu2016synchronization}
\bibinfo{author}{Lu, F.}, \bibinfo{author}{Zeng, Q.}, \& \bibinfo{author}{Duan, H.} (\bibinfo{year}{2016}).
\newblock \bibinfo{title}{Synchronization-core-based discovery of processes with decomposable cyclic dependencies}.
\newblock {\it \bibinfo{journal}{ACM Transactions on Knowledge Discovery from Data}\/},  {\it \bibinfo{volume}{10}\/}, \bibinfo{pages}{1--29}.
\bibitem[{Lundberg \& Lee(2017)}]{lundberg2017unified}
\bibinfo{author}{Lundberg, S.~M.}, \& \bibinfo{author}{Lee, S.-I.} (\bibinfo{year}{2017}).
\newblock \bibinfo{title}{A unified approach to interpreting model predictions}.
\newblock In {\it \bibinfo{booktitle}{{Proceedings of the 31st Conference on Neural Information Processing Systems}}\/} (pp. \bibinfo{pages}{4765--4774}).
\newblock \bibinfo{publisher}{ACM}.
\bibitem[{Maggi et~al.(2012)Maggi, Bose \& van~der Aalst}]{maggi2012efficient}
\bibinfo{author}{Maggi, F.~M.}, \bibinfo{author}{Bose, R.~J.}, \& \bibinfo{author}{van~der Aalst, W. M.~P.} (\bibinfo{year}{2012}).
\newblock \bibinfo{title}{Efficient discovery of understandable declarative process models from event logs}.
\newblock In {\it \bibinfo{booktitle}{{Proceedings of the 24th International Conference on Advanced Information Systems Engineering}}\/} (pp. \bibinfo{pages}{270--285}).
\newblock \bibinfo{publisher}{Springer}.
\bibitem[{Maggi et~al.(2018)Maggi, Di~Ciccio, Di~Francescomarino \& Kala}]{maggi2018parallel}
\bibinfo{author}{Maggi, F.~M.}, \bibinfo{author}{Di~Ciccio, C.}, \bibinfo{author}{Di~Francescomarino, C.}, \& \bibinfo{author}{Kala, T.} (\bibinfo{year}{2018}).
\newblock \bibinfo{title}{Parallel algorithms for the automated discovery of declarative process models}.
\newblock {\it \bibinfo{journal}{Information Systems}\/},  {\it \bibinfo{volume}{74}\/}, \bibinfo{pages}{136--152}.
\bibitem[{Maggi et~al.(2014)Maggi, Di~Francescomarino, Dumas \& Ghidini}]{maggi2014predictive}
\bibinfo{author}{Maggi, F.~M.}, \bibinfo{author}{Di~Francescomarino, C.}, \bibinfo{author}{Dumas, M.}, \& \bibinfo{author}{Ghidini, C.} (\bibinfo{year}{2014}).
\newblock \bibinfo{title}{Predictive monitoring of business processes}.
\newblock In {\it \bibinfo{booktitle}{{Proceedings of the 26th International Conference on Advanced Information Systems Engineering}}\/} (pp. \bibinfo{pages}{457--472}).
\newblock \bibinfo{publisher}{Springer}.
\bibitem[{Maier et~al.(2019)Maier, Syben, Stimpel, W{\"u}rfl, Hoffmann, Schebesch, Fu, Mill, Kling \& Christiansen}]{maier2019learning}
\bibinfo{author}{Maier, A.~K.}, \bibinfo{author}{Syben, C.}, \bibinfo{author}{Stimpel, B.}, \bibinfo{author}{W{\"u}rfl, T.}, \bibinfo{author}{Hoffmann, M.}, \bibinfo{author}{Schebesch, F.}, \bibinfo{author}{Fu, W.}, \bibinfo{author}{Mill, L.}, \bibinfo{author}{Kling, L.}, \& \bibinfo{author}{Christiansen, S.} (\bibinfo{year}{2019}).
\newblock \bibinfo{title}{Learning with known operators reduces maximum error bounds}.
\newblock {\it \bibinfo{journal}{Nature Machine Intelligence}\/},  {\it \bibinfo{volume}{1}\/}, \bibinfo{pages}{373--380}.
\bibitem[{Maita et~al.(2015)Maita, Martins, Paz, Peres \& Fantinato}]{maita2015process}
\bibinfo{author}{Maita, A. R.~C.}, \bibinfo{author}{Martins, L.~C.}, \bibinfo{author}{Paz, C. R.~L.}, \bibinfo{author}{Peres, S.~M.}, \& \bibinfo{author}{Fantinato, M.} (\bibinfo{year}{2015}).
\newblock \bibinfo{title}{Process mining through artificial neural networks and support vector machines: A systematic literature review}.
\newblock {\it \bibinfo{journal}{Business Process Management Journal}\/},  {\it \bibinfo{volume}{21}\/}, \bibinfo{pages}{1391--1415}.
\bibitem[{Mannhardt et~al.(2016)Mannhardt, de~Leoni, Reijers \& van~der Aalst}]{mannhardt2016decision}
\bibinfo{author}{Mannhardt, F.}, \bibinfo{author}{de~Leoni, M.}, \bibinfo{author}{Reijers, H.~A.}, \& \bibinfo{author}{van~der Aalst, W. M.~P.} (\bibinfo{year}{2016}).
\newblock \bibinfo{title}{{Decision mining revisited -- Discovering overlapping rules}}.
\newblock In {\it \bibinfo{booktitle}{{Proceedings of the 28th International Conference on Advanced Information Systems Engineering}}\/} (pp. \bibinfo{pages}{377--392}).
\newblock \bibinfo{publisher}{Springer}.
\bibitem[{M{\'a}rquez-Chamorro et~al.(2017{\natexlab{a}})M{\'a}rquez-Chamorro, Resinas \& Ruiz-Cortés}]{marquez2017predictive}
\bibinfo{author}{M{\'a}rquez-Chamorro, A.~E.}, \bibinfo{author}{Resinas, M.}, \& \bibinfo{author}{Ruiz-Cortés, A.} (\bibinfo{year}{2017}{\natexlab{a}}).
\newblock \bibinfo{title}{Predictive monitoring of business processes: A survey}.
\newblock {\it \bibinfo{journal}{IEEE Transactions on Services Computing}\/},  {\it \bibinfo{volume}{11}\/}, \bibinfo{pages}{962--977}. \DOIprefix\doi{10.1109/TSC.2017.2772256}.
\bibitem[{M{\'a}rquez-Chamorro et~al.(2017{\natexlab{b}})M{\'a}rquez-Chamorro, Resinas, Ruiz-Cortés \& Toro}]{marquez2017run}
\bibinfo{author}{M{\'a}rquez-Chamorro, A.~E.}, \bibinfo{author}{Resinas, M.}, \bibinfo{author}{Ruiz-Cortés, A.}, \& \bibinfo{author}{Toro, M.} (\bibinfo{year}{2017}{\natexlab{b}}).
\newblock \bibinfo{title}{Run-time prediction of business process indicators using evolutionary decision rules}.
\newblock {\it \bibinfo{journal}{Expert Systems with Applications}\/},  {\it \bibinfo{volume}{87}\/}, \bibinfo{pages}{1--14}.
\bibitem[{Martens et~al.(2014)Martens, Fettke \& Loos}]{martens2014genetic}
\bibinfo{author}{Martens, A.}, \bibinfo{author}{Fettke, P.}, \& \bibinfo{author}{Loos, P.} (\bibinfo{year}{2014}).
\newblock \bibinfo{title}{A genetic algorithm for the inductive derivation of reference models using minimal graph-edit distance applied to real-world business process data}.
\newblock In {\it \bibinfo{booktitle}{{Proceedings of the 2014 Multikonferenz Wirtschaftsinformatik}}\/} (pp. \bibinfo{pages}{1613--1626}).
\newblock \bibinfo{publisher}{Universit{\"a}t Paderborn}.
\bibitem[{M{\u{a}}ru{\c{s}}ter et~al.(2002)M{\u{a}}ru{\c{s}}ter, Weijters, van~der Aalst \& Van Den~Bosch}]{maruster2002process}
\bibinfo{author}{M{\u{a}}ru{\c{s}}ter, L.}, \bibinfo{author}{Weijters, A.~J.}, \bibinfo{author}{van~der Aalst, W. M.~P.}, \& \bibinfo{author}{Van Den~Bosch, A.} (\bibinfo{year}{2002}).
\newblock \bibinfo{title}{Process mining: Discovering direct successors in process logs}.
\newblock In {\it \bibinfo{booktitle}{{Proceedings of the 5th International Conference on Discovery Science}}\/} (pp. \bibinfo{pages}{364--373}).
\newblock \bibinfo{publisher}{Springer}.
\bibitem[{M{\u{a}}ru{\c{s}}ter et~al.(2006)M{\u{a}}ru{\c{s}}ter, Weijters, van~der Aalst \& Van Den~Bosch}]{muarucster2006rule}
\bibinfo{author}{M{\u{a}}ru{\c{s}}ter, L.}, \bibinfo{author}{Weijters, A.~J.}, \bibinfo{author}{van~der Aalst, W. M.~P.}, \& \bibinfo{author}{Van Den~Bosch, A.} (\bibinfo{year}{2006}).
\newblock \bibinfo{title}{A rule-based approach for process discovery: Dealing with noise and imbalance in process logs}.
\newblock {\it \bibinfo{journal}{{Data Mining and Knowledge Discovery}}\/},  {\it \bibinfo{volume}{13}\/}, \bibinfo{pages}{67--87}.
\bibitem[{McCoy(2002)}]{mccoy2002business}
\bibinfo{author}{McCoy, D.~W.} (\bibinfo{year}{2002}).
\newblock \bibinfo{title}{{Business Activity Monitoring: Calm Before the Storm}}.
\newblock \URLprefix \url{https://www.gartner.com/resources/105500/105562/105562.pdf}.
\bibitem[{Mehdiyev et~al.(2020)Mehdiyev, Evermann \& Fettke}]{mehdiyev2020novel}
\bibinfo{author}{Mehdiyev, N.}, \bibinfo{author}{Evermann, J.}, \& \bibinfo{author}{Fettke, P.} (\bibinfo{year}{2020}).
\newblock \bibinfo{title}{A novel business process prediction model using a deep learning method}.
\newblock {\it \bibinfo{journal}{Business \& Information Systems Engineering}\/},  {\it \bibinfo{volume}{62}\/}, \bibinfo{pages}{143--157}. \DOIprefix\doi{10.1007/s12599-018-0551-3}.
\bibitem[{Mendling et~al.(2018)Mendling, Decker, Hull, Reijers \& Weber}]{mendling2018machine}
\bibinfo{author}{Mendling, J.}, \bibinfo{author}{Decker, G.}, \bibinfo{author}{Hull, R.}, \bibinfo{author}{Reijers, H.~A.}, \& \bibinfo{author}{Weber, I.} (\bibinfo{year}{2018}).
\newblock \bibinfo{title}{How do machine learning, robotic process automation, and blockchains affect the human factor in business process management?}
\newblock {\it \bibinfo{journal}{Communications of the Association for Information Systems}\/},  {\it \bibinfo{volume}{43}\/}, \bibinfo{pages}{297--320}. \DOIprefix\doi{10.17705/1CAIS.04319}.
\bibitem[{Metzger et~al.(2020)Metzger, Kley \& Palm}]{metzger2020triggering}
\bibinfo{author}{Metzger, A.}, \bibinfo{author}{Kley, T.}, \& \bibinfo{author}{Palm, A.} (\bibinfo{year}{2020}).
\newblock \bibinfo{title}{Triggering proactive business process adaptations via online reinforcement learning}.
\newblock In {\it \bibinfo{booktitle}{{Proceedings of the 18th International Conference on Business Process Management}}\/} (pp. \bibinfo{pages}{273--290}).
\newblock \bibinfo{publisher}{Springer}.
\bibitem[{Metzger et~al.(2019)Metzger, Neubauer, Bohn \& Pohl}]{metzger2019proactive}
\bibinfo{author}{Metzger, A.}, \bibinfo{author}{Neubauer, A.}, \bibinfo{author}{Bohn, P.}, \& \bibinfo{author}{Pohl, K.} (\bibinfo{year}{2019}).
\newblock \bibinfo{title}{Proactive process adaptation using deep learning ensembles}.
\newblock In {\it \bibinfo{booktitle}{{Proceedings of the 31st International Conference on Advanced Information Systems Engineering}}\/} (pp. \bibinfo{pages}{547--562}).
\newblock \bibinfo{publisher}{Springer}.
\bibitem[{Mitchell(1997)}]{mitchell.1997}
\bibinfo{author}{Mitchell, T.~M.} (\bibinfo{year}{1997}).
\newblock {\it \bibinfo{title}{{Machine Learning}}\/}.
\newblock \bibinfo{publisher}{McGraw-Hill Education}.
\bibitem[{Mohri et~al.(2018)Mohri, Rostamizadeh \& Talwalkar}]{mohri2018foundations}
\bibinfo{author}{Mohri, M.}, \bibinfo{author}{Rostamizadeh, A.}, \& \bibinfo{author}{Talwalkar, A.} (\bibinfo{year}{2018}).
\newblock {\it \bibinfo{title}{{Foundations of Machine Learning}}\/}.
\newblock \bibinfo{publisher}{MIT Press}.
\bibitem[{Montani \& Leonardi(2012)}]{montani2012retrieval}
\bibinfo{author}{Montani, S.}, \& \bibinfo{author}{Leonardi, G.} (\bibinfo{year}{2012}).
\newblock \bibinfo{title}{{Retrieval and clustering for business process monitoring: Results and improvements}}.
\newblock In {\it \bibinfo{booktitle}{{Proceedings of the 20th International Conference on Case-Based Reasoning}}\/} (pp. \bibinfo{pages}{269--283}).
\newblock \bibinfo{publisher}{Springer}.
\bibitem[{Mothilal et~al.(2020)Mothilal, Sharma \& Tan}]{mothilal2020explaining}
\bibinfo{author}{Mothilal, R.~K.}, \bibinfo{author}{Sharma, A.}, \& \bibinfo{author}{Tan, C.} (\bibinfo{year}{2020}).
\newblock \bibinfo{title}{Explaining machine learning classifiers through diverse counterfactual explanations}.
\newblock In {\it \bibinfo{booktitle}{{Proceedings of the 2020 Conference on Fairness, Accountability, and Transparency}}\/} (pp. \bibinfo{pages}{607--617}).
\newblock \bibinfo{publisher}{ACM}.
\bibitem[{zur Muehlen \& Ho(2005)}]{Muehlen2006}
\bibinfo{author}{zur Muehlen, M.}, \& \bibinfo{author}{Ho, D. T.-Y.} (\bibinfo{year}{2005}).
\newblock \bibinfo{title}{{Risk Management in the BPM Lifecycle}}.
\newblock In {\it \bibinfo{booktitle}{{Proceedings of the 3rd International Conference on Business Process Management (Workshops)}}\/} (pp. \bibinfo{pages}{454--466}).
\newblock \bibinfo{publisher}{Springer}.
\newblock \DOIprefix\doi{10.1007/11678564\_42}.
\bibitem[{Murphy(2012)}]{murphy2012machine}
\bibinfo{author}{Murphy, K.~P.} (\bibinfo{year}{2012}).
\newblock {\it \bibinfo{title}{Machine learning: a probabilistic perspective}\/}.
\newblock \bibinfo{publisher}{MIT press}.
\bibitem[{Mustansir et~al.(2022)Mustansir, Shahzad \& Malik}]{Mustansir2022}
\bibinfo{author}{Mustansir, A.}, \bibinfo{author}{Shahzad, K.}, \& \bibinfo{author}{Malik, M.~K.} (\bibinfo{year}{2022}).
\newblock \bibinfo{title}{Towards automatic business process redesign: {An} {NLP} based approach to extract redesign suggestions}.
\newblock {\it \bibinfo{journal}{Automated Software Engineering}\/},  {\it \bibinfo{volume}{29}\/}, \bibinfo{pages}{1--24}. \DOIprefix\doi{10.1007/s10515-021-00316-8}.
\bibitem[{Neu et~al.(2021)Neu, Lahann \& Fettke}]{Neu2021}
\bibinfo{author}{Neu, D.~A.}, \bibinfo{author}{Lahann, J.}, \& \bibinfo{author}{Fettke, P.} (\bibinfo{year}{2021}).
\newblock \bibinfo{title}{A systematic literature review on state-of-the-art deep learning methods for process prediction}.
\newblock {\it \bibinfo{journal}{Artificial Intelligence Review}\/},  {\it \bibinfo{volume}{55}\/}, \bibinfo{pages}{801--827}. \DOIprefix\doi{10.1007/s10462-021-09960-8}.
\bibitem[{Neuendorf(2017)}]{neuendorf2017content}
\bibinfo{author}{Neuendorf, K.~A.} (\bibinfo{year}{2017}).
\newblock {\it \bibinfo{title}{{The Content Analysis Guidebook}}\/}.
\newblock \bibinfo{publisher}{Sage Publications}.
\bibitem[{Nguyen et~al.(2019)Nguyen, Lee, Kim, Ko \& Comuzzi}]{nguyen2019autoencoders}
\bibinfo{author}{Nguyen, H. T.~C.}, \bibinfo{author}{Lee, S.}, \bibinfo{author}{Kim, J.}, \bibinfo{author}{Ko, J.}, \& \bibinfo{author}{Comuzzi, M.} (\bibinfo{year}{2019}).
\newblock \bibinfo{title}{{Autoencoders for improving quality of process event logs}}.
\newblock {\it \bibinfo{journal}{Expert Systems with Applications}\/},  {\it \bibinfo{volume}{131}\/}, \bibinfo{pages}{132--147}. \DOIprefix\doi{10.1016/j.eswa.2019.04.052}.
\bibitem[{Nguyen et~al.(2016)Nguyen, Slominski, Muthusamy, Ishakian \& Nahrstedt}]{nguyen2016process}
\bibinfo{author}{Nguyen, P.}, \bibinfo{author}{Slominski, A.}, \bibinfo{author}{Muthusamy, V.}, \bibinfo{author}{Ishakian, V.}, \& \bibinfo{author}{Nahrstedt, K.} (\bibinfo{year}{2016}).
\newblock \bibinfo{title}{Process trace clustering: A heterogeneous information network approach}.
\newblock In {\it \bibinfo{booktitle}{{Proceedings of the 2016 SIAM International Conference on Data Mining}}\/} (pp. \bibinfo{pages}{279--287}).
\newblock \bibinfo{publisher}{SIAM}.
\bibitem[{Nichol et~al.(2018)Nichol, Achiam \& Schulman}]{nichol2018first}
\bibinfo{author}{Nichol, A.}, \bibinfo{author}{Achiam, J.}, \& \bibinfo{author}{Schulman, J.} (\bibinfo{year}{2018}).
\newblock \bibinfo{title}{{On first-order meta-learning algorithms}}.
\newblock {\it \bibinfo{journal}{arXiv preprint arXiv:1803.02999}\/},  (pp. \bibinfo{pages}{1--15}). \href{http://arxiv.org/abs/1803.02999}{\tt arXiv:1803.02999}.
\bibitem[{Nigam \& Caswell(2003)}]{nigam2003business}
\bibinfo{author}{Nigam, A.}, \& \bibinfo{author}{Caswell, N.~S.} (\bibinfo{year}{2003}).
\newblock \bibinfo{title}{{Business artifacts: An approach to operational specification}}.
\newblock {\it \bibinfo{journal}{IBM Systems Journal}\/},  {\it \bibinfo{volume}{42}\/}, \bibinfo{pages}{428--445}.
\bibitem[{Nolle et~al.(2018{\natexlab{a}})Nolle, Luettgen, Seeliger \& Mühlhäuser}]{nolle.2018}
\bibinfo{author}{Nolle, T.}, \bibinfo{author}{Luettgen, S.}, \bibinfo{author}{Seeliger, A.}, \& \bibinfo{author}{Mühlhäuser, M.} (\bibinfo{year}{2018}{\natexlab{a}}).
\newblock \bibinfo{title}{Analyzing business process anomalies using autoencoders}.
\newblock {\it \bibinfo{journal}{Machine Learning}\/},  {\it \bibinfo{volume}{107}\/}, \bibinfo{pages}{1875--1893}. \DOIprefix\doi{10.1007/s10994-018-5702-8}.
\bibitem[{Nolle et~al.(2019)Nolle, Luettgen, Seeliger \& Mühlhäuser}]{nolle2019binet}
\bibinfo{author}{Nolle, T.}, \bibinfo{author}{Luettgen, S.}, \bibinfo{author}{Seeliger, A.}, \& \bibinfo{author}{Mühlhäuser, M.} (\bibinfo{year}{2019}).
\newblock \bibinfo{title}{{BINet: Multi-perspective business process anomaly classification}}.
\newblock {\it \bibinfo{journal}{Information Systems}\/},  {\it \bibinfo{volume}{103}\/}, \bibinfo{pages}{101458}. \DOIprefix\doi{10.1016/j.is.2019.101458}.
\bibitem[{Nolle et~al.(2018{\natexlab{b}})Nolle, Seeliger \& M{\"u}hlh{\"a}user}]{nolle2018binet}
\bibinfo{author}{Nolle, T.}, \bibinfo{author}{Seeliger, A.}, \& \bibinfo{author}{M{\"u}hlh{\"a}user, M.} (\bibinfo{year}{2018}{\natexlab{b}}).
\newblock \bibinfo{title}{{BINet: Multivariate business process anomaly detection using deep learning}}.
\newblock In {\it \bibinfo{booktitle}{{Proceedings of the 16th International Conference on Business Process Management}}\/} (pp. \bibinfo{pages}{271--287}).
\newblock \bibinfo{publisher}{Springer}.
\bibitem[{Nolle et~al.(2020)Nolle, Seeliger, Thoma \& Mühlhäuser}]{nolle2020deepalign}
\bibinfo{author}{Nolle, T.}, \bibinfo{author}{Seeliger, A.}, \bibinfo{author}{Thoma, N.}, \& \bibinfo{author}{Mühlhäuser, M.} (\bibinfo{year}{2020}).
\newblock \bibinfo{title}{{DeepAlign: Alignment-based process anomaly correction using recurrent neural networks}}.
\newblock In {\it \bibinfo{booktitle}{{Proceedings of the 32nd International Conference on Advanced Information Systems Engineering}}\/} (pp. \bibinfo{pages}{319--333}).
\newblock \bibinfo{publisher}{Springer}.
\bibitem[{Oberdorf et~al.(2023)Oberdorf, Schaschek, Weinzierl, Stein, Matzner \& Flath}]{oberdorf2023predictive}
\bibinfo{author}{Oberdorf, F.}, \bibinfo{author}{Schaschek, M.}, \bibinfo{author}{Weinzierl, S.}, \bibinfo{author}{Stein, N.}, \bibinfo{author}{Matzner, M.}, \& \bibinfo{author}{Flath, C.~M.} (\bibinfo{year}{2023}).
\newblock \bibinfo{title}{Predictive end-to-end enterprise process network monitoring}.
\newblock {\it \bibinfo{journal}{Business \& Information Systems Engineering}\/},  (pp. \bibinfo{pages}{49--64}).
\bibitem[{Okoli(2015)}]{okoli2015guide}
\bibinfo{author}{Okoli, C.} (\bibinfo{year}{2015}).
\newblock \bibinfo{title}{A guide to conducting a standalone systematic literature review}.
\newblock {\it \bibinfo{journal}{Communications of the Association for Information Systems}\/},  {\it \bibinfo{volume}{37}\/}, \bibinfo{pages}{879--910}.
\bibitem[{OpenAI(2023)}]{openai2023}
\bibinfo{author}{OpenAI} (\bibinfo{year}{2023}).
\newblock \bibinfo{title}{Gpt-4 technical report}.
\newblock {\it \bibinfo{journal}{arXiv preprint arXiv:2303.08774v3}\/},  (pp. \bibinfo{pages}{1--100}). \href{http://arxiv.org/abs/2303.08774v3}{\tt arXiv:2303.08774v3}.
\bibitem[{Otter et~al.(2020)Otter, Medina \& Kalita}]{otter2020survey}
\bibinfo{author}{Otter, D.~W.}, \bibinfo{author}{Medina, J.~R.}, \& \bibinfo{author}{Kalita, J.~K.} (\bibinfo{year}{2020}).
\newblock \bibinfo{title}{A survey of the usages of deep learning for natural language processing}.
\newblock {\it \bibinfo{journal}{IEEE Transactions on Neural Networks and Learning Systems}\/},  {\it \bibinfo{volume}{32}\/}, \bibinfo{pages}{604--624}.
\bibitem[{Pan \& Yang(2009)}]{pan2009survey}
\bibinfo{author}{Pan, S.~J.}, \& \bibinfo{author}{Yang, Q.} (\bibinfo{year}{2009}).
\newblock \bibinfo{title}{A survey on transfer learning}.
\newblock {\it \bibinfo{journal}{{IEEE Transactions on Knowledge and Data Engineering}}\/},  {\it \bibinfo{volume}{22}\/}, \bibinfo{pages}{1345--1359}. \DOIprefix\doi{10.1109/TKDE.2009.191}.
\bibitem[{Park \& Song(2019)}]{park2019prediction}
\bibinfo{author}{Park, G.}, \& \bibinfo{author}{Song, M.} (\bibinfo{year}{2019}).
\newblock \bibinfo{title}{Prediction-based resource allocation using {LSTM} and minimum cost and maximum flow algorithm}.
\newblock In {\it \bibinfo{booktitle}{{Proceedings of the 1st International Conference on Process Mining}}\/} (pp. \bibinfo{pages}{121--128}).
\newblock \bibinfo{publisher}{IEEE}.
\bibitem[{Park \& Song(2020)}]{park2020predicting}
\bibinfo{author}{Park, G.}, \& \bibinfo{author}{Song, M.} (\bibinfo{year}{2020}).
\newblock \bibinfo{title}{Predicting performances in business processes using deep neural networks}.
\newblock {\it \bibinfo{journal}{Decision Support Systems}\/},  {\it \bibinfo{volume}{129}\/}, \bibinfo{pages}{113191}.
\bibitem[{Paré et~al.(2015)Paré, Trudel, Jaana \& Kitsiou}]{pare.2015}
\bibinfo{author}{Paré, G.}, \bibinfo{author}{Trudel, M.-C.}, \bibinfo{author}{Jaana, M.}, \& \bibinfo{author}{Kitsiou, S.} (\bibinfo{year}{2015}).
\newblock \bibinfo{title}{Synthesizing information systems knowledge: A typology of literature reviews}.
\newblock {\it \bibinfo{journal}{Information \& Management}\/},  {\it \bibinfo{volume}{52}\/}, \bibinfo{pages}{183 -- 199}. \DOIprefix\doi{10.1016/j.im.2014.08.008}.
\bibitem[{Pasquadibisceglie et~al.(2019)Pasquadibisceglie, Appice, Castellano \& Malerba}]{pasquadibisceglie2019using}
\bibinfo{author}{Pasquadibisceglie, V.}, \bibinfo{author}{Appice, A.}, \bibinfo{author}{Castellano, G.}, \& \bibinfo{author}{Malerba, D.} (\bibinfo{year}{2019}).
\newblock \bibinfo{title}{Using convolutional neural networks for predictive process analytics}.
\newblock In {\it \bibinfo{booktitle}{{Proceedings of the 1st International Conference on Process Mining}}\/} (pp. \bibinfo{pages}{129--136}).
\newblock \bibinfo{publisher}{IEEE}.
\bibitem[{Pasquadibisceglie et~al.(2020{\natexlab{a}})Pasquadibisceglie, Appice, Castellano \& Malerba}]{pasquadibisceglie2020predictive}
\bibinfo{author}{Pasquadibisceglie, V.}, \bibinfo{author}{Appice, A.}, \bibinfo{author}{Castellano, G.}, \& \bibinfo{author}{Malerba, D.} (\bibinfo{year}{2020}{\natexlab{a}}).
\newblock \bibinfo{title}{Predictive process mining meets computer vision}.
\newblock In {\it \bibinfo{booktitle}{{Proceedings of the 18th International Conference of Business Process Management (Forum)}}\/} (pp. \bibinfo{pages}{176--192}).
\newblock \bibinfo{publisher}{Springer}.
\bibitem[{Pasquadibisceglie et~al.(2021{\natexlab{a}})Pasquadibisceglie, Appice, Castellano \& Malerba}]{pasquadibisceglie2021multi}
\bibinfo{author}{Pasquadibisceglie, V.}, \bibinfo{author}{Appice, A.}, \bibinfo{author}{Castellano, G.}, \& \bibinfo{author}{Malerba, D.} (\bibinfo{year}{2021}{\natexlab{a}}).
\newblock \bibinfo{title}{A multi-view deep learning approach for predictive business process monitoring}.
\newblock {\it \bibinfo{journal}{IEEE Transactions on Services Computing}\/},  {\it \bibinfo{volume}{15}\/}, \bibinfo{pages}{2382--2395}.
\bibitem[{Pasquadibisceglie et~al.(2020{\natexlab{b}})Pasquadibisceglie, Appice, Castellano, Malerba \& Modugno}]{pasquadibisceglie2020orange}
\bibinfo{author}{Pasquadibisceglie, V.}, \bibinfo{author}{Appice, A.}, \bibinfo{author}{Castellano, G.}, \bibinfo{author}{Malerba, D.}, \& \bibinfo{author}{Modugno, G.} (\bibinfo{year}{2020}{\natexlab{b}}).
\newblock \bibinfo{title}{{ORANGE: Outcome-oriented predictive process monitoring based on image encoding and CNNs}}.
\newblock {\it \bibinfo{journal}{IEEE Access}\/},  {\it \bibinfo{volume}{8}\/}, \bibinfo{pages}{184073--184086}.
\bibitem[{Pasquadibisceglie et~al.(2021{\natexlab{b}})Pasquadibisceglie, Castellano, Appice \& Malerba}]{pasquadibisceglie2021fox}
\bibinfo{author}{Pasquadibisceglie, V.}, \bibinfo{author}{Castellano, G.}, \bibinfo{author}{Appice, A.}, \& \bibinfo{author}{Malerba, D.} (\bibinfo{year}{2021}{\natexlab{b}}).
\newblock \bibinfo{title}{{FOX: A neuro-fuzzy model for process outcome prediction and explanation}}.
\newblock In {\it \bibinfo{booktitle}{{Proceedings of the 3rd International Conference on Process Mining}}\/} (pp. \bibinfo{pages}{112--119}).
\newblock \bibinfo{publisher}{IEEE}.
\bibitem[{Pauwels \& Calders(2020)}]{pauwels2020bayesian}
\bibinfo{author}{Pauwels, S.}, \& \bibinfo{author}{Calders, T.} (\bibinfo{year}{2020}).
\newblock \bibinfo{title}{Bayesian network based predictions of business processes}.
\newblock In {\it \bibinfo{booktitle}{{Proceedings of the 18th International Conference of Business Process Management (Forum)}}\/} (pp. \bibinfo{pages}{159--175}).
\newblock \bibinfo{publisher}{Springer}.
\bibitem[{Pauwels \& Calders(2021)}]{pauwels2021incremental}
\bibinfo{author}{Pauwels, S.}, \& \bibinfo{author}{Calders, T.} (\bibinfo{year}{2021}).
\newblock \bibinfo{title}{{Incremental predictive process monitoring: The next activity case}}.
\newblock In {\it \bibinfo{booktitle}{{Proceedings of the 19th International Conference on Business Process Management}}\/} (pp. \bibinfo{pages}{123--140}).
\newblock \bibinfo{publisher}{Springer}.
\bibitem[{Pegoraro et~al.(2021)Pegoraro, Uysal, Georgi \& van~der Aalst}]{pegorarotextaware}
\bibinfo{author}{Pegoraro, M.}, \bibinfo{author}{Uysal, M.~S.}, \bibinfo{author}{Georgi, D.~B.}, \& \bibinfo{author}{van~der Aalst, W. M.~P.} (\bibinfo{year}{2021}).
\newblock \bibinfo{title}{{Text-aware predictive monitoring of business processes}}.
\newblock In {\it \bibinfo{booktitle}{{Proceedings of 24th International Conference on Business Information Systems}}\/} (pp. \bibinfo{pages}{120--129}).
\newblock \bibinfo{publisher}{Springer}.
\bibitem[{Pery et~al.(2022)Pery, Rafiei, Simon \& van~der Aalst}]{pery2022trustworthy}
\bibinfo{author}{Pery, A.}, \bibinfo{author}{Rafiei, M.}, \bibinfo{author}{Simon, M.}, \& \bibinfo{author}{van~der Aalst, W.} (\bibinfo{year}{2022}).
\newblock \bibinfo{title}{{Trustworthy artificial intelligence and process mining: Challenges and opportunities}}.
\newblock In {\it \bibinfo{booktitle}{{Proceedings of 3rd International Conference on Process Mining (Workshops)}}\/} (pp. \bibinfo{pages}{395--407}).
\newblock \bibinfo{publisher}{IEEE}.
\bibitem[{Pfeiffer et~al.(2021)Pfeiffer, Lahann \& Fettke}]{pfeiffer2021multivariate}
\bibinfo{author}{Pfeiffer, P.}, \bibinfo{author}{Lahann, J.}, \& \bibinfo{author}{Fettke, P.} (\bibinfo{year}{2021}).
\newblock \bibinfo{title}{{Multivariate business process representation learning utilizing gramian angular fields and convolutional neural networks}}.
\newblock In {\it \bibinfo{booktitle}{{Proceedings of the 19th International Conference on Business Process Management}}\/} (pp. \bibinfo{pages}{327--344}).
\newblock \bibinfo{publisher}{Springer}.
\bibitem[{Pflug \& Rinderle-Ma(2016)}]{pflug2016application}
\bibinfo{author}{Pflug, J.}, \& \bibinfo{author}{Rinderle-Ma, S.} (\bibinfo{year}{2016}).
\newblock \bibinfo{title}{Application of dynamic instance queuing to activity sequences in cooperative business process scenarios}.
\newblock {\it \bibinfo{journal}{International Journal of Cooperative Information Systems}\/},  {\it \bibinfo{volume}{25}\/}, \bibinfo{pages}{1650002}.
\bibitem[{Pham et~al.(2021)Pham, Ahn, Kim \& Kim}]{pham2021process}
\bibinfo{author}{Pham, D.-L.}, \bibinfo{author}{Ahn, H.}, \bibinfo{author}{Kim, K.-S.}, \& \bibinfo{author}{Kim, K.~P.} (\bibinfo{year}{2021}).
\newblock \bibinfo{title}{{Process-aware enterprise social network prediction and experiment using LSTM neural network models}}.
\newblock {\it \bibinfo{journal}{IEEE Access}\/},  {\it \bibinfo{volume}{9}\/}, \bibinfo{pages}{57922--57940}.
\bibitem[{Philipp et~al.(2020)Philipp, Jacob, Robert \& Beyerer}]{philipp2020predictive}
\bibinfo{author}{Philipp, P.}, \bibinfo{author}{Jacob, R.}, \bibinfo{author}{Robert, S.}, \& \bibinfo{author}{Beyerer, J.} (\bibinfo{year}{2020}).
\newblock \bibinfo{title}{Predictive analysis of business processes using neural networks with attention mechanism}.
\newblock In {\it \bibinfo{booktitle}{{Proceedings of the 2020 International Conference on Artificial Intelligence in Information and Communication}}\/} (pp. \bibinfo{pages}{225--230}).
\newblock \bibinfo{publisher}{IEEE}.
\bibitem[{Polan{\v{c}}i{\v{c}} et~al.(2020)Polan{\v{c}}i{\v{c}}, Jage{\v{c}}i{\v{c}} \& Kous}]{polanvcivc2020empirical}
\bibinfo{author}{Polan{\v{c}}i{\v{c}}, G.}, \bibinfo{author}{Jage{\v{c}}i{\v{c}}, S.}, \& \bibinfo{author}{Kous, K.} (\bibinfo{year}{2020}).
\newblock \bibinfo{title}{An empirical investigation of the effectiveness of optical recognition of hand-drawn business process elements by applying machine learning}.
\newblock {\it \bibinfo{journal}{IEEE Access}\/},  {\it \bibinfo{volume}{8}\/}, \bibinfo{pages}{206118--206131}. \DOIprefix\doi{10.1109/ACCESS.2020.3034603}.
\bibitem[{Polato et~al.(2014)Polato, Sperduti, Burattin \& de~Leoni}]{polato2014data}
\bibinfo{author}{Polato, M.}, \bibinfo{author}{Sperduti, A.}, \bibinfo{author}{Burattin, A.}, \& \bibinfo{author}{de~Leoni, M.} (\bibinfo{year}{2014}).
\newblock \bibinfo{title}{Data-aware remaining time prediction of business process instances}.
\newblock In {\it \bibinfo{booktitle}{{Proceedings of the 2014 International Joint Conference on Neural Networks}}\/} (pp. \bibinfo{pages}{816--823}).
\newblock \bibinfo{publisher}{IEEE}.
\bibitem[{Polato et~al.(2018)Polato, Sperduti, Burattin \& de~Leoni}]{polato2018time}
\bibinfo{author}{Polato, M.}, \bibinfo{author}{Sperduti, A.}, \bibinfo{author}{Burattin, A.}, \& \bibinfo{author}{de~Leoni, M.} (\bibinfo{year}{2018}).
\newblock \bibinfo{title}{Time and activity sequence prediction of business process instances}.
\newblock {\it \bibinfo{journal}{Computing}\/},  {\it \bibinfo{volume}{100}\/}, \bibinfo{pages}{1005--1031}.
\bibitem[{Qafari \& van~der Aalst(2019)}]{qafari2019fairness}
\bibinfo{author}{Qafari, M.~S.}, \& \bibinfo{author}{van~der Aalst, W. M.~P.} (\bibinfo{year}{2019}).
\newblock \bibinfo{title}{Fairness-aware process mining}.
\newblock In {\it \bibinfo{booktitle}{{Proceedings of the 2019 OTM Confederated International Conferences "On the Move to Meaningful Internet Systems"}}\/} (pp. \bibinfo{pages}{182--192}).
\newblock \bibinfo{publisher}{Springer}.
\bibitem[{Qian et~al.(2020)Qian, Wen, Kumar, Lin, Lin, Zong, Li \& Wang}]{qian2020approach}
\bibinfo{author}{Qian, C.}, \bibinfo{author}{Wen, L.}, \bibinfo{author}{Kumar, A.}, \bibinfo{author}{Lin, L.}, \bibinfo{author}{Lin, L.}, \bibinfo{author}{Zong, Z.}, \bibinfo{author}{Li, S.}, \& \bibinfo{author}{Wang, J.} (\bibinfo{year}{2020}).
\newblock \bibinfo{title}{An approach for process model extraction by multi-grained text classification}.
\newblock In {\it \bibinfo{booktitle}{{Proceedings of the 32nd International Conference on Advanced Information Systems Engineering}}\/} (pp. \bibinfo{pages}{268--282}).
\newblock \bibinfo{publisher}{Springer}.
\bibitem[{Qiao et~al.(2011)Qiao, Akkiraju \& Rembert}]{qiao2011towards}
\bibinfo{author}{Qiao, M.}, \bibinfo{author}{Akkiraju, R.}, \& \bibinfo{author}{Rembert, A.~J.} (\bibinfo{year}{2011}).
\newblock \bibinfo{title}{Towards efficient business process clustering and retrieval: combining language modeling and structure matching}.
\newblock In {\it \bibinfo{booktitle}{Proceedings of the 9th International Conference on Business Process Management}\/} (pp. \bibinfo{pages}{199--214}).
\newblock \bibinfo{organization}{Springer}.
\bibitem[{Rama-Maneiro et~al.(2021)Rama-Maneiro, Vidal \& Lama}]{rama2020deep}
\bibinfo{author}{Rama-Maneiro, E.}, \bibinfo{author}{Vidal, J.~C.}, \& \bibinfo{author}{Lama, M.} (\bibinfo{year}{2021}).
\newblock \bibinfo{title}{Deep learning for predictive business process monitoring: Review and benchmark}.
\newblock {\it \bibinfo{journal}{IEEE Transactions on Services Computing}\/},  (pp. \bibinfo{pages}{1--18}). \DOIprefix\doi{10.1109/TSC.2021.3139807}.
\bibitem[{Recker \& Mendling(2016)}]{Recker2016}
\bibinfo{author}{Recker, J.}, \& \bibinfo{author}{Mendling, J.} (\bibinfo{year}{2016}).
\newblock \bibinfo{title}{{The state of the art of business process management research as published in the BPM conference: Recommendations for progressing the field}}.
\newblock {\it \bibinfo{journal}{Business \& Information Systems Engineering}\/},  {\it \bibinfo{volume}{58}\/}, \bibinfo{pages}{55--72}. \DOIprefix\doi{10.1007/s12599-015-0411-3}.
\bibitem[{Ribeiro et~al.(2016)Ribeiro, Singh \& Guestrin}]{ribeiro2016should}
\bibinfo{author}{Ribeiro, M.~T.}, \bibinfo{author}{Singh, S.}, \& \bibinfo{author}{Guestrin, C.} (\bibinfo{year}{2016}).
\newblock \bibinfo{title}{{Why should I trust you?: Explaining the predictions of any classifier}}.
\newblock In {\it \bibinfo{booktitle}{{Proceedings of the 22nd International Conference on Knowledge Discovery and Data Mining}}\/} (pp. \bibinfo{pages}{1135--1144}).
\newblock \bibinfo{publisher}{ACM}.
\bibitem[{Rizzi et~al.(2020)Rizzi, Di~Francescomarino \& Maggi}]{rizzi2020explainability}
\bibinfo{author}{Rizzi, W.}, \bibinfo{author}{Di~Francescomarino, C.}, \& \bibinfo{author}{Maggi, F.~M.} (\bibinfo{year}{2020}).
\newblock \bibinfo{title}{Explainability in predictive process monitoring: When understanding helps improving}.
\newblock In {\it \bibinfo{booktitle}{{Proceedings of the 18th International Conference on Business Process Management (Forum)}}\/} (pp. \bibinfo{pages}{141--158}).
\newblock \bibinfo{publisher}{Springer}.
\bibitem[{Rogge-Solti \& Kasneci(2014)}]{rogge2014temporal}
\bibinfo{author}{Rogge-Solti, A.}, \& \bibinfo{author}{Kasneci, G.} (\bibinfo{year}{2014}).
\newblock \bibinfo{title}{Temporal anomaly detection in business processes}.
\newblock In {\it \bibinfo{booktitle}{{Proceedings of the 12th International Conference on Business Process Management}}\/} (pp. \bibinfo{pages}{234--249}).
\newblock \bibinfo{publisher}{Springer}.
\bibitem[{Romano et~al.(2021)Romano, Le, La~Cava, Gregg, Goldberg, Chakraborty, Ray, Himmelstein, Fu \& Moore}]{romano2021pmlb}
\bibinfo{author}{Romano, J.~D.}, \bibinfo{author}{Le, T.~T.}, \bibinfo{author}{La~Cava, W.}, \bibinfo{author}{Gregg, J.~T.}, \bibinfo{author}{Goldberg, D.~J.}, \bibinfo{author}{Chakraborty, P.}, \bibinfo{author}{Ray, N.~L.}, \bibinfo{author}{Himmelstein, D.}, \bibinfo{author}{Fu, W.}, \& \bibinfo{author}{Moore, J.~H.} (\bibinfo{year}{2021}).
\newblock \bibinfo{title}{Pmlb v1.0: an open source dataset collection for benchmarking machine learning methods}.
\newblock {\it \bibinfo{journal}{arXiv preprint arXiv:2012.00058v2}\/}, .
\bibitem[{Rozinat \& van~der Aalst(2006)}]{rozinat2006decision}
\bibinfo{author}{Rozinat, A.}, \& \bibinfo{author}{van~der Aalst, W. M.~P.} (\bibinfo{year}{2006}).
\newblock \bibinfo{title}{Decision mining in {ProM}}.
\newblock In {\it \bibinfo{booktitle}{{Proceedings of the 4th International Conference on Business Process Management}}\/} (pp. \bibinfo{pages}{420--425}).
\newblock \bibinfo{publisher}{Springer}.
\bibitem[{Rumelhart et~al.(1985)Rumelhart, Hinton \& Williams}]{rumelhart1985learning}
\bibinfo{author}{Rumelhart, D.~E.}, \bibinfo{author}{Hinton, G.~E.}, \& \bibinfo{author}{Williams, R.~J.} (\bibinfo{year}{1985}).
\newblock {\it \bibinfo{title}{{Learning Internal Representations by Error Propagation}}\/}.
\newblock \bibinfo{type}{Technical Report} California University \bibinfo{address}{USA}.
\bibitem[{Russel \& Norvig(2016)}]{Russel2016}
\bibinfo{author}{Russel, S.}, \& \bibinfo{author}{Norvig, P.} (\bibinfo{year}{2016}).
\newblock {\it \bibinfo{title}{{Artificial Intelligence: A Modern Approach}}\/}.
\newblock (\bibinfo{edition}{3rd} ed.).
\newblock \bibinfo{publisher}{Prentice Hall}.
\bibitem[{Samiri et~al.(2017)Samiri, Najib, El~Fazziki \& Boukachour}]{samiri2017toward}
\bibinfo{author}{Samiri, M.~Y.}, \bibinfo{author}{Najib, M.}, \bibinfo{author}{El~Fazziki, A.}, \& \bibinfo{author}{Boukachour, J.} (\bibinfo{year}{2017}).
\newblock \bibinfo{title}{Toward a self-adaptive workflow management system through learning and prediction models}.
\newblock {\it \bibinfo{journal}{Arabian Journal for Science and Engineering}\/},  {\it \bibinfo{volume}{42}\/}, \bibinfo{pages}{897--912}.
\bibitem[{Saraeian et~al.(2019)Saraeian, Shirazi \& Motameni}]{saraeian2019optimal}
\bibinfo{author}{Saraeian, S.}, \bibinfo{author}{Shirazi, B.}, \& \bibinfo{author}{Motameni, H.} (\bibinfo{year}{2019}).
\newblock \bibinfo{title}{Optimal autonomous architecture for uncertain processes management}.
\newblock {\it \bibinfo{journal}{Information Sciences}\/},  {\it \bibinfo{volume}{501}\/}, \bibinfo{pages}{84--99}.
\bibitem[{Sarno et~al.(2013{\natexlab{a}})Sarno, Ginardi, Pamungkas \& Sunaryono}]{sarno2013clustering}
\bibinfo{author}{Sarno, R.}, \bibinfo{author}{Ginardi, H.}, \bibinfo{author}{Pamungkas, E.~W.}, \& \bibinfo{author}{Sunaryono, D.} (\bibinfo{year}{2013}{\natexlab{a}}).
\newblock \bibinfo{title}{Clustering of erp business process fragments}.
\newblock In {\it \bibinfo{booktitle}{Proceedings of the 2013 International Conference on Computer, Control, Informatics and its Applications}\/} (pp. \bibinfo{pages}{319--324}).
\newblock \bibinfo{organization}{IEEE}.
\bibitem[{Sarno et~al.(2013{\natexlab{b}})Sarno, Sari, Ginardi, Sunaryono \& Mukhlash}]{sarno2013decision}
\bibinfo{author}{Sarno, R.}, \bibinfo{author}{Sari, P. L.~I.}, \bibinfo{author}{Ginardi, H.}, \bibinfo{author}{Sunaryono, D.}, \& \bibinfo{author}{Mukhlash, I.} (\bibinfo{year}{2013}{\natexlab{b}}).
\newblock \bibinfo{title}{Decision mining for multi choice workflow patterns}.
\newblock In {\it \bibinfo{booktitle}{{Proceedings of the 2013 International Conference on Computer, Control, Informatics and its Applications}}\/} (pp. \bibinfo{pages}{337--342}).
\newblock \bibinfo{publisher}{IEEE}.
\bibitem[{Sarno \& Sungkono(201612016a)}]{sarno2016hidden}
\bibinfo{author}{Sarno, R.}, \& \bibinfo{author}{Sungkono, K.~R.} (\bibinfo{year}{201612016a}).
\newblock \bibinfo{title}{Hidden markov model for process mining of parallel business processes}.
\newblock {\it \bibinfo{journal}{{International Review on Computers and Software}}\/},  {\it \bibinfo{volume}{11}\/}, \bibinfo{pages}{290--300}.
\bibitem[{Sarno \& Sungkono(201622016b)}]{sarno2016coupled}
\bibinfo{author}{Sarno, R.}, \& \bibinfo{author}{Sungkono, K.~R.} (\bibinfo{year}{201622016b}).
\newblock \bibinfo{title}{Coupled hidden markov model for process mining of invisible prime tasks}.
\newblock {\it \bibinfo{journal}{{International Review on Computers and Software}}\/},  {\it \bibinfo{volume}{11}\/}, \bibinfo{pages}{539--547}.
\bibitem[{Savickas \& Vasilecas(2018)}]{savickas2018belief}
\bibinfo{author}{Savickas, T.}, \& \bibinfo{author}{Vasilecas, O.} (\bibinfo{year}{2018}).
\newblock \bibinfo{title}{Belief network discovery from event logs for business process analysis}.
\newblock {\it \bibinfo{journal}{Computers in Industry}\/},  {\it \bibinfo{volume}{100}\/}, \bibinfo{pages}{258--266}.
\bibitem[{Seeliger et~al.(2021)Seeliger, Luettgen, Nolle \& Mühlhäuser}]{seeliger2021learning}
\bibinfo{author}{Seeliger, A.}, \bibinfo{author}{Luettgen, S.}, \bibinfo{author}{Nolle, T.}, \& \bibinfo{author}{Mühlhäuser, M.} (\bibinfo{year}{2021}).
\newblock \bibinfo{title}{Learning of process representations using recurrent neural networks}.
\newblock In {\it \bibinfo{booktitle}{{Proceedings of the 33rd International Conference on Advanced Information Systems Engineering}}\/} (pp. \bibinfo{pages}{109--124}).
\newblock \bibinfo{publisher}{Springer}.
\bibitem[{Seeliger et~al.(2018)Seeliger, Nolle \& Mühlhäuser}]{seeliger2018finding}
\bibinfo{author}{Seeliger, A.}, \bibinfo{author}{Nolle, T.}, \& \bibinfo{author}{Mühlhäuser, M.} (\bibinfo{year}{2018}).
\newblock \bibinfo{title}{Finding structure in the unstructured: Hybrid feature set clustering for process discovery}.
\newblock In {\it \bibinfo{booktitle}{{Proceedings of the 16th International Conference on Business Process Management}}\/} (pp. \bibinfo{pages}{288--304}).
\newblock \bibinfo{publisher}{Springer}.
\bibitem[{Senderovich et~al.(2017)Senderovich, Di~Francescomarino, Ghidini, Jorbina \& Maggi}]{senderovich2017intra}
\bibinfo{author}{Senderovich, A.}, \bibinfo{author}{Di~Francescomarino, C.}, \bibinfo{author}{Ghidini, C.}, \bibinfo{author}{Jorbina, K.}, \& \bibinfo{author}{Maggi, F.~M.} (\bibinfo{year}{2017}).
\newblock \bibinfo{title}{Intra and inter-case features in predictive process monitoring: A tale of two dimensions}.
\newblock In {\it \bibinfo{booktitle}{{Proceedings of the 15th International Conference on Business Process Management}}\/} (pp. \bibinfo{pages}{306--323}).
\newblock \bibinfo{publisher}{Springer}.
\bibitem[{Senderovich et~al.(2014)Senderovich, Weidlich, Gal \& Mandelbaum}]{senderovich2014mining}
\bibinfo{author}{Senderovich, A.}, \bibinfo{author}{Weidlich, M.}, \bibinfo{author}{Gal, A.}, \& \bibinfo{author}{Mandelbaum, A.} (\bibinfo{year}{2014}).
\newblock \bibinfo{title}{Mining resource scheduling protocols}.
\newblock In {\it \bibinfo{booktitle}{{Proceedings of the 12th International Conference on Business Process Management}}\/} (pp. \bibinfo{pages}{200--216}).
\newblock \bibinfo{publisher}{Springer}.
\bibitem[{Senderovich et~al.(2015)Senderovich, Weidlich, Gal \& Mandelbaum}]{senderovich2015queue}
\bibinfo{author}{Senderovich, A.}, \bibinfo{author}{Weidlich, M.}, \bibinfo{author}{Gal, A.}, \& \bibinfo{author}{Mandelbaum, A.} (\bibinfo{year}{2015}).
\newblock \bibinfo{title}{Queue mining for delay prediction in multi-class service processes}.
\newblock {\it \bibinfo{journal}{Information Systems}\/},  {\it \bibinfo{volume}{53}\/}, \bibinfo{pages}{278--295}.
\bibitem[{Settles(2009)}]{settles2009active}
\bibinfo{author}{Settles, B.} (\bibinfo{year}{2009}).
\newblock {\it \bibinfo{title}{{Active Learning Literature Survey}}\/}.
\newblock \bibinfo{type}{Technical Report} {University of Wisconsin-Madison} \bibinfo{address}{USA}.
\bibitem[{Shraga et~al.(2020)Shraga, Gal, Schumacher, Senderovich \& Weidlich}]{shraga2020process}
\bibinfo{author}{Shraga, R.}, \bibinfo{author}{Gal, A.}, \bibinfo{author}{Schumacher, D.}, \bibinfo{author}{Senderovich, A.}, \& \bibinfo{author}{Weidlich, M.} (\bibinfo{year}{2020}).
\newblock \bibinfo{title}{Process discovery with context-aware process trees}.
\newblock {\it \bibinfo{journal}{Information Systems}\/},  {\it \bibinfo{volume}{106}\/}, \bibinfo{pages}{101533}.
\bibitem[{Silva et~al.(2005)Silva, Zhang \& Shanahan}]{silva2005probabilistic}
\bibinfo{author}{Silva, R.}, \bibinfo{author}{Zhang, J.}, \& \bibinfo{author}{Shanahan, J.~G.} (\bibinfo{year}{2005}).
\newblock \bibinfo{title}{Probabilistic workflow mining}.
\newblock In {\it \bibinfo{booktitle}{{Proceedings of the 11th International Conference on Knowledge Discovery in Data Mining}}\/} (pp. \bibinfo{pages}{275--284}).
\newblock \bibinfo{publisher}{ACM}.
\bibitem[{Sim et~al.(2022)Sim, Sutrisnowati, Won, Lee \& Bae}]{sim2022automatic}
\bibinfo{author}{Sim, S.}, \bibinfo{author}{Sutrisnowati, R.~A.}, \bibinfo{author}{Won, S.}, \bibinfo{author}{Lee, S.}, \& \bibinfo{author}{Bae, H.} (\bibinfo{year}{2022}).
\newblock \bibinfo{title}{Automatic conversion of event data to event logs using cnn and event density embedding}.
\newblock {\it \bibinfo{journal}{IEEE Access}\/},  {\it \bibinfo{volume}{10}\/}, \bibinfo{pages}{15994--16009}.
\bibitem[{Sindhgatta et~al.(2020)Sindhgatta, Moreira, Ouyang \& Barros}]{sindhgatta2020exploring}
\bibinfo{author}{Sindhgatta, R.}, \bibinfo{author}{Moreira, C.}, \bibinfo{author}{Ouyang, C.}, \& \bibinfo{author}{Barros, A.} (\bibinfo{year}{2020}).
\newblock \bibinfo{title}{Exploring interpretable predictive models for business processes}.
\newblock In {\it \bibinfo{booktitle}{{Proceedings of the 18th International Conference on Business Process Management}}\/} (pp. \bibinfo{pages}{257--272}).
\newblock \bibinfo{publisher}{Springer}.
\bibitem[{Sommers et~al.(2021)Sommers, Menkovski \& Fahland}]{sommers2021process}
\bibinfo{author}{Sommers, D.}, \bibinfo{author}{Menkovski, V.}, \& \bibinfo{author}{Fahland, D.} (\bibinfo{year}{2021}).
\newblock \bibinfo{title}{Process discovery using graph neural networks}.
\newblock In {\it \bibinfo{booktitle}{{Proceedings of the 3rd International Conference on Process Mining}}\/} (pp. \bibinfo{pages}{40--47}).
\newblock \bibinfo{publisher}{IEEE}.
\bibitem[{Stierle et~al.(2021{\natexlab{a}})Stierle, Brunk, Weinzierl, Zilker, Matzner \& Becker}]{stierlexpbpm}
\bibinfo{author}{Stierle, M.}, \bibinfo{author}{Brunk, J.}, \bibinfo{author}{Weinzierl, S.}, \bibinfo{author}{Zilker, S.}, \bibinfo{author}{Matzner, M.}, \& \bibinfo{author}{Becker, J.} (\bibinfo{year}{2021}{\natexlab{a}}).
\newblock \bibinfo{title}{{Bringing light into the darkness -- A systematic literature review on explainable predictive business process monitoring techniques}}.
\newblock In {\it \bibinfo{booktitle}{{Proceedings of the 29th European Conference on Information Systems}}\/} (pp. \bibinfo{pages}{1--12}).
\newblock \bibinfo{publisher}{AIS}.
\bibitem[{Stierle et~al.(2021{\natexlab{b}})Stierle, Weinzierl, Harl \& Matzner}]{stierle2021technique}
\bibinfo{author}{Stierle, M.}, \bibinfo{author}{Weinzierl, S.}, \bibinfo{author}{Harl, M.}, \& \bibinfo{author}{Matzner, M.} (\bibinfo{year}{2021}{\natexlab{b}}).
\newblock \bibinfo{title}{A technique for determining relevance scores of process activities using graph-based neural networks}.
\newblock {\it \bibinfo{journal}{Decision Support Systems}\/},  {\it \bibinfo{volume}{144}\/}, \bibinfo{pages}{113511}. \DOIprefix\doi{10.1016/j.dss.2021.113511}.
\bibitem[{Tavares et~al.(2019)Tavares, Ceravolo, da~Costa, Damiani \& Junior}]{tavares2019overlapping}
\bibinfo{author}{Tavares, G.~M.}, \bibinfo{author}{Ceravolo, P.}, \bibinfo{author}{da~Costa, V. G.~T.}, \bibinfo{author}{Damiani, E.}, \& \bibinfo{author}{Junior, S.~B.} (\bibinfo{year}{2019}).
\newblock \bibinfo{title}{Overlapping analytic stages in online process mining}.
\newblock In {\it \bibinfo{booktitle}{{Proceedings of the 2019 IEEE International Conference on Services Computing }}\/} (pp. \bibinfo{pages}{167--175}).
\newblock \bibinfo{publisher}{IEEE}.
\bibitem[{Tax et~al.(2016)Tax, Sidorova, Haakma \& van~der Aalst}]{tax2016event}
\bibinfo{author}{Tax, N.}, \bibinfo{author}{Sidorova, N.}, \bibinfo{author}{Haakma, R.}, \& \bibinfo{author}{van~der Aalst, W. M.~P.} (\bibinfo{year}{2016}).
\newblock \bibinfo{title}{Event abstraction for process mining using supervised learning techniques}.
\newblock In {\it \bibinfo{booktitle}{{Proceedings of SAI Intelligent Systems Conference}}\/} (pp. \bibinfo{pages}{251--269}).
\newblock \bibinfo{publisher}{Springer}.
\bibitem[{Tax et~al.(2017)Tax, Verenich, La~Rosa \& Dumas}]{tax2017predictive}
\bibinfo{author}{Tax, N.}, \bibinfo{author}{Verenich, I.}, \bibinfo{author}{La~Rosa, M.}, \& \bibinfo{author}{Dumas, M.} (\bibinfo{year}{2017}).
\newblock \bibinfo{title}{{Predictive business process monitoring with LSTM neural networks}}.
\newblock In {\it \bibinfo{booktitle}{{Proceedings of the 29th International Conference on Advanced Information Systems Engineering}}\/} (pp. \bibinfo{pages}{477--492}).
\newblock \bibinfo{publisher}{Springer}.
\bibitem[{Taymouri et~al.(2021{\natexlab{a}})Taymouri, La~Rosa, Dumas \& Maggi}]{Taymouri2021}
\bibinfo{author}{Taymouri, F.}, \bibinfo{author}{La~Rosa, M.}, \bibinfo{author}{Dumas, M.}, \& \bibinfo{author}{Maggi, F.~M.} (\bibinfo{year}{2021}{\natexlab{a}}).
\newblock \bibinfo{title}{Business process variant analysis: Survey and classification}.
\newblock {\it \bibinfo{journal}{Knowledge-Based Systems}\/},  {\it \bibinfo{volume}{211}\/}, \bibinfo{pages}{106557}. \DOIprefix\doi{10.1016/j.knosys.2020.106557}.
\bibitem[{Taymouri et~al.(2021{\natexlab{b}})Taymouri, La~Rosa \& Erfani}]{taymouri2021deep}
\bibinfo{author}{Taymouri, F.}, \bibinfo{author}{La~Rosa, M.}, \& \bibinfo{author}{Erfani, S.} (\bibinfo{year}{2021}{\natexlab{b}}).
\newblock \bibinfo{title}{A deep adversarial model for suffix and remaining time prediction of event sequences}.
\newblock In {\it \bibinfo{booktitle}{{Proceedings of the 2021 SIAM International Conference on Data Mining}}\/} (pp. \bibinfo{pages}{522--530}).
\newblock \bibinfo{publisher}{SIAM}.
\bibitem[{Taymouri et~al.(2020)Taymouri, La~Rosa, Erfani, Bozorgi \& Verenich}]{taymouri2020predictive}
\bibinfo{author}{Taymouri, F.}, \bibinfo{author}{La~Rosa, M.}, \bibinfo{author}{Erfani, S.}, \bibinfo{author}{Bozorgi, Z.~D.}, \& \bibinfo{author}{Verenich, I.} (\bibinfo{year}{2020}).
\newblock \bibinfo{title}{{Predictive business process monitoring via generative adversarial nets: The case of next event prediction}}.
\newblock In {\it \bibinfo{booktitle}{{Proceedings of the 18th International Conference on Business Process Management}}\/} (pp. \bibinfo{pages}{237--256}).
\newblock \bibinfo{publisher}{Springer}.
\bibitem[{Teinemaa et~al.(2019)Teinemaa, Dumas, La~Rosa \& Maggi}]{teinemaa2019outcome}
\bibinfo{author}{Teinemaa, I.}, \bibinfo{author}{Dumas, M.}, \bibinfo{author}{La~Rosa, M.}, \& \bibinfo{author}{Maggi, F.~M.} (\bibinfo{year}{2019}).
\newblock \bibinfo{title}{Outcome-oriented predictive process monitoring: Review and benchmark}.
\newblock {\it \bibinfo{journal}{ACM Transactions on Knowledge Discovery from Data}\/},  {\it \bibinfo{volume}{13}\/}, \bibinfo{pages}{1--57}. \DOIprefix\doi{10.1145/3301300}.
\bibitem[{Teinemaa et~al.(2016)Teinemaa, Dumas, Maggi \& Di~Francescomarino}]{teinemaa2016predictive}
\bibinfo{author}{Teinemaa, I.}, \bibinfo{author}{Dumas, M.}, \bibinfo{author}{Maggi, F.~M.}, \& \bibinfo{author}{Di~Francescomarino, C.} (\bibinfo{year}{2016}).
\newblock \bibinfo{title}{Predictive business process monitoring with structured and unstructured data}.
\newblock In {\it \bibinfo{booktitle}{{Proceedings of the 14th International Conference on Business Process Management}}\/} (pp. \bibinfo{pages}{401--417}).
\newblock \bibinfo{publisher}{Springer}.
\bibitem[{Teinemaa et~al.(2018)Teinemaa, Tax, de~Leoni, Dumas \& Maggi}]{teinemaa2018alarm}
\bibinfo{author}{Teinemaa, I.}, \bibinfo{author}{Tax, N.}, \bibinfo{author}{de~Leoni, M.}, \bibinfo{author}{Dumas, M.}, \& \bibinfo{author}{Maggi, F.~M.} (\bibinfo{year}{2018}).
\newblock \bibinfo{title}{Alarm-based prescriptive process monitoring}.
\newblock In {\it \bibinfo{booktitle}{{Proceedings of the 16th International Conference on Business Process Management (Forum)}}\/} (pp. \bibinfo{pages}{91--107}).
\newblock \bibinfo{publisher}{Springer}.
\bibitem[{Tello et~al.(2019)Tello, Gianini, Mizouni \& Damiani}]{tello2019machine}
\bibinfo{author}{Tello, G.}, \bibinfo{author}{Gianini, G.}, \bibinfo{author}{Mizouni, R.}, \& \bibinfo{author}{Damiani, E.} (\bibinfo{year}{2019}).
\newblock \bibinfo{title}{Machine learning-based framework for log-lifting in business process mining applications}.
\newblock In {\it \bibinfo{booktitle}{{Proceedings of the 17th International Conference on Business Process Management}}\/} (pp. \bibinfo{pages}{232--249}).
\newblock \bibinfo{publisher}{Springer}.
\bibitem[{Theis \& Darabi(2019)}]{theis2019decay}
\bibinfo{author}{Theis, J.}, \& \bibinfo{author}{Darabi, H.} (\bibinfo{year}{2019}).
\newblock \bibinfo{title}{Decay replay mining to predict next process events}.
\newblock {\it \bibinfo{journal}{IEEE Access}\/},  {\it \bibinfo{volume}{7}\/}, \bibinfo{pages}{119787--119803}.
\bibitem[{Theis \& Darabi(2020)}]{theis2020adversarial}
\bibinfo{author}{Theis, J.}, \& \bibinfo{author}{Darabi, H.} (\bibinfo{year}{2020}).
\newblock \bibinfo{title}{Adversarial system variant approximation to quantify process model generalization}.
\newblock {\it \bibinfo{journal}{IEEE Access}\/},  {\it \bibinfo{volume}{8}\/}, \bibinfo{pages}{194410--194427}.
\bibitem[{Tiwari et~al.(2008)Tiwari, Turner \& Majeed}]{tiwari2008review}
\bibinfo{author}{Tiwari, A.}, \bibinfo{author}{Turner, C.~J.}, \& \bibinfo{author}{Majeed, B.} (\bibinfo{year}{2008}).
\newblock \bibinfo{title}{A review of business process mining: State-of-the-art and future trends}.
\newblock {\it \bibinfo{journal}{Business Process Management Journal}\/},  {\it \bibinfo{volume}{14}\/}, \bibinfo{pages}{5--22}.
\bibitem[{Trkman(2010)}]{Trkman2010}
\bibinfo{author}{Trkman, P.} (\bibinfo{year}{2010}).
\newblock \bibinfo{title}{The critical success factors of business process management}.
\newblock {\it \bibinfo{journal}{International Journal of Information Management}\/},  {\it \bibinfo{volume}{30}\/}, \bibinfo{pages}{125--134}. \DOIprefix\doi{10.1016/j.ijinfomgt.2009.07.003}.
\bibitem[{Turner et~al.(2008)Turner, Tiwari \& Mehnen}]{turner2008genetic}
\bibinfo{author}{Turner, C.~J.}, \bibinfo{author}{Tiwari, A.}, \& \bibinfo{author}{Mehnen, J.} (\bibinfo{year}{2008}).
\newblock \bibinfo{title}{A genetic programming approach to business process mining}.
\newblock In {\it \bibinfo{booktitle}{{Proceedings of the 10th Annual Conference on Genetic and Evolutionary Computation}}\/} (pp. \bibinfo{pages}{1307--1314}).
\bibitem[{Valdés et~al.(2022{\natexlab{a}})Valdés, Céspedes-González \& Pou}]{valdes2022conformance}
\bibinfo{author}{Valdés, J.~J.}, \bibinfo{author}{Céspedes-González, Y.}, \& \bibinfo{author}{Pou, A.} (\bibinfo{year}{2022}{\natexlab{a}}).
\newblock \bibinfo{title}{Process mining as a time series analysis tool via conformance checking}.
\newblock In {\it \bibinfo{booktitle}{{Proceedings of the 2021 Future Technologies Conference}}\/} (pp. \bibinfo{pages}{636--649}).
\newblock \bibinfo{publisher}{Springer}.
\newblock \DOIprefix\doi{10.1007/978-3-030-89906-6_42}.
\bibitem[{Valdés et~al.(2022{\natexlab{b}})Valdés, Céspedes-González \& Pou}]{valdes2022graphkernel}
\bibinfo{author}{Valdés, J.~J.}, \bibinfo{author}{Céspedes-González, Y.}, \& \bibinfo{author}{Pou, A.} (\bibinfo{year}{2022}{\natexlab{b}}).
\newblock \bibinfo{title}{Process mining as a time series analysis tool via graph kernels}.
\newblock In {\it \bibinfo{booktitle}{{Proceedings of the 2021 Future Technologies Conference}}\/} (pp. \bibinfo{pages}{832--847}).
\newblock \bibinfo{publisher}{Springer}.
\newblock \DOIprefix\doi{10.1007/978-3-030-89906-6_53}.
\bibitem[{Varela-Vaca et~al.(2019)Varela-Vaca, Galindo, Ramos-Guti{\'e}rrez, G{\'o}mez-L{\'o}pez \& Benavides}]{varela2019process}
\bibinfo{author}{Varela-Vaca, {\'A}.~J.}, \bibinfo{author}{Galindo, J.~A.}, \bibinfo{author}{Ramos-Guti{\'e}rrez, B.}, \bibinfo{author}{G{\'o}mez-L{\'o}pez, M.~T.}, \& \bibinfo{author}{Benavides, D.} (\bibinfo{year}{2019}).
\newblock \bibinfo{title}{Process mining to unleash variability management: Discovering configuration workflows using logs}.
\newblock In {\it \bibinfo{booktitle}{{Proceedings of the 23rd International Systems and Software Product Line Conference}}\/} (pp. \bibinfo{pages}{265--276}).
\newblock \bibinfo{publisher}{ACM}.
\bibitem[{V{\'a}zquez-Barreiros et~al.(2014)V{\'a}zquez-Barreiros, Mucientes \& Lama}]{vazquez2014genetic}
\bibinfo{author}{V{\'a}zquez-Barreiros, B.}, \bibinfo{author}{Mucientes, M.}, \& \bibinfo{author}{Lama, M.} (\bibinfo{year}{2014}).
\newblock \bibinfo{title}{A genetic algorithm for process discovery guided by completeness, precision and simplicity}.
\newblock In {\it \bibinfo{booktitle}{{Proceedings of the 12th International Conference on Business Process Management}}\/} (pp. \bibinfo{pages}{118--133}).
\newblock \bibinfo{publisher}{Springer}.
\bibitem[{V{\'a}zquez-Barreiros et~al.(2015)V{\'a}zquez-Barreiros, Mucientes \& Lama}]{vazquez2015prodigen}
\bibinfo{author}{V{\'a}zquez-Barreiros, B.}, \bibinfo{author}{Mucientes, M.}, \& \bibinfo{author}{Lama, M.} (\bibinfo{year}{2015}).
\newblock \bibinfo{title}{{ProDiGen: Mining complete, precise and minimal structure process models with a genetic algorithm}}.
\newblock {\it \bibinfo{journal}{Information Sciences}\/},  {\it \bibinfo{volume}{294}\/}, \bibinfo{pages}{315--333}.
\bibitem[{Verenich et~al.(2016)Verenich, Dumas, La~Rosa, Maggi \& Di~Francescomarino}]{verenich2016minimizing}
\bibinfo{author}{Verenich, I.}, \bibinfo{author}{Dumas, M.}, \bibinfo{author}{La~Rosa, M.}, \bibinfo{author}{Maggi, F.~M.}, \& \bibinfo{author}{Di~Francescomarino, C.} (\bibinfo{year}{2016}).
\newblock \bibinfo{title}{Minimizing overprocessing waste in business processes via predictive activity ordering}.
\newblock In {\it \bibinfo{booktitle}{{Proceedings of the 28th International Conference on Advanced Information Systems Engineering}}\/} (pp. \bibinfo{pages}{186--202}).
\newblock \bibinfo{publisher}{Springer}.
\bibitem[{Verenich et~al.(2019)Verenich, Dumas, La~Rosa, Maggi \& Teinemaa}]{verenich2019survey}
\bibinfo{author}{Verenich, I.}, \bibinfo{author}{Dumas, M.}, \bibinfo{author}{La~Rosa, M.}, \bibinfo{author}{Maggi, F.~M.}, \& \bibinfo{author}{Teinemaa, I.} (\bibinfo{year}{2019}).
\newblock \bibinfo{title}{Survey and cross-benchmark comparison of remaining time prediction methods in business process monitoring}.
\newblock {\it \bibinfo{journal}{ACM Transactions on Intelligent Systems and Technology}\/},  {\it \bibinfo{volume}{10}\/}, \bibinfo{pages}{1--34}. \DOIprefix\doi{10.1145/3331449}.
\bibitem[{Vergidis et~al.(2007)Vergidis, Tiwari, Majeed \& Roy}]{vergidis2007optimisation}
\bibinfo{author}{Vergidis, K.}, \bibinfo{author}{Tiwari, A.}, \bibinfo{author}{Majeed, B.}, \& \bibinfo{author}{Roy, R.} (\bibinfo{year}{2007}).
\newblock \bibinfo{title}{Optimisation of business process designs: An algorithmic approach with multiple objectives}.
\newblock {\it \bibinfo{journal}{International Journal of Production Economics}\/},  {\it \bibinfo{volume}{109}\/}, \bibinfo{pages}{105--121}.
\bibitem[{Vidgof et~al.(2023)Vidgof, Bachhofner \& Mendling}]{vidgof2023large}
\bibinfo{author}{Vidgof, M.}, \bibinfo{author}{Bachhofner, S.}, \& \bibinfo{author}{Mendling, J.} (\bibinfo{year}{2023}).
\newblock \bibinfo{title}{Large language models for business process management: Opportunities and challenges}.
\newblock {\it \bibinfo{journal}{arXiv preprint arXiv:2304.04309}\/},  (pp. \bibinfo{pages}{1--16}). \href{http://arxiv.org/abs/2304.04309}{\tt arXiv:2304.04309}.
\bibitem[{Voulodimos et~al.(2018)Voulodimos, Doulamis, Doulamis \& Protopapadakis}]{voulodimos2018deep}
\bibinfo{author}{Voulodimos, A.}, \bibinfo{author}{Doulamis, N.}, \bibinfo{author}{Doulamis, A.}, \& \bibinfo{author}{Protopapadakis, E.} (\bibinfo{year}{2018}).
\newblock \bibinfo{title}{Deep learning for computer vision: A brief review}.
\newblock {\it \bibinfo{journal}{Computational Intelligence and Neuroscience}\/},  {\it \bibinfo{volume}{2018}\/}.
\bibitem[{Wahid et~al.(2019)Wahid, Adi, Bae \& Choi}]{wahid2019predictive}
\bibinfo{author}{Wahid, N.~A.}, \bibinfo{author}{Adi, T.~N.}, \bibinfo{author}{Bae, H.}, \& \bibinfo{author}{Choi, Y.} (\bibinfo{year}{2019}).
\newblock \bibinfo{title}{{Predictive business process monitoring -- Remaining time prediction using deep neural network with entity embedding}}.
\newblock {\it \bibinfo{journal}{Procedia Computer Science}\/},  {\it \bibinfo{volume}{161}\/}, \bibinfo{pages}{1080--1088}.
\bibitem[{Wang et~al.(2012)Wang, Wong, Ding, Guo \& Wen}]{wang2012efficient}
\bibinfo{author}{Wang, J.}, \bibinfo{author}{Wong, R.~K.}, \bibinfo{author}{Ding, J.}, \bibinfo{author}{Guo, Q.}, \& \bibinfo{author}{Wen, L.} (\bibinfo{year}{2012}).
\newblock \bibinfo{title}{Efficient selection of process mining algorithms}.
\newblock {\it \bibinfo{journal}{IEEE Transactions on Services Computing}\/},  {\it \bibinfo{volume}{6}\/}, \bibinfo{pages}{484--496}.
\bibitem[{Wang et~al.(2019)Wang, Yu, Liu \& Sun}]{wang2019outcome}
\bibinfo{author}{Wang, J.}, \bibinfo{author}{Yu, D.}, \bibinfo{author}{Liu, C.}, \& \bibinfo{author}{Sun, X.} (\bibinfo{year}{2019}).
\newblock \bibinfo{title}{{Outcome-oriented predictive process monitoring with attention-based bidirectional LSTM neural networks}}.
\newblock In {\it \bibinfo{booktitle}{{Proceedings of the 2019 IEEE International Conference on Web Services}}\/} (pp. \bibinfo{pages}{360--367}).
\newblock \bibinfo{publisher}{IEEE}.
\bibitem[{Wanner et~al.(2020)Wanner, Wissuchek \& Janiesch}]{wanner2020machine}
\bibinfo{author}{Wanner, J.}, \bibinfo{author}{Wissuchek, C.}, \& \bibinfo{author}{Janiesch, C.} (\bibinfo{year}{2020}).
\newblock \bibinfo{title}{Machine learning and complex event processing: A review of real-time data analytics for the industrial internet of things}.
\newblock {\it \bibinfo{journal}{Enterprise Modelling and Information Systems Architectures (EMISAJ)}\/},  {\it \bibinfo{volume}{15}\/}, \bibinfo{pages}{1--1}.
\bibitem[{Watkins(1989)}]{watkins1989learning}
\bibinfo{author}{Watkins, C. J. C.~H.} (\bibinfo{year}{1989}).
\newblock \bibinfo{title}{{Learning from Delayed Rewards}}.
\bibitem[{Webster \& Watson(2002)}]{webster2002analyzing}
\bibinfo{author}{Webster, J.}, \& \bibinfo{author}{Watson, R.~T.} (\bibinfo{year}{2002}).
\newblock \bibinfo{title}{Analyzing the past to prepare for the future: Writing a literature review}.
\newblock {\it \bibinfo{journal}{MIS Quarterly}\/},  (pp. \bibinfo{pages}{xiii--xxiii}).
\bibitem[{Weinzierl et~al.(2020{\natexlab{a}})Weinzierl, Dunzer, Zilker \& Matzner}]{weinzierl2020prescriptive}
\bibinfo{author}{Weinzierl, S.}, \bibinfo{author}{Dunzer, S.}, \bibinfo{author}{Zilker, S.}, \& \bibinfo{author}{Matzner, M.} (\bibinfo{year}{2020}{\natexlab{a}}).
\newblock \bibinfo{title}{Prescriptive business process monitoring for recommending next best actions}.
\newblock In {\it \bibinfo{booktitle}{{Proceedings of the 18th International Conference of Business Process Management (Forum)}}\/} (pp. \bibinfo{pages}{193--209}).
\newblock \bibinfo{publisher}{Springer}.
\bibitem[{Weinzierl et~al.(2022)Weinzierl, Wolf, Pauli, Beverungen \& Matzner}]{weinzierl2022detecting}
\bibinfo{author}{Weinzierl, S.}, \bibinfo{author}{Wolf, V.}, \bibinfo{author}{Pauli, T.}, \bibinfo{author}{Beverungen, D.}, \& \bibinfo{author}{Matzner, M.} (\bibinfo{year}{2022}).
\newblock \bibinfo{title}{{Detecting temporal workarounds in business processes -- A deep-learning-based method for analysing event log data}}.
\newblock {\it \bibinfo{journal}{Journal of Business Analytics}\/},  {\it \bibinfo{volume}{5}\/}, \bibinfo{pages}{76--100}.
\bibitem[{Weinzierl et~al.(2020{\natexlab{b}})Weinzierl, Zilker, Stierle, Matzner \& Park}]{weinzierl2020predictiveWi}
\bibinfo{author}{Weinzierl, S.}, \bibinfo{author}{Zilker, S.}, \bibinfo{author}{Stierle, M.}, \bibinfo{author}{Matzner, M.}, \& \bibinfo{author}{Park, G.} (\bibinfo{year}{2020}{\natexlab{b}}).
\newblock \bibinfo{title}{From predictive to prescriptive process monitoring: Recommending the next best actions instead of calculating the next most likely events.}
\newblock In {\it \bibinfo{booktitle}{Proceedings of the 15th International Conference on Wirtschaftsinformatik}\/} (pp. \bibinfo{pages}{364--368}).
\bibitem[{Wetzstein et~al.(2009)Wetzstein, Leitner, Rosenberg, Brandic, Dustdar \& Leymann}]{wetzstein2009monitoring}
\bibinfo{author}{Wetzstein, B.}, \bibinfo{author}{Leitner, P.}, \bibinfo{author}{Rosenberg, F.}, \bibinfo{author}{Brandic, I.}, \bibinfo{author}{Dustdar, S.}, \& \bibinfo{author}{Leymann, F.} (\bibinfo{year}{2009}).
\newblock \bibinfo{title}{Monitoring and analyzing influential factors of business process performance}.
\newblock In {\it \bibinfo{booktitle}{{Proceedings of the 2009 IEEE International Enterprise Distributed Object Computing Conference}}\/} (pp. \bibinfo{pages}{141--150}).
\newblock \bibinfo{publisher}{IEEE}.
\bibitem[{Weytjens \& De~Weerdt(2021)}]{weytjens2021learning}
\bibinfo{author}{Weytjens, H.}, \& \bibinfo{author}{De~Weerdt, J.} (\bibinfo{year}{2021}).
\newblock \bibinfo{title}{Learning uncertainty with artificial neural networks for improved remaining time prediction of business processes}.
\newblock In {\it \bibinfo{booktitle}{{Proceedings of the 19th International Conference on Business Process Management}}\/} (pp. \bibinfo{pages}{141--157}).
\newblock \bibinfo{publisher}{Springer}.
\bibitem[{Weytjens \& Weerdt(2021)}]{weytjens2021creating}
\bibinfo{author}{Weytjens, H.}, \& \bibinfo{author}{Weerdt, J.~D.} (\bibinfo{year}{2021}).
\newblock \bibinfo{title}{Creating unbiased public benchmark datasets with data leakage prevention for predictive process monitoring}.
\newblock In {\it \bibinfo{booktitle}{{Proceedings of the 19th International Conference on Business Process Management (Workshops)}}\/} (pp. \bibinfo{pages}{18--29}).
\newblock \bibinfo{publisher}{Springer}.
\bibitem[{Wibisono et~al.(2015)Wibisono, Nisafani, Bae \& Park}]{wibisono2015fly}
\bibinfo{author}{Wibisono, A.}, \bibinfo{author}{Nisafani, A.~S.}, \bibinfo{author}{Bae, H.}, \& \bibinfo{author}{Park, Y.-J.} (\bibinfo{year}{2015}).
\newblock \bibinfo{title}{{On-the-fly performance-aware human resource allocation in the business process management systems environment using na{\"\i}ve Bayes}}.
\newblock In {\it \bibinfo{booktitle}{{Proceedings of the 3rd Asia-Pacific Conference on Business Process Management}}\/} (pp. \bibinfo{pages}{70--80}).
\newblock \bibinfo{publisher}{Springer}.
\bibitem[{Wickramanayake et~al.(2022)Wickramanayake, He, Ouyang, Moreira, Xu \& Sindhgatta}]{wickramanayake2022building}
\bibinfo{author}{Wickramanayake, B.}, \bibinfo{author}{He, Z.}, \bibinfo{author}{Ouyang, C.}, \bibinfo{author}{Moreira, C.}, \bibinfo{author}{Xu, Y.}, \& \bibinfo{author}{Sindhgatta, R.} (\bibinfo{year}{2022}).
\newblock \bibinfo{title}{Building interpretable models for business process prediction using shared and specialised attention mechanisms}.
\newblock {\it \bibinfo{journal}{Knowledge-Based Systems}\/},  {\it \bibinfo{volume}{248}\/}, \bibinfo{pages}{108773}.
\bibitem[{Wu et~al.(2022)Wu, Xiao, Sun, Zhang, Ma \& He}]{wu2022survey}
\bibinfo{author}{Wu, X.}, \bibinfo{author}{Xiao, L.}, \bibinfo{author}{Sun, Y.}, \bibinfo{author}{Zhang, J.}, \bibinfo{author}{Ma, T.}, \& \bibinfo{author}{He, L.} (\bibinfo{year}{2022}).
\newblock \bibinfo{title}{A survey of human-in-the-loop for machine learning}.
\newblock {\it \bibinfo{journal}{Future Generation Computer Systems}\/},  {\it \bibinfo{volume}{135}\/}, \bibinfo{pages}{364--381}.
\bibitem[{Xu et~al.(2013)Xu, Tao \& Xu}]{xu2013survey}
\bibinfo{author}{Xu, C.}, \bibinfo{author}{Tao, D.}, \& \bibinfo{author}{Xu, C.} (\bibinfo{year}{2013}).
\newblock \bibinfo{title}{{A Survey on Multi-View Learning}}.
\newblock {\it \bibinfo{journal}{arXiv preprint arXiv:1304.5634}\/},  (pp. \bibinfo{pages}{1--59}). \href{http://arxiv.org/abs/1304.5634}{\tt arXiv:1304.5634}.
\bibitem[{Xu et~al.(2016)Xu, Liu, Zhao, Yongchareon \& Ding}]{xu2016resource}
\bibinfo{author}{Xu, J.}, \bibinfo{author}{Liu, C.}, \bibinfo{author}{Zhao, X.}, \bibinfo{author}{Yongchareon, S.}, \& \bibinfo{author}{Ding, Z.} (\bibinfo{year}{2016}).
\newblock \bibinfo{title}{Resource management for business process scheduling in the presence of availability constraints}.
\newblock {\it \bibinfo{journal}{ACM Transactions on Management Information Systems}\/},  {\it \bibinfo{volume}{7}\/}, \bibinfo{pages}{1--26}.
\bibitem[{Xu \& Liu(2019)}]{xu2019profile}
\bibinfo{author}{Xu, J.}, \& \bibinfo{author}{Liu, J.} (\bibinfo{year}{2019}).
\newblock \bibinfo{title}{A profile clustering based event logs repairing approach for process mining}.
\newblock {\it \bibinfo{journal}{IEEE Access}\/},  {\it \bibinfo{volume}{7}\/}, \bibinfo{pages}{17872--17881}.
\bibitem[{Yang et~al.(2018)Yang, Ouyang, Pan, Yu \& ter Hofstede}]{yang2018finding}
\bibinfo{author}{Yang, J.}, \bibinfo{author}{Ouyang, C.}, \bibinfo{author}{Pan, M.}, \bibinfo{author}{Yu, Y.}, \& \bibinfo{author}{ter Hofstede, A.~H.} (\bibinfo{year}{2018}).
\newblock \bibinfo{title}{Finding the “liberos”: Discover organizational models with overlaps}.
\newblock In {\it \bibinfo{booktitle}{{Proceedings of the 16th International Conference on Business Process Management}}\/} (pp. \bibinfo{pages}{339--355}).
\newblock \bibinfo{publisher}{Springer}.
\bibitem[{Yang et~al.(2019)Yang, Liu, Chen \& Tong}]{yang2019federated}
\bibinfo{author}{Yang, Q.}, \bibinfo{author}{Liu, Y.}, \bibinfo{author}{Chen, T.}, \& \bibinfo{author}{Tong, Y.} (\bibinfo{year}{2019}).
\newblock \bibinfo{title}{{Federated machine learning: Concept and applications}}.
\newblock {\it \bibinfo{journal}{ACM Transactions on Intelligent Systems and Technology}\/},  {\it \bibinfo{volume}{10}\/}, \bibinfo{pages}{1--19}. \DOIprefix\doi{10.1145/3298981}.
\bibitem[{Yang et~al.(2017)Yang, Dong, Sun, Zhou, Farneth, Xiong, Burd \& Marsic}]{yang2017data}
\bibinfo{author}{Yang, S.}, \bibinfo{author}{Dong, X.}, \bibinfo{author}{Sun, L.}, \bibinfo{author}{Zhou, Y.}, \bibinfo{author}{Farneth, R.~A.}, \bibinfo{author}{Xiong, H.}, \bibinfo{author}{Burd, R.~S.}, \& \bibinfo{author}{Marsic, I.} (\bibinfo{year}{2017}).
\newblock \bibinfo{title}{A data-driven process recommender framework}.
\newblock In {\it \bibinfo{booktitle}{{Proceedings of the 23rd International Conference on Knowledge Discovery and Data Mining}}\/} (pp. \bibinfo{pages}{2111--2120}).
\newblock \bibinfo{publisher}{ACM}.
\bibitem[{Yeshchenko et~al.(2019)Yeshchenko, Di~Ciccio, Mendling \& Polyvyanyy}]{yeshchenko2019comprehensive}
\bibinfo{author}{Yeshchenko, A.}, \bibinfo{author}{Di~Ciccio, C.}, \bibinfo{author}{Mendling, J.}, \& \bibinfo{author}{Polyvyanyy, A.} (\bibinfo{year}{2019}).
\newblock \bibinfo{title}{Comprehensive process drift detection with visual analytics}.
\newblock In {\it \bibinfo{booktitle}{{Proceedings of the 38th International Conference on Conceptual Modeling}}\/} (pp. \bibinfo{pages}{119--135}).
\newblock \bibinfo{publisher}{Springer}.
\bibitem[{Yeshchenko et~al.(2018)Yeshchenko, Durier, Revoredo, Mendling \& Santoro}]{yeshchenko2018context}
\bibinfo{author}{Yeshchenko, A.}, \bibinfo{author}{Durier, F.}, \bibinfo{author}{Revoredo, K.}, \bibinfo{author}{Mendling, J.}, \& \bibinfo{author}{Santoro, F.} (\bibinfo{year}{2018}).
\newblock \bibinfo{title}{Context-aware predictive process monitoring: The impact of news sentiment}.
\newblock In {\it \bibinfo{booktitle}{{Proceedings of the 2018 OTM Confederated International Conferences "On the Move to Meaningful Internet Systems"}}\/} (pp. \bibinfo{pages}{586--603}).
\newblock \bibinfo{publisher}{Springer}.
\bibitem[{Zairi(1997)}]{Zairi1997}
\bibinfo{author}{Zairi, M.} (\bibinfo{year}{1997}).
\newblock \bibinfo{title}{Business process management: A boundaryless approach to modern competitiveness}.
\newblock {\it \bibinfo{journal}{Business Process Management Journal}\/},  {\it \bibinfo{volume}{3}\/}, \bibinfo{pages}{64--80}. \DOIprefix\doi{10.1108/14637159710161585}.
\bibitem[{Zanzotto(2019)}]{zanzotto2019human}
\bibinfo{author}{Zanzotto, F.~M.} (\bibinfo{year}{2019}).
\newblock \bibinfo{title}{Human-in-the-loop artificial intelligence}.
\newblock {\it \bibinfo{journal}{Journal of Artificial Intelligence Research}\/},  {\it \bibinfo{volume}{64}\/}, \bibinfo{pages}{243--252}.
\bibitem[{Zhang \& Yang(2021)}]{zhang2021survey}
\bibinfo{author}{Zhang, Y.}, \& \bibinfo{author}{Yang, Q.} (\bibinfo{year}{2021}).
\newblock \bibinfo{title}{A survey on multi-task learning}.
\newblock {\it \bibinfo{journal}{IEEE Transactions on Knowledge and Data Engineering}\/},  (pp. \bibinfo{pages}{1--20}). \DOIprefix\doi{10.1109/TKDE.2021.3070203}.
\bibitem[{Zhao et~al.(2016)Zhao, Liu, Dai \& Ma}]{zhao2016entropy}
\bibinfo{author}{Zhao, W.}, \bibinfo{author}{Liu, H.}, \bibinfo{author}{Dai, W.}, \& \bibinfo{author}{Ma, J.} (\bibinfo{year}{2016}).
\newblock \bibinfo{title}{An entropy-based clustering ensemble method to support resource allocation in business process management}.
\newblock {\it \bibinfo{journal}{Knowledge and Information Systems}\/},  {\it \bibinfo{volume}{48}\/}, \bibinfo{pages}{305--330}. \DOIprefix\doi{10.1007/s10115-015-0879-7}.
\bibitem[{Zhao et~al.(2020)Zhao, Pu \& Jiang}]{zhao2020human}
\bibinfo{author}{Zhao, W.}, \bibinfo{author}{Pu, S.}, \& \bibinfo{author}{Jiang, D.} (\bibinfo{year}{2020}).
\newblock \bibinfo{title}{A human resource allocation method for business processes using team faultlines}.
\newblock {\it \bibinfo{journal}{Applied Intelligence}\/},  {\it \bibinfo{volume}{50}\/}, \bibinfo{pages}{2887--2900}.
\bibitem[{Zhu(2005)}]{Zhu2005SemiSupervisedLL}
\bibinfo{author}{Zhu, X.} (\bibinfo{year}{2005}).
\newblock {\it \bibinfo{title}{{Semi-Supervised Learning Literature Survey}}\/}.
\newblock \bibinfo{type}{Technical Report} {University of Wisconsin-Madison} \bibinfo{address}{USA}.

\end{thebibliography}





\end{document}